\documentclass[sigconf]{acmart}
\def\FilingLoopFullPaper{1}

\usepackage{enumitem}
\usepackage{array}
\usepackage{tabularx}
\usepackage{colortbl}
\usepackage{float}
\usepackage[breakable,skins]{tcolorbox}
\definecolor{MRFound}{HTML}{35648D}
\definecolor{MRMet}{HTML}{286D62}
\definecolor{MRPanel}{HTML}{F1F5F9}

\definecolor{PromptReviewer}{HTML}{35648D}
\definecolor{PromptFiler}{HTML}{286D62}
\definecolor{PromptEditor}{HTML}{704DA1}
\newtcolorbox{AgentPrompt}[3][]{%
  enhanced jigsaw,
  breakable,
  colback=#2!4!white,
  colframe=#2!38!white,
  colbacktitle=#2!12!white,
  coltitle=#2!90!black,
  fonttitle=\sffamily\bfseries\fontsize{9}{11}\selectfont,
  fontupper=\fontsize{9}{11.2}\selectfont,
  title={#3},
  title after break={#3\hfill\normalfont\sffamily\fontsize{8}{10}\selectfont continued},
  boxrule=0.35pt,
  leftrule=1.6pt,
  arc=1.5pt,
  outer arc=1.5pt,
  left=9pt,right=9pt,top=6pt,bottom=6pt,
  toptitle=4pt,bottomtitle=4pt,
  before skip=6pt,after skip=6pt,
  pad at break*=3pt,
  before upper={\setlength{\parindent}{0pt}\setlength{\parskip}{3pt}%
    \setlength{\emergencystretch}{1.5em}},
  #1
}
\newcommand{\PromptPart}[1]{%
  \par\smallskip\noindent\textbf{#1}\par\nobreak
}
\newcommand{\PromptField}[1]{{\fontsize{8.2}{10.3}\selectfont\texttt{#1}}}
\newenvironment{PromptItems}{%
  \begin{itemize}[leftmargin=1.2em,label=\textbullet,itemsep=1.5pt,
    topsep=2pt,parsep=0pt,partopsep=0pt]
}{\end{itemize}}

\renewcommand\footnotetextcopyrightpermission[1]{}

\fancypagestyle{filingloopfull}{%
  \fancyhf{}%
  \fancyfoot[C]{\footnotesize\thepage}%
  \renewcommand{\headrulewidth}{0pt}%
}
\acmConference{}{}{}
\acmBooktitle{}
\AtBeginDocument{%
  }

\begin{document}

\title[FilingLoop]{FilingLoop: Evaluating and Evolving LLM Agents for Corporate Disclosure Review and Revision}
% Author information is enabled only by sigconf-full.tex.
\author{Wentao Zhang}
\affiliation{%
  \institution{Nanyang Technological University}
  \country{Singapore}}

\author{Yilei Zhao}
\affiliation{%
  \institution{Nanyang Technological University}
  \country{Singapore}}

\author{Guangyan Gan}
\affiliation{%
  \institution{Nanyang Technological University}
  \country{Singapore}}

\author{Shuo Sun}
\affiliation{%
  \institution{The Hong Kong University of Science and Technology (Guangzhou)}
  \city{Guangzhou}
  \country{China}}

\author{Lin William Cong}
\authornote{Corresponding author.\newline Email (Wentao Zhang): \href{mailto:zhangwent963@gmail.com}{\nolinkurl{zhangwent963@gmail.com}}.}
\affiliation{%
  \institution{Nanyang Technological University}
  \country{Singapore}}

\renewcommand{\shortauthors}{Zhang et al.}

\begin{abstract}
Public disclosure repositories such as SEC EDGAR support informed investment through web access to corporate filings.
Supporting their review with language model agents requires discovering unstated deficiencies across related documents, acquiring missing company facts, and verifying that requested revisions satisfy evidence-backed requirements.
Existing financial question-answering evaluations provide limited evidence of these interactive capabilities.
We introduce \textbf{FilingLoop}, an interactive benchmark and a self-evolving agent framework for corporate disclosure review and revision.
Grounded in historical U.S.\ Securities and Exchange Commission correspondence, FilingLoop reconstructs 380 predominantly IPO review tasks from 96 companies under distinct information constraints.
An evaluator separate from the acting agents scores final filings and responses against 6,099 eligible requirements and audits unsupported claims and harmful revisions.
Our framework couples iterative review and revision within each case with procedural learning across cases, converting requirement-level evaluation feedback into role-specific guidance without updating model weights.
Updates affect only subsequent cases, allowing procedural adaptation to be assessed without retrying scored tasks.
On the evaluated stream, evolution increases requirement fulfillment from 24.28\% to 27.96\% and full comment resolution from 16.90\% to 20.75\%, with a 5.8\% increase in total cost over the fixed-policy variant.
FilingLoop connects the concerns an agent discovers with the company facts it obtains and the revisions it commits, providing requirement-level feedback for procedures used in later review tasks.
\end{abstract}

\maketitle
\pagestyle{filingloopfull}
\raggedbottom

% Generated from the same aggregate as the benchmark overview figure.
\newcommand{\BenchCompanies}{96}
\newcommand{\BenchPackages}{98}
\newcommand{\BenchTasks}{380}
\newcommand{\BenchComments}{3,012}
\newcommand{\BenchRequirements}{6,099}
\newcommand{\BenchReviewedFilings}{378}
\newcommand{\BenchReviewStart}{May 2016}
\newcommand{\BenchReviewEnd}{March 2026}
\newcommand{\BenchRegistrationPackages}{97}
\newcommand{\BenchRegistrationTasks}{378}
\newcommand{\BenchAnnualPackages}{1}
\newcommand{\BenchAnnualTasks}{2}
\newcommand{\BenchCompoundPct}{62.7\%}
\newcommand{\BenchFactsPct}{50.5\%}
\newcommand{\BenchDisclosurePct}{76.5\%}
\newcommand{\BenchCommentsMedian}{4}
\newcommand{\BenchCommentsPninety}{21}
\newcommand{\BenchCommentsMax}{60}
\newcommand{\BenchUnitsMax}{6}

% Generated offline by scripts/build_main_results.py; see notes/main-results/metrics.json.
\newcommand{\MRJS}{15.32}
\newcommand{\MRCompanyCredit}{14.71}
\newcommand{\MRDR}{32.64}
\newcommand{\MRRF}{24.28}
\newcommand{\MRFR}{16.90}
\newcommand{\MRComplete}{99.74}
\newcommand{\MRcRF}{68.50}
\newcommand{\MRDiscoveredUnits}{2,162}
\newcommand{\MRCompleted}{379}
\newcommand{\MRInput}{6,098.59}
\newcommand{\MRCacheRead}{5,539.13}
\newcommand{\MRCacheWrite}{412.91}
\newcommand{\MROutput}{35.59}
\newcommand{\MRCost}{2,380.15}
\newcommand{\MRTime}{33.28}
\newcommand{\MRFixedElapsedHours}{33.28}
\newcommand{\MRActorRecordedHours}{455.04}
\newcommand{\MRActorMeanMinutes}{71.85}
\newcommand{\MRTaskConcurrency}{16}
\newcommand{\MRParallelIdealHours}{28.44}
\newcommand{\MRPausedHours}{53.02}
\newcommand{\MRCalendarHours}{86.30}
\newcommand{\MRCostPerTask}{6.26}
\newcommand{\MRSharedCost}{495.34}
\newcommand{\MRCacheRate}{90.83}
\newcommand{\MRUnitsMet}{1,481}
\newcommand{\MRCommentsFound}{983}
\newcommand{\MRCommentsResolved}{509}
\newcommand{\MRMedianMinutes}{72.8}
\newcommand{\MRMedianMinutesPrecise}{72.84}
\newcommand{\MRPninetyMinutes}{118.8}
\newcommand{\MRFilerInput}{2,848.36}
\newcommand{\MRFilerOutput}{16.47}
\newcommand{\MRFilerCost}{740.30}
\newcommand{\MRFilerAcceptedPercent}{99.93}
\newcommand{\MRFilerAccepted}{1,473}
\newcommand{\MRFilerInvocations}{1,474}
\newcommand{\MRFilerCompletePercent}{99.74}
\newcommand{\MRFilerCompleteCases}{379}
\newcommand{\MRFilerMedianMinutes}{32.86}
\newcommand{\MRFilerPninetyMinutes}{58.22}
\newcommand{\MRFilerRecordedHours}{226.78}
\newcommand{\MRFilerMeanMinutes}{35.81}
\newcommand{\MRFilerScore}{51.13}
\newcommand{\MRFilerScoreCredit}{48.06}
\newcommand{\MRFilerScoreCompanies}{94}
\newcommand{\MRReviewerInput}{3,041.21}
\newcommand{\MRReviewerOutput}{15.47}
\newcommand{\MRReviewerCost}{1,144.51}
\newcommand{\MRReviewerAcceptedPercent}{100.00}
\newcommand{\MRReviewerAccepted}{1,239}
\newcommand{\MRReviewerInvocations}{1,239}
\newcommand{\MRReviewerCompletePercent}{100.00}
\newcommand{\MRReviewerCompleteCases}{380}
\newcommand{\MRReviewerMedianMinutes}{33.46}
\newcommand{\MRReviewerPninetyMinutes}{53.34}
\newcommand{\MRReviewerRecordedHours}{228.26}
\newcommand{\MRReviewerMeanMinutes}{36.04}
\newcommand{\MRReviewerScore}{30.45}
\newcommand{\MRReviewerScoreCredit}{29.23}
\newcommand{\MRReviewerScoreCompanies}{96}
\newcommand{\MRRC}{35.45}
\newcommand{\MRRCNumerator}{2,162}
\newcommand{\MRRCDenominator}{6,099}
\newcommand{\MRRRA}{61.47}
\newcommand{\MRRRANumerator}{509}
\newcommand{\MRRRADenominator}{828}
\newcommand{\MRcFR}{51.78}
\newcommand{\MRcFRNumerator}{509}
\newcommand{\MRcFRDenominator}{983}
\newcommand{\MRHER}{5.80}
\newcommand{\MRHERNumerator}{22}
\newcommand{\MRHERDenominator}{379}
\newcommand{\MRJointRequirements}{24.28}
\newcommand{\MRJointRequirementsNumerator}{1,481}
\newcommand{\MRJointRequirementsDenominator}{6,099}
\newcommand{\MRJointComments}{16.90}
\newcommand{\MRJointCommentsNumerator}{509}
\newcommand{\MRJointCommentsDenominator}{3,012}
\newcommand{\MRFixedTaskAcc}{3.95}
\newcommand{\MRFixedTaskAccNumerator}{15}
\newcommand{\MRFixedTaskAccDenominator}{380}
\newcommand{\MRReviewerTaskAcc}{13.16}
\newcommand{\MRReviewerTaskAccNumerator}{50}
\newcommand{\MRReviewerTaskAccDenominator}{380}
\newcommand{\MRFilerTaskAcc}{27.21}
\newcommand{\MRFilerTaskAccNumerator}{77}
\newcommand{\MRFilerTaskAccDenominator}{283}
\newcommand{\MRFixedCompanyAcc}{1.04}
\newcommand{\MRFixedCompanyAccNumerator}{1}
\newcommand{\MRFixedCompanyAccDenominator}{96}
\newcommand{\MRReviewerCompanyAcc}{1.04}
\newcommand{\MRReviewerCompanyAccNumerator}{1}
\newcommand{\MRReviewerCompanyAccDenominator}{96}
\newcommand{\MRFilerCompanyAcc}{5.26}
\newcommand{\MRFilerCompanyAccNumerator}{5}
\newcommand{\MRFilerCompanyAccDenominator}{95}
\newcommand{\MRFilerEmptyTargets}{125}

% Generated offline by scripts/build_baseline_results.py; see notes/main-results/mini-swe-gpt61/metrics.json.
\newcommand{\MRSWERequirements}{21.99}
\newcommand{\MRSWERequirementsNumerator}{1,341}
\newcommand{\MRSWERequirementsDenominator}{6,099}
\newcommand{\MRSWEComments}{14.61}
\newcommand{\MRSWECommentsNumerator}{440}
\newcommand{\MRSWECommentsDenominator}{3,012}
\newcommand{\MRSWETaskAcc}{4.47}
\newcommand{\MRSWETaskAccNumerator}{17}
\newcommand{\MRSWETaskAccDenominator}{380}
\newcommand{\MRSWECompanyAcc}{0.00}
\newcommand{\MRSWECompanyAccNumerator}{0}
\newcommand{\MRSWECompanyAccDenominator}{96}
\newcommand{\MRSWEComplete}{99.74}
\newcommand{\MRSWECompleted}{379}
\newcommand{\MRSWETasks}{380}
\newcommand{\MRSWEInput}{3,676.76}
\newcommand{\MRSWEOutput}{24.14}
\newcommand{\MRSWECost}{1,353.33}
\newcommand{\MRSWEHours}{183.61}
\newcommand{\MRSWEMeanMinutes}{28.99}

% Generated by scripts/build_main_table_diagnostics.py; no inference.
\newcommand{\MTFixedFound}{2,162}
\newcommand{\MTFixedMet}{1,481}
\newcommand{\MTFixedUnassessed}{9}
\newcommand{\MTFixedTotal}{6,099}
\newcommand{\MTFixedDiscovery}{29.00}
\newcommand{\MTFixedConditional}{68.50}
\newcommand{\MTFixedReviewerInput}{3,041.21}
\newcommand{\MTFixedReviewerOutput}{15.47}
\newcommand{\MTFixedReviewerCost}{1,144.51}
\newcommand{\MTFixedReviewerHours}{228.26}
\newcommand{\MTFixedReviewerMeanMinutes}{36.04}
\newcommand{\MTFixedReviewerComplete}{100.00}
\newcommand{\MTFixedReviewerCompleted}{380}
\newcommand{\MTFixedFilerInput}{2,848.36}
\newcommand{\MTFixedFilerOutput}{16.47}
\newcommand{\MTFixedFilerCost}{740.30}
\newcommand{\MTFixedFilerHours}{226.78}
\newcommand{\MTFixedFilerMeanMinutes}{35.81}
\newcommand{\MTFixedFilerComplete}{99.74}
\newcommand{\MTFixedFilerCompleted}{379}
\newcommand{\MTSWEFound}{2,171}
\newcommand{\MTSWEMet}{1,341}
\newcommand{\MTSWEUnassessed}{1}
\newcommand{\MTSWETotal}{6,099}
\newcommand{\MTSWEDiscovery}{35.60}
\newcommand{\MTSWEConditional}{61.77}
\newcommand{\MTSWEReviewerInput}{2,330.39}
\newcommand{\MTSWEReviewerOutput}{7.37}
\newcommand{\MTSWEReviewerCost}{676.50}
\newcommand{\MTSWEReviewerHours}{89.54}
\newcommand{\MTSWEReviewerMeanMinutes}{14.14}
\newcommand{\MTSWEReviewerComplete}{100.00}
\newcommand{\MTSWEReviewerCompleted}{380}
\newcommand{\MTSWEFilerInput}{1,101.65}
\newcommand{\MTSWEFilerOutput}{11.56}
\newcommand{\MTSWEFilerCost}{497.95}
\newcommand{\MTSWEFilerHours}{94.07}
\newcommand{\MTSWEFilerMeanMinutes}{14.85}
\newcommand{\MTSWEFilerComplete}{99.74}
\newcommand{\MTSWEFilerCompleted}{379}

% FilingLoop--Evolving values used in the main comparison.
\newcommand{\MREvoRequirements}{27.96}
\newcommand{\MREvoRequirementsNumerator}{1,705}
\newcommand{\MREvoRequirementsDenominator}{6,099}
\newcommand{\MREvoComments}{20.75}
\newcommand{\MREvoCommentsNumerator}{625}
\newcommand{\MREvoCommentsDenominator}{3,012}
\newcommand{\MREvoTaskAcc}{5.00}
\newcommand{\MREvoTaskAccNumerator}{19}
\newcommand{\MREvoTaskAccDenominator}{380}
\newcommand{\MREvoCompanyAcc}{2.08}
\newcommand{\MREvoCompanyAccNumerator}{2}
\newcommand{\MREvoCompanyAccDenominator}{96}
\newcommand{\MREvoComplete}{99.74}
\newcommand{\MREvoCompleted}{379}
\newcommand{\MREvoInput}{6,432.51}
\newcommand{\MREvoOutput}{37.90}
\newcommand{\MREvoCost}{2,519.23}
\newcommand{\MREvoHours}{469.36}
\newcommand{\MREvoMeanMinutes}{74.11}
\newcommand{\MREvoReviewerRequirements}{36.86}
\newcommand{\MREvoReviewerRequirementsNumerator}{2,248}
\newcommand{\MREvoReviewerRequirementsDenominator}{6,099}
\newcommand{\MREvoReviewerComments}{33.93}
\newcommand{\MREvoReviewerCommentsNumerator}{1,022}
\newcommand{\MREvoReviewerCommentsDenominator}{3,012}
\newcommand{\MREvoReviewerTaskAcc}{14.21}
\newcommand{\MREvoReviewerTaskAccNumerator}{54}
\newcommand{\MREvoReviewerTaskAccDenominator}{380}
\newcommand{\MREvoReviewerCompanyAcc}{2.08}
\newcommand{\MREvoReviewerCompanyAccNumerator}{2}
\newcommand{\MREvoReviewerCompanyAccDenominator}{96}
\newcommand{\MREvoReviewerComplete}{100.00}
\newcommand{\MREvoReviewerCompleted}{380}
\newcommand{\MREvoReviewerInput}{3,126.36}
\newcommand{\MREvoReviewerOutput}{16.17}
\newcommand{\MREvoReviewerCost}{1,185.71}
\newcommand{\MREvoReviewerHours}{232.83}
\newcommand{\MREvoReviewerMeanMinutes}{36.76}
% Paired Filer diagnostics use the same Fixed target pool.
% Share denominators with Fixed Filer, not with the parent or Reviewer row.
\newcommand{\MREvoFilerRequirements}{78.86}
\newcommand{\MREvoFilerRequirementsNumerator}{1,705}
\newcommand{\MREvoFilerRequirementsDenominator}{\MRDiscoveredUnits}
\newcommand{\MREvoFilerComments}{63.58}
\newcommand{\MREvoFilerCommentsNumerator}{625}
\newcommand{\MREvoFilerCommentsDenominator}{\MRcFRDenominator}
\newcommand{\MREvoFilerTaskAcc}{32.16}
\newcommand{\MREvoFilerTaskAccNumerator}{91}
\newcommand{\MREvoFilerTaskAccDenominator}{\MRFilerTaskAccDenominator}
\newcommand{\MREvoFilerCompanyAcc}{6.32}
\newcommand{\MREvoFilerCompanyAccNumerator}{6}
\newcommand{\MREvoFilerCompanyAccDenominator}{\MRFilerCompanyAccDenominator}
\newcommand{\MREvoFilerComplete}{99.74}
\newcommand{\MREvoFilerCompleted}{379}
\newcommand{\MREvoFilerInput}{3,076.23}
\newcommand{\MREvoFilerOutput}{17.54}
\newcommand{\MREvoFilerCost}{793.60}
\newcommand{\MREvoFilerHours}{236.53}
\newcommand{\MREvoFilerMeanMinutes}{37.35}

% Generated from tables/main_results.tex by scripts/build_rq_analysis.py.
% pp changes subtract displayed percentages; relative gains use counts.
\newcommand{\RQFixedReq}{24.28}
\newcommand{\RQFixedReqCount}{1,481}
\newcommand{\RQFixedComment}{16.90}
\newcommand{\RQFixedCommentCount}{509}
\newcommand{\RQFixedTask}{3.95}
\newcommand{\RQFixedTaskCount}{15}
\newcommand{\RQFixedCompany}{1.04}
\newcommand{\RQFixedCompanyCount}{1}
\newcommand{\RQFixedCost}{2,380.15}
\newcommand{\RQFixedHours}{455.04}
\newcommand{\RQFixedMinutes}{71.85}
\newcommand{\RQEvoReq}{27.96}
\newcommand{\RQEvoReqCount}{1,705}
\newcommand{\RQEvoComment}{20.75}
\newcommand{\RQEvoCommentCount}{625}
\newcommand{\RQEvoTask}{5.00}
\newcommand{\RQEvoTaskCount}{19}
\newcommand{\RQEvoCompany}{2.08}
\newcommand{\RQEvoCompanyCount}{2}
\newcommand{\RQEvoCost}{2,519.23}
\newcommand{\RQEvoHours}{469.36}
\newcommand{\RQEvoMinutes}{74.11}
\newcommand{\RQOpusReq}{23.74}
\newcommand{\RQOpusReqCount}{1,448}
\newcommand{\RQOpusComment}{16.37}
\newcommand{\RQOpusCommentCount}{493}
\newcommand{\RQOpusTask}{5.26}
\newcommand{\RQOpusTaskCount}{20}
\newcommand{\RQOpusCompany}{0.00}
\newcommand{\RQOpusCompanyCount}{0}
\newcommand{\RQOpusCost}{2,327.73}
\newcommand{\RQOpusHours}{212.99}
\newcommand{\RQOpusMinutes}{33.63}
\newcommand{\RQReviewerReq}{36.86}
\newcommand{\RQReviewerReqCount}{2,248}
\newcommand{\RQReviewerComment}{33.93}
\newcommand{\RQReviewerCommentCount}{1,022}
\newcommand{\RQReviewerTask}{14.21}
\newcommand{\RQReviewerTaskCount}{54}
\newcommand{\RQReviewerCompany}{2.08}
\newcommand{\RQReviewerCompanyCount}{2}
\newcommand{\RQReviewerCost}{1,185.71}
\newcommand{\RQReviewerHours}{232.83}
\newcommand{\RQReviewerMinutes}{36.76}
\newcommand{\RQFilerReq}{78.86}
\newcommand{\RQFilerReqCount}{1,705}
\newcommand{\RQFilerComment}{63.58}
\newcommand{\RQFilerCommentCount}{625}
\newcommand{\RQFilerTask}{32.16}
\newcommand{\RQFilerTaskCount}{91}
\newcommand{\RQFilerCompany}{6.32}
\newcommand{\RQFilerCompanyCount}{6}
\newcommand{\RQFilerCost}{793.60}
\newcommand{\RQFilerHours}{236.53}
\newcommand{\RQFilerMinutes}{37.35}
\newcommand{\RQDeepSeekReq}{19.56}
\newcommand{\RQDeepSeekReqCount}{1,193}
\newcommand{\RQDeepSeekComment}{12.55}
\newcommand{\RQDeepSeekCommentCount}{378}
\newcommand{\RQDeepSeekTask}{3.42}
\newcommand{\RQDeepSeekTaskCount}{13}
\newcommand{\RQDeepSeekCompany}{0.00}
\newcommand{\RQDeepSeekCompanyCount}{0}
\newcommand{\RQDeepSeekCost}{392.47}
\newcommand{\RQDeepSeekHours}{143.22}
\newcommand{\RQDeepSeekMinutes}{22.61}
\newcommand{\RQReqGain}{3.68}
\newcommand{\RQMoreReq}{224}
\newcommand{\RQCommentGain}{3.85}
\newcommand{\RQMoreComment}{116}
\newcommand{\RQTaskGain}{1.05}
\newcommand{\RQMoreTask}{4}
\newcommand{\RQCompanyGain}{1.04}
\newcommand{\RQMoreCompany}{1}
\newcommand{\RQReviewerReqGain}{1.41}
\newcommand{\RQReviewerCommentGain}{1.29}
\newcommand{\RQFilerReqGain}{10.36}
\newcommand{\RQFilerCommentGain}{11.80}
\newcommand{\RQBestExternalReqGain}{4.22}
\newcommand{\RQBestExternalCommentGain}{4.38}
\newcommand{\RQFixedCodexReqGain}{2.15}
\newcommand{\RQFixedCodexCommentGain}{2.06}
\newcommand{\RQFixedSweReqGain}{2.29}
\newcommand{\RQFixedSweCommentGain}{2.29}
\newcommand{\RQReqGrowth}{15.1}
\newcommand{\RQCommentGrowth}{22.8}
\newcommand{\RQCostGrowth}{5.8}
\newcommand{\RQTimeGrowth}{3.1}
\newcommand{\RQInputGrowth}{5.5}
\newcommand{\RQOutputGrowth}{6.5}
\newcommand{\RQFixedUnitCost}{1.61}
\newcommand{\RQEvoUnitCost}{1.48}
\newcommand{\RQUnitSaving}{8.1}
\newcommand{\RQExtraCost}{139.08}
\newcommand{\RQExtraMinutes}{2.26}
\newcommand{\RQMinCompletion}{99.21}
\newcommand{\RQCodexCostSaving}{28.0}
\newcommand{\RQClaudeCostSaving}{32.2}

\section{Introduction}
\label{sec:introduction}

\begin{figure}[t]
  \centering
  \includegraphics[width=0.8\columnwidth]{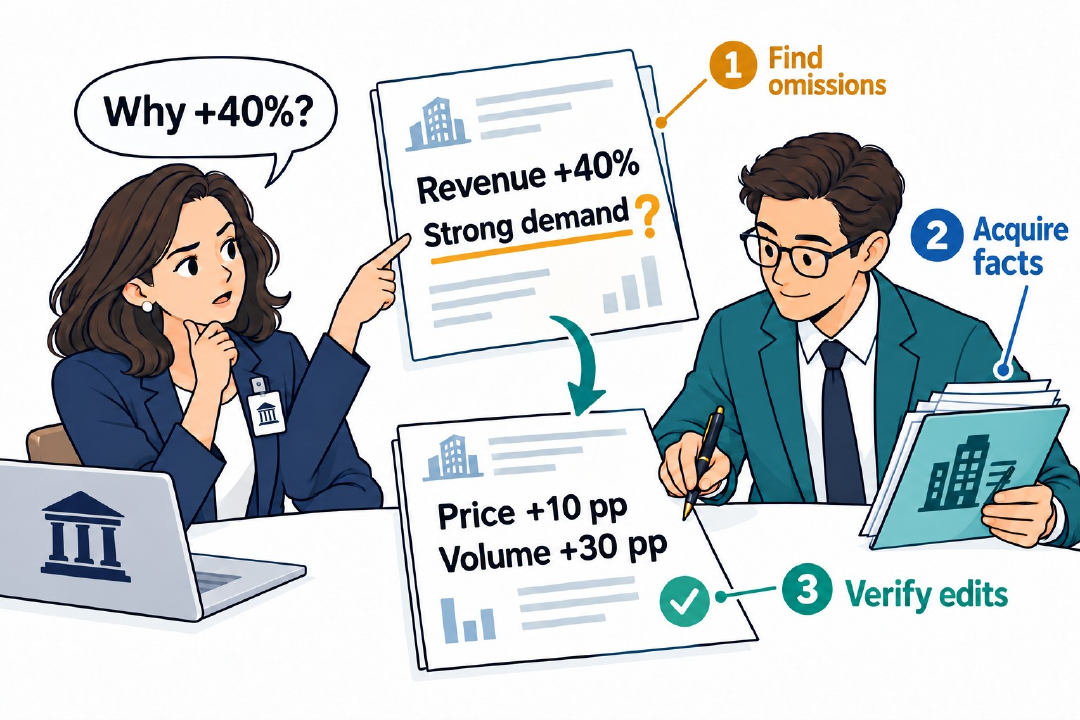}
  \caption{\textbf{Disclosure review challenges.} (1) Discover omissions, (2) acquire missing facts, and (3) verify revisions. Reviewer (left); issuer counsel (right).}
  \Description{A reviewer questions a vague revenue-growth explanation; issuer counsel checks company records and revises the disclosure. Three numbered callouts identify omission discovery, fact acquisition, and verification of document changes. Illustrative contributions of ten percentage points from price and thirty from volume explain forty percent revenue growth.}
  \label{fig:overview}
\ifdefined\FilingLoopFullPaper\else\vspace{-0.3cm}\fi
\end{figure}

The U.S. Securities and Exchange Commission (SEC) provides public access to corporate filings through its Electronic Data Gathering, Analysis, and Retrieval (EDGAR) system, enabling investors to assess company performance, risks, and accounting choices~\cite{healy2001information}. For companies preparing to go public, producing reliable disclosures involves repeated exchanges between regulatory reviewers and company counsel~\cite{johnston2017regulatory}.\footnote{\url{https://www.sec.gov/search-filings/edgar-search-assistance/accessing-edgar-data}} Reviewers identify consequential gaps across filings and correspondence, while counsel obtains facts and revises the relevant documents. As illustrated in Figure~\ref{fig:overview}, a company may explain its 40\% revenue growth simply as a result of strong demand, without providing details about the factors driving that growth. The information-access challenge is to discover omissions from the evidence available at each review round and verify their resolution in the appropriate document version.

Connecting issue discovery, evidence acquisition, and document revision poses three challenges for agent evaluation. \textbf{(1)~Discovering unstated issues.} Financial question-answering and web-workflow benchmarks evaluate predefined questions or tasks~\cite{chen2021finqa,zhu2021tat,chen2022convfinqa,islam2023financebench,krumdick2024bizbench,xie2024finben,bigeard2025finance,drouin2024workarena}, whereas reviewers must first determine which disclosure gaps warrant scrutiny. \textbf{(2)~Acquiring missing facts.} Agent benchmarks evaluate tool use and interactive workflows, including web tasks~\cite{liu2024agentbench,mialon2024gaia,drouin2024workarena}. Disclosure review additionally requires targeted requests for company facts absent from the public documents. \textbf{(3)~Verifying actual fulfillment.} Interactive benchmarks emphasize evaluation of final states~\cite{yao2024tau,barres2025tau}. In disclosure review, each requirement must be checked against the relevant document. A company may claim to have addressed a reviewer’s comment, but the requirement remains unmet if the revised filing still lacks the requested information.

\ifdefined\FilingLoopFullPaper
The revenue example also shows why a persuasive response is insufficient.
Counsel must obtain a supported breakdown, place it in the disclosure if
requested, and check that the explanation remains consistent with the rest
of the filing. A response that merely promises a revision leaves the
disclosure unchanged. Conversely, a well-executed amendment to one passage
does not resolve other requests in the same comment. Evaluating the entire
sequence requires keeping track of what was requested, what evidence became
available, and what ultimately changed in each document.
\fi

We evaluate \emph{disclosure resolution} in the final filing and response, then use requirement-level feedback to improve review procedures on later tasks. Reasoning-and-action agents, iterative refinement, and learned workflows provide building blocks~\cite{yao2022react,madaan2023self,wang2024agent}. Disclosure review requires tracing each concern from source evidence to verified corrections across document versions. Feedback on recurring investigation and verification failures can guide reusable procedures~\cite{zhao2024expel,yuksekgonul2024textgrad,agrawal2026gepa}, while company-specific facts remain separate from transferable guidance.

To study these capabilities, we introduce \textbf{FilingLoop}, an interactive benchmark and self-evolving agent framework grounded in historical SEC initial public offering (IPO) review trajectories. Agents inspect full filings while reference comments remain hidden. Reviewer and issuer counsel operate under role-specific information constraints, with evidence-bounded access to company facts. Independent evaluation checks atomic requirements in final filings and responses and audits unsupported claims and harmful actions. Evolving updates both roles' guidance from completed cases after scores are locked; guidance is frozen within batches and applies only to subsequent cases. Together, the benchmark and framework offer three contributions:
\begin{itemize}[leftmargin=0pt]
  \item \textbf{Benchmark and framework.} We reconstruct SEC review rounds, control access to historical records and company facts, and check each requested action in the final filing or response.
  \item \textbf{Procedural evolution for both roles.} We distill requirement-level feedback into role-specific guidance, coupling iterative review and revision with procedural learning across cases while keeping model weights fixed throughout the ordered task stream.
  \item \textbf{Comprehensive evaluation.} Across eight configurations on 380 tasks, FilingLoop--Evolving achieves the highest requirement and comment accuracies. Compared with Fixed, procedural evolution improves both discovery and conditional revision while reducing cost per fulfilled requirement.
\end{itemize}

\begin{figure*}[t]
  \centering
  \includegraphics[width=\textwidth]{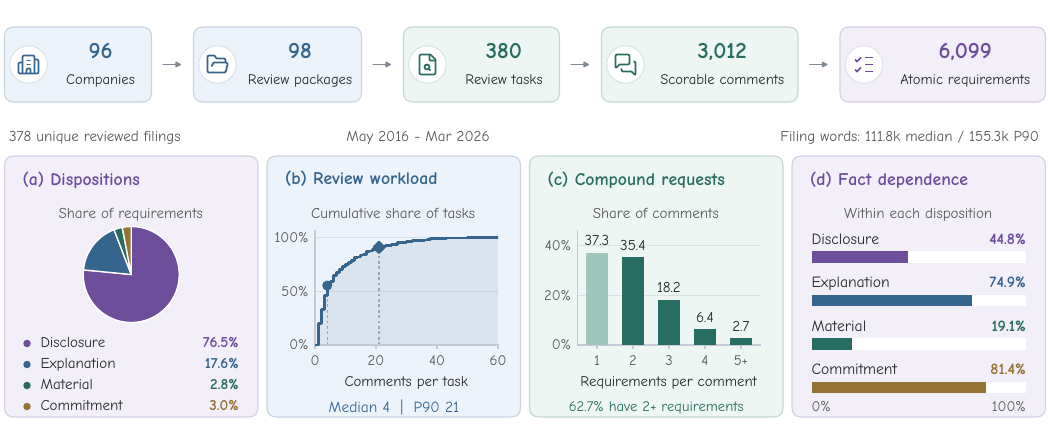}
  \caption{\textbf{Benchmark scope and task characteristics.}
  Top: corpus hierarchy and coverage.
  (a) Requirement disposition shares;
  (b) cumulative distribution of comments per task;
  (c) requirements per comment (5+ includes the full tail);
  (d) company-fact dependence within each disposition.}
  \Description{A scope diagram connects \BenchCompanies{} companies,
  \BenchPackages{} review packages, \BenchTasks{} review tasks,
  \BenchComments{} scorable comments, and \BenchRequirements{} requirement occurrences.
  The collection contains \BenchReviewedFilings{} unique reviewed filings, with
  review dates from \BenchReviewStart{} to \BenchReviewEnd{}.
  Below, four charts in one row: pie sectors show disposition shares,
  an empirical cumulative curve shows comments per task, bars show percentages
  of comments by requirement count, and horizontal bars show company-fact dependence.
  The fact-dependence bars use a zero-to-one-hundred-percent scale.
  Colors consistently identify the four requirement dispositions.}
  \label{fig:benchmark-statistics}
\end{figure*}

\section{Related Work}
\label{sec:related_work}

\subsection{Financial Disclosure Analysis}
\label{sec:rw_finance}

Corporate disclosure reduces information gaps between firms and investors. Healy and Palepu~\cite{healy2001information} review its relationship with information asymmetry and capital markets, while Johnston and Petacchi~\cite{johnston2017regulatory} examine financial reporting oversight through SEC comment letters. This literature motivates studying disclosure quality within the review process. Financial language models provide complementary computational capabilities: BloombergGPT~\cite{wu2023bloomberggpt} incorporates financial data into large-scale pretraining, and FinGPT~\cite{yang2023fingpt} develops an open framework for adapting language models to downstream financial applications.

Financial benchmarks examine several capabilities needed to interpret disclosures. FinQA~\cite{chen2021finqa} and TAT-QA~\cite{zhu2021tat} test numerical reasoning over financial text and tables; ConvFinQA~\cite{chen2022convfinqa} extends this setting to conversational reasoning. FinanceBench~\cite{islam2023financebench} emphasizes evidence-supported answers from company documents, whereas BizBench~\cite{krumdick2024bizbench} examines quantitative reasoning and program synthesis. FinBen~\cite{xie2024finben} broadens evaluation to extraction, generation, forecasting, and financial decision-making. ESGBenchmark~\cite{zhao2026advancing} further evaluates environmental, social, and governance (ESG) understanding and analysis based on corporate sustainability reports. These benchmarks measure complementary aspects of financial understanding, without making regulatory review and document revision their central evaluation loop. Extracting a reported number and assessing whether its explanation is adequate are different objectives. Reviewers must identify omissions before receiving explicit questions and obtain the facts needed to substantiate revisions.

Agentic financial research moves beyond isolated predictions. FinRobot~\cite{zhou2024finrobot} organizes specialized agents and financial toolchains, while Finance Agent Benchmark~\cite{bigeard2025finance} evaluates expert-authored research questions using search and SEC filings. Disclosure review adds a distinct requirement: agents must discover consequential omissions, request missing company facts, and substantiate corrections in the appropriate documents. FilingLoop studies this process through a benchmark reconstructed from historical SEC IPO review trajectories, with evaluation of issue discovery and requirement fulfillment in final filings and formal responses. Each reference concern states what must change and the evidence supporting it, so we can check the final filing and response against that request.

\subsection{Interactive Agent Benchmarks}
\label{sec:rw_interaction}

Interactive agents combine reasoning with actions that gather evidence or change an environment. ReAct~\cite{yao2022react} interleaves reasoning and tool use, while AgentBench~\cite{liu2024agentbench} and GAIA~\cite{mialon2024gaia} evaluate agents across environments and realistic assistant tasks. Effective evidence access remains a separate concern: Lost in the Middle~\cite{liu2024lost} demonstrates that access to a long context does not ensure reliable use of information throughout it. Review therefore requires reliable evidence access across filings and supporting documents.

Multi-agent frameworks structure collaboration through roles and communication. AutoGen~\cite{wu2023autogen} supports configurable agent conversations, CAMEL~\cite{li2023camel} studies cooperation through role-playing, and MetaGPT~\cite{hong2024metagpt} organizes specialized roles around standard operating procedures. Complementary benchmarks evaluate task completion in operational settings. WorkArena~\cite{drouin2024workarena} evaluates browser-mediated enterprise tasks, and TheAgentCompany~\cite{xu2026theagentcompany} simulates digital work with tools and coworkers. AppWorld~\cite{trivedi2024appworld} evaluates resulting application states and checks unintended changes. These settings assess completed tasks and the consequences of agent actions, extending evaluation beyond the quality of final replies.

Interaction itself can also determine success. $\tau$-bench~\cite{yao2024tau} combines user dialogue, domain policies, and tools with final-state evaluation; $\tau^2$-Bench~\cite{barres2025tau} extends coordination to a shared environment in which both participants act. FilingLoop builds on these evaluation principles through a regulatory reviewer and issuer counsel with distinct responsibilities and information access. Their exchanges produce revisions to a persistent filing, and an independent evaluator checks whether each disclosure requirement is fulfilled in the relevant deliverable, including checks for unsupported claims and harmful changes to otherwise adequate disclosures.

\subsection{Self-Evolving Agents}
\label{sec:rw_evolution}

Feedback can improve both an agent's current output and its behavior on later tasks. Self-Refine~\cite{madaan2023self} iteratively critiques and revises outputs, while Reflexion~\cite{shinn2023reflexion} stores verbal feedback to guide subsequent trials. For experience transfer, ExpeL~\cite{zhao2024expel} extracts insights from task trajectories, and Voyager~\cite{wang2023voyager} accumulates executable skills during embodied exploration. Agent Workflow Memory~\cite{wang2024agent} learns reusable routines for web navigation in offline and online settings. These approaches retain experience as language, code, or workflows without updating model weights. Applying experience transfer to professional review requires role-specific procedures learned from feedback on jointly produced documents.

Other work optimizes prompts and agent structures. Automatic Prompt Engineer~\cite{zhou2022large} generates and selects instructions, whereas OPRO~\cite{yang2024large} conditions new candidates on previous solutions and scores. ProTeGi~\cite{pryzant2023automatic} combines textual critiques with search, and TextGrad~\cite{yuksekgonul2024textgrad} propagates textual feedback through computation graphs. GEPA~\cite{agrawal2026gepa} uses trajectory reflection to evolve prompts and combine complementary improvements. At the system level, Automated Design of Agentic Systems~\cite{hu2025automated} searches over code-defined agents, while AFlow~\cite{zhang2025aflow} uses Monte Carlo tree search to construct agentic workflows. They edit prompts or agent design and use varied feedback to select which of the new versions to keep.

FilingLoop contributes a disclosure-review environment and an evidence-linked feedback protocol for updating both professional roles within a fixed workflow. Within-case review and revision are available to both the evolving and fixed-policy agents; only the evolving group carries learned guidance forward. Scores are locked before feedback is released, skills are frozen within each batch, and updates apply only to subsequent cases. The comparison evaluates a domain-specific application of procedural reflection against a matched fixed-policy control, using each task's first-attempt score; it does not compare alternative prompt-search algorithms.

\section{FilingLoop}
\label{sec:filingloop}

\begin{figure*}[t]
  \ifdefined\FilingLoopFullPaper\else\vspace{-0.3cm}\fi
  \centering
  \includegraphics[width=\textwidth]{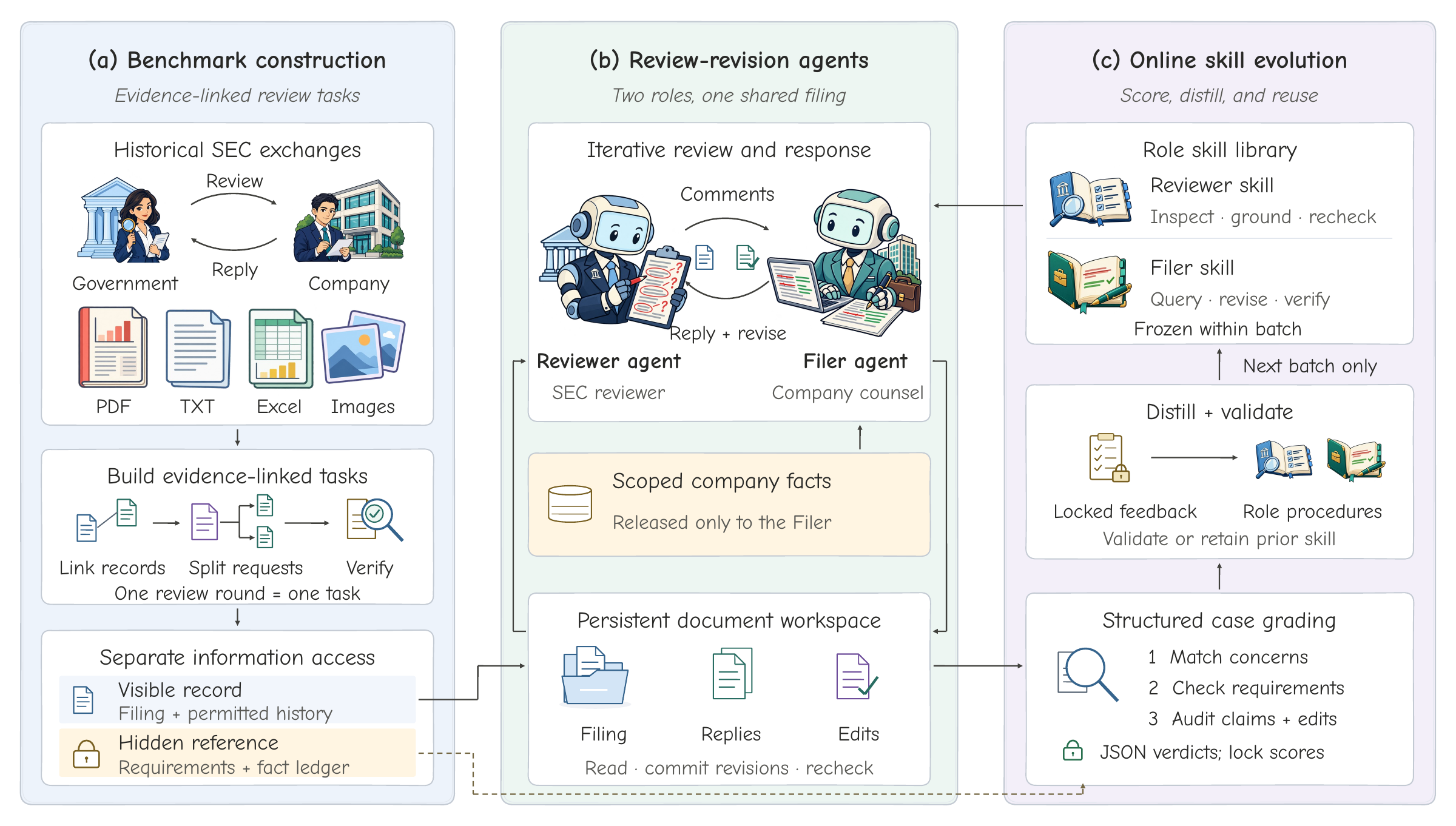}
  \caption{\textbf{FilingLoop benchmark and self-evolving agents.}
  (a) Historical SEC exchanges yield tasks with visible records and hidden references.
  (b) Reviewer and Filer agents iteratively review and revise disclosures.
  (c) Structured grading locks scores before feedback updates role skills
  for later batches, without replaying cases.}
  \Description{Three vertical nested panels show benchmark construction,
  two-role review and revision, and online skill evolution. Government and
  company cartoons identify the historical participants in the benchmark.
  PDF reports, text files, Excel workbooks and image cards show source materials.
  A navy-suited robot with a government building and red-annotated issue
  clipboard represents the regulatory Reviewer. A teal-suited robot with
  a company building, briefcase and laptop edits a disclosure with visible
  red and green revisions, representing the company-counsel Filer.
  Visible task records enter the
  document workspace, while an amber dashed path supplies hidden reference
  evidence directly to independent evaluation. Three case-level grading
  stages match comments, check requirements, and audit claims and edits.
  Structured JSON verdicts are validated before deterministic scoring.
  An upward feedback chain
  connects locked scores, procedural editing and role-specific skill storage.
  Blue checklist playbooks with magnifiers depict reusable Reviewer procedures;
  green editing playbooks with pens depict reusable Filer procedures. These
  icons appear both after distillation and in the role skill library.
  Only future batches receive updated guidance.}
  \label{fig:framework}
  \ifdefined\FilingLoopFullPaper\else\vspace{-0.3cm}\fi
\end{figure*}

FilingLoop couples an interactive disclosure benchmark with a framework
for improving the procedures of agents acting as Reviewer and Filer.
Figure~\ref{fig:framework} connects historical task construction, review
and revision within a case, and skill evolution across cases. This separation
is essential: both fixed and evolving agents can revise their work during
an interaction, but only the latter update their guidance for subsequent tasks.
We describe the benchmark and its evaluation criteria before presenting
the review and revision workflow and cross-case learning procedure.

% Active main framework figure; the previous illustration remains archived.

\subsection{FilingLoop Benchmark}
\label{sec:benchmark}

\subsubsection{Task Construction.}
The benchmark reconstructs disclosure-review workflows from the SEC's public EDGAR archive, primarily covering IPO registration filings with a small extension of annual reports from an existing issuer. Both domestic and foreign companies are included. Filings, amendments, staff letters, company responses, and exhibits retain their original identifiers and dates to preserve provenance across document versions.

Records are grouped by company and review package, with administrative correspondence distinguished from substantive responses. Within each package, dates, references, comment numbers, and source spans establish links between the reviewed filing, reviewer comments, company replies, and subsequent revisions. For annual reports, the reporting period and original accession must also be identified, as an SEC file number may cover multiple reporting years. Recurring concerns are tracked as issue threads across review rounds, while each occurrence retains its supporting evidence and applicable requirements. This organization allows the benchmark to trace how individual disclosure issues are raised, addressed, and revisited across successive review rounds.

Each task reconstructs one historical review round using a fixed local snapshot of permitted EDGAR records. Agents see the reviewed filing and earlier history; current historical comments, replies, and subsequent amendments remain hidden. Document visibility follows the reconstructed review history, which may differ from the dates on which correspondence became public. The hidden reference decomposes compound comments into atomic requirements, each linked to a disposition, target deliverable, supporting evidence, applicable comment IDs, and any required company facts. Construction combines model-assisted annotation, independent checking, reconciliation, and source validation. The project team verified the original collection. Annual-report records underwent additional annotation and audit passes with exact source checks. Model agreement is not treated as measured label accuracy.

Task eligibility is determined before agent execution. A scored comment must be identifiable from the permitted records and contain at least one applicable requirement supported by the available evidence. Requirements that depend on company facts are retained only when those facts can be attributed to evidence from the current round. Ambiguous source links and unsupported requirements are preserved in the audit trail rather than resolved through inference. Registration effectiveness is not considered evidence that a filing has no remaining deficiencies, and therefore does not serve as a basis for constructing negative tasks. Figure~\ref{fig:benchmark-statistics} summarizes the evaluable collection. Filing lengths are measured by whitespace-separated word counts in the primary filing for each task, excluding documents provided as prior-round history. Appendix~\ref{app:data-construction} details reconstruction, annotation, and eligibility checks.

\ifdefined\FilingLoopFullPaper
\noindent\textbf{Eligibility and the unit of annotation.}
The discovery pool contained 300 company candidates. Reconstruction yielded
412 review rounds, of which \BenchTasks{} are eligible; the other 32 remain
in exclusion records. A further 93 terminal diagnostics are excluded because
effectiveness or closure does not establish the absence of disclosure
deficiencies. Fact-evidence checks remove 318 requirement occurrences lacking
the necessary current-round support. The retained tasks therefore measure
resolution where the requested action and its evidentiary basis can be
reconstructed, rather than treating every archived exchange as scorable.

A requirement specifies an action and its expected deliverable, rather than
a string that must be copied from the historical response. Filing changes,
explanations to staff, supporting material, and commitments are checked in
their respective artifacts. A comment containing several such requests is
decomposed so that a supported explanation does not earn credit for an
unmade amendment. Requirements are attached to the rounds in which they
apply: a later follow-up does not retroactively add an obligation to an
earlier task. Repeated requests remain separate occurrences when they recur
in later rounds. Appendix~\ref{app:requirement-annotation} details annotation
and reconciliation, and Appendix~\ref{app:eligibility-and-coverage} records
the eligibility rules, excluded records, and final corpus coverage.
\fi

\subsubsection{Task Characteristics.}
The benchmark contains \BenchTasks{} review tasks from \BenchCompanies{}
companies and \BenchPackages{} review packages, with \BenchComments{}
scorable comments and \BenchRequirements{} applicable requirement
occurrences. These levels are distinct: one task can contain many comments,
and a recurring issue can contribute requirements in more than one round.
The \BenchRegistrationTasks{} registration tasks span
\BenchRegistrationPackages{} packages. Another \BenchAnnualTasks{} tasks
come from one issuer's annual-report review in 2026. This deliberately small
extension adds no companies and does not establish broad annual-report coverage.
Figure~\ref{fig:benchmark-statistics} reports distributions recomputed using
the same eligibility rules as the evaluation harness. All eligible tasks form
a single evaluation stream, with no separate training or validation split. The online
protocol below permits learning only from already scored tasks.

These statistics identify three demands on the agents. First, agents must
discover multiple issues within a filing: the median task contains
\BenchCommentsMedian{} scorable comments, with a 90th percentile of
\BenchCommentsPninety{}. Second, \BenchCompoundPct{} of comments contain
multiple requirements, so addressing one request need not resolve the whole
comment. Third, \BenchFactsPct{} of requirements are marked as needing
company facts. Disclosure changes account for \BenchDisclosurePct{} of requirements;
the remainder call for explanations to staff, supporting material, or
commitments. These are disposition types, not financial topic labels.
They motivate evaluation of discovery, evidence acquisition, and
verification in the required document. Section~\ref{sec:experiments} examines
how effectively agents meet these demands in the evaluated task stream.

\subsubsection{Outcome Evaluation.}
An evaluator separate from the acting Reviewer assesses frozen artifacts
using one grading model in three stages. \textbf{(1)~Matching} aligns generated
comments with historical concerns by substance and requested action; a grounded
additional concern is not automatically an error. \textbf{(2)~Fulfillment}
checks all requirements of matched scorable comments in one case-level request,
sharing the complete final filing across judgments. Each requirement receives
a binary verdict, an artifact identifier, a supporting quotation when available,
and a reason. Disclosure changes require evidence in the final filing;
a promise in a reply does not suffice. \textbf{(3)~Audit} jointly checks
extracted company assertions and all final-document changes against permitted
source evidence and released facts. Missing support alone is distinguished
from a contradiction.

Applicable evaluation stages return schema-constrained JSON. The controller
checks identifier coverage and quotation provenance before computing scores.
Invalid outputs are retried, and persistent failures remain incomplete
assessments. Case-level evaluation shares the final filing across comments
without resubmitting it. These checks establish structure and source
provenance; semantic validity still depends on interpreting the evidence
and requested disclosure changes.
Human--grader agreement and cross-grader sensitivity are not quantified.

Each assessment records discovery, requirement fulfillment, full comment
resolution, and harm. Discovery credits a reference concern when it matches
a generated comment. End-to-end fulfillment requires both discovery of the
concern and verification of the requested action in its final deliverable.
A comment is fully resolved only when all its requirements pass.
Undiscovered concerns and their requirements remain
in the respective denominators, so missed issues cannot inflate accuracy.

Harm records unfounded additional concerns, comments with contradicted
company assertions, and tasks containing harmful filing changes. These
counts supplement accuracy without being subtracted from it.
Section~\ref{sec:experimental-metrics} describes aggregate accuracies, role
diagnostics, completion, and resources; Appendix~\ref{app:evaluation-metrics}
gives their formal definitions and reporting scopes.

\subsection{Self-Evolving Agents}
\label{sec:evolving-agents}

\subsubsection{Review and Revision.}
The two roles use a common language model with distinct instructions,
outputs, and information access. The Reviewer
inspects permitted documents and raises grounded concerns with source
references. The Filer responds to these concerns, requests missing company
information when needed, and submits concrete revisions. Both roles use
document-reading and shell tools in isolated workspaces containing the allowed
record and navigation aids. Hidden references are held by the environment,
not copied into actor workspaces. Persistent documents and structured state
carry the interaction across turns rather than relying on an unrestricted
shared conversation history.

Revisions are submitted as source replacements linked to comment IDs.
The controller accepts a replacement only when its target can be located
uniquely, allowing whitespace normalization but rejecting ambiguous matches.
Accepted changes are committed to the authoritative draft. If a response also
requests facts, completed edits are preserved while the Filer finalizes the
pending work after receiving an answer. The fact service releases only
evidence-supported information linked to the relevant current-round comments.
Released facts enter the Filer's private history, while the Reviewer sees the
resulting formal responses and disclosures. An empty service answer indicates
limited evidence coverage, not that the real company possesses no information.

The Reviewer then reassesses the revised filing and responses. Unresolved
comments and newly raised concerns can initiate another exchange, with prior
responses and committed revisions preserved. Detailed rereview rationales are
logged for audit rather than forwarded to the next Filer turn.
Interaction ends when no active work remains or the exchange budget is
exhausted. This stopping decision controls execution only: final scoring
retains all eligible requirements, including concerns the Reviewer did not
raise or close.
The fixed-policy baseline shares this workflow, providing a matched control
for the joint skill updates applied to subsequent cases under identical base prompts.

\ifdefined\FilingLoopFullPaper
\subsubsection{Information and Artifact Control.}
Both roles can inspect the permitted record and current draft, but their
private conversations remain separate. Fact requests and released answers
enter the Filer's history. The Reviewer can use that information only when
the Filer includes it in a formal response or the revised disclosure.
The service withholds reference identifiers and original historical replies,
so acquiring facts does not reveal the hidden comment letter. Its answers
retain their status as company statements supported by the archived
evidence; the fact service does not independently certify the company's
underlying analyses or commitments.

The authoritative draft also remains separate from workspace files. Ordered
edit transactions identify the source text, replacement, and motivating
comment IDs; the host records whether each operation was committed.
Supporting-document changes additionally identify the authorized artifact
and its base hash. This distinction matters when an agent reports a change
that its edit failed to apply. After a fact request, finalization receives
the current draft and edit outcomes, retains committed work, and addresses
pending responses and rejected edits. It cannot initiate another fact
request within that phase. Appendix~\ref{app:tools-and-artifacts} gives the
submission contracts, and Appendix~\ref{app:fact-service} describes fact
release and finalization.
\fi

\subsubsection{Online Skill Evolution.}
We evaluate procedural adaptation through an ordered task stream, pooling
all eligible cases. A seeded company order and round-robin scheduling preserve
chronological order within each company; a batch contains at most one task
per company. Both experimental arms receive the same ordered batches, base
prompts, model, tools, and execution budgets, starting with empty learned
skills. Task concurrency controls scheduling within a batch independently
of the batch size. Skills remain frozen until every task in both arms has
completed execution and scoring. Thus, completion order cannot change the
guidance available to other tasks running concurrently in that batch.

After first-attempt results are locked, a skill editor receives the evolving
arm's completed-batch feedback: discovery outcomes, applicable requirements,
fulfillment decisions, evidence quotations, and reasons. It proposes reusable
procedural guidance separately for the Reviewer and Filer, or explicitly
retains the current guidance. Let $s_b^r$ denote the frozen skill used by
role $r$ (Reviewer or Filer) in batch $b$, and $\mathcal F_b$ the evolving
arm's locked batch feedback. The update is
\begin{equation}
 s_{b+1}^{r}=\operatorname{ValidateOrKeep}
 \bigl(\operatorname{Edit}(s_b^{r},\mathcal F_b),s_b^{r}\bigr).
\label{eq:skill-update}
\end{equation}
Validation checks structure, role identity, company identifiers, and direct
copying of reference text. Invalid proposals can be revised using validation
feedback; if no valid update is produced, the incumbent skill is retained.
Guidance length is advisory; accepted updates are not truncated. Validation enforces
content and formatting constraints without selecting updates by held-out
performance.

Only later, unattempted tasks use the new guidance. Model weights and base
instructions remain fixed, completed cases retain their original scores,
and skill updates stop after the penultimate batch. We record skill
versions, update decisions, task outcomes, elapsed time, and model usage,
including editing costs. This is online adaptation with historical scoring
feedback, not evaluation on an independent held-out test set. Earlier rounds
of a company may inform later rounds; unseen-company generalization requires
a separate evaluation protocol.

\ifdefined\FilingLoopFullPaper
\noindent\textbf{What transfers between tasks.}
Each skill is a Markdown document with role metadata and procedural
instructions. Reviewer guidance can specify how to inspect an explanation
for missing evidence; Filer guidance can specify how to verify a committed
change against every request in a compound comment. These are procedures
that the representation can express, rather than new facts about the next
issuer. Fixed role instructions take precedence, and any assertion about
the current issuer still requires support from its permitted records or
company facts that the service releases for that issuer's current review round.

The editor can return \texttt{no\_change} or a complete replacement skill.
It receives validation errors for invalid proposals and has at most three
attempts per role. Known company identifiers, private source paths, and
sufficiently long exact reference copies are rejected. These textual checks
constrain what is retained without establishing that every paraphrased case
detail has been removed. Incompletely assessed batches produce no update.
Policy hashes bind each subsequent task to the guidance frozen before it
starts, allowing the recorded outcome to be related to the procedure
actually available during execution. Appendix~\ref{app:agent-evolution}
provides the skill format, validation rules, and update schedule.
\fi

\section{Empirical Studies}
\label{sec:experiments}

\ifdefined\FilingLoopFullPaper
  % Generated from the current main table; no model calls.
\newcommand{\AGFixedUndiscoveredReq}{3,937}
\newcommand{\AGFixedOpenReq}{681}
\newcommand{\AGFixedMetReq}{1,481}
\newcommand{\AGFixedUndiscoveredComment}{2,029}
\newcommand{\AGFixedOpenComment}{474}
\newcommand{\AGFixedMetComment}{509}
\newcommand{\AGFixedSharedCost}{495.34}
\newcommand{\AGFixedSharedPct}{20.8}
\newcommand{\AGFixedSharedInput}{209.02}
\newcommand{\AGFixedSharedOutput}{3.65}
\newcommand{\AGFixedMissedShareReq}{85.3}
\newcommand{\AGFixedMissedShareComment}{81.1}
\newcommand{\AGEvoUndiscoveredReq}{3,851}
\newcommand{\AGEvoOpenReq}{543}
\newcommand{\AGEvoMetReq}{1,705}
\newcommand{\AGEvoUndiscoveredComment}{1,990}
\newcommand{\AGEvoOpenComment}{397}
\newcommand{\AGEvoMetComment}{625}
\newcommand{\AGEvoSharedCost}{539.92}
\newcommand{\AGEvoSharedPct}{21.4}
\newcommand{\AGEvoSharedInput}{229.92}
\newcommand{\AGEvoSharedOutput}{4.19}
\newcommand{\AGEvoMissedShareReq}{87.6}
\newcommand{\AGEvoMissedShareComment}{83.4}
\newcommand{\AGExtraCost}{139.08}
\newcommand{\AGExtraHours}{14.32}
\newcommand{\AGMarginalReqCost}{0.62}
\newcommand{\AGMarginalCommentCost}{1.20}
\newcommand{\AGMarginalReqMinutes}{3.84}
\newcommand{\AGFixedCompletedUnresolved}{364}
\newcommand{\AGEvoCompletedUnresolved}{360}
\newcommand{\AGExtraReviewerCost}{41.20}
\newcommand{\AGExtraFilerCost}{53.30}
\newcommand{\AGExtraSharedCost}{44.58}

\fi

We evaluate discovery, fulfillment, resource use, and harmful revisions
under fixed and evolving policies. Our study addresses four research questions:
\begin{itemize}[leftmargin=0pt, nosep]
  \item \textbf{RQ1:} How effectively do FilingLoop and existing agents discover
  and resolve disclosure deficiencies across backbones?
  \item \textbf{RQ2:} How does cross-task procedural evolution affect
  end-to-end resolution and role-level performance?
  \item \textbf{RQ3:} What are the trade-offs between resolution quality, token
  usage, monetary cost, and execution time?
  \item \textbf{RQ4:} How often do filing revisions introduce harm, and how
  completely are they audited?
\end{itemize}

% RQ analyses and Figure 4 use the current main table as their numeric source.

\subsection{Experimental Setup}
\label{sec:experimental-setup}

\noindent\textbf{Data and task stream.}
The evaluation uses all \BenchTasks{} eligible review tasks from
\BenchCompanies{} companies in the benchmark described in
Section~\ref{sec:benchmark}. These comprise \BenchRegistrationTasks{}
registration-review tasks and \BenchAnnualTasks{} annual-report tasks,
pooled without a training or validation split. All configurations receive
the same seeded company order, chronological rounds within each company,
and permitted records. Current historical comments, subsequent replies,
and amendments remain hidden from the acting agents throughout each task.

\noindent\textbf{Online protocol.}
Each task is attempted once per stream. Both FilingLoop variants start
with empty learned skills and identical base instructions. The main
comparison uses order seed~0 and batches of at most 16 tasks,
with no repeated company within a batch. Skills remain frozen until both
variants finish scoring the batch. The evolving variant then updates its
guidance using its own locked feedback, exclusively for subsequent tasks.
Scores are never replaced by replay results; earlier company rounds may
nevertheless inform later rounds under this online adaptation protocol.
The reported differences describe this seeded stream and are limited to the
evaluated task sample and order.

\noindent\textbf{Execution and assessment.}
The reference configuration permits up to three review--revision exchanges
and 100 agent steps per role invocation, including follow-up phases.
A step is one model response and its resulting tool calls.
We run up to \MRTaskConcurrency{} tasks concurrently and record resource
usage separately for each role invocation.
A single \textit{gpt-6.1-sol} grader
assesses frozen artifacts through the matching, fulfillment, and audit stages
in Section~\ref{sec:benchmark}. The same model performs fact matching and
skill editing, with grading-policy identifiers retained for each assessment.
The evaluator was manually reviewed; all configurations use the same grading
model, prompts, and artifact checks.
Acting roles receive no
current-task grading feedback. Manifests record model identifiers, inference
settings, software versions, data fingerprints, and policy versions.
We use \texttt{high} as the default LLM reasoning effort; inference
options are retained in the execution manifests.
Because step definitions differ across frameworks, we report measured
token usage, cost, and execution time alongside nominal step budgets.
Appendix~\ref{app:experimental-details} gives construction, information
boundaries, and reproduction details.

\subsection{Evaluation Metrics}
\label{sec:experimental-metrics}

\ifdefined\FilingLoopFullPaper
% Extended metric exposition; the submission retains its compact definitions.
\noindent\textbf{Requirement and comment resolution.}
For task $t$, let $\mathcal C_t$ be the scorable reference comments and
$\mathcal U_{ti}$ the applicable requirements of comment $i$. Let
$d_{ti}\in\{0,1\}$ indicate discovery of that comment and
$b_{tij}\in\{0,1\}$ indicate fulfillment of requirement $j$ in its requested
deliverable. Over the evaluated task set $\mathcal T$, system accuracies are
\begin{align}
\mathrm{Acc}_{\rm req}
&=\frac{\sum_{t\in\mathcal T}\sum_{i\in\mathcal C_t}
\sum_{j\in\mathcal U_{ti}}d_{ti}b_{tij}}
{\sum_{t\in\mathcal T}\sum_{i\in\mathcal C_t}|\mathcal U_{ti}|},
\label{eq:full-system-req}\\
\mathrm{Acc}_{\rm comment}
&=\frac{\sum_{t\in\mathcal T}\sum_{i\in\mathcal C_t}
d_{ti}\prod_{j\in\mathcal U_{ti}}b_{tij}}
{\sum_{t\in\mathcal T}|\mathcal C_t|}.
\label{eq:full-system-comment}
\end{align}
All accuracies are reported as percentages. Requirement Accuracy gives equal
weight to annotated requirement occurrences, while Comment Accuracy gives
equal weight to comments and requires every constituent obligation to pass.
The denominators are fixed before execution. Missing a concern therefore
earns no system credit without removing the concern or its requirements.
Requirements can share evidence or prerequisites; their count represents
annotated obligations rather than independent causes of failure.

\noindent\textbf{Task and company resolution.}
Task Accuracy is the fraction of tasks for which assessment is complete and
every applicable requirement passes. Company Accuracy requires the same
condition across all review rounds of an issuer. These stricter measures
withhold a pass when even one obligation remains unmet. Task completion
instead measures whether execution and assessment finish, independently of
resolution quality. Failed or unassessed tasks receive no system credit,
and their known reference targets remain in the system denominators.

\noindent\textbf{Reviewer and Filer diagnostics.}
Reviewer scores measure discovery over the full reference inventory.
Requirement-level discovery credit inherits the parent-comment match; it
does not imply that the Reviewer separately articulated every atomic
requirement. Post-hoc Filer scores compare the two variants on Fixed's
assigned reference-comment IDs: \MRDiscoveredUnits{} requirements from
\MRcFRDenominator{} comments, covering \MRFilerTaskAccDenominator{} tasks
and \MRFilerCompanyAccDenominator{} companies. Both Filer numerators retain
only fulfillment linked to those IDs. Extra Evolving discoveries contribute
to system scores, while tasks and companies with no assigned targets are
excluded from the conditional Filer accuracies.

This common pool fixes the scope of the diagnostic, rather than the comments
shown to the agents. Each Filer still receives its own Reviewer's requests
and retains the interaction that produced its final artifacts. Consequently,
the paired Filer scores describe fulfillment under the joint policies;
they do not isolate the effect of updating Filer alone. Reviewer and Filer
scores describe different stages and are not additive contributions to
system accuracy.

\noindent\textbf{Harm and audit coverage.}
The evaluator separately checks unfounded additional concerns, contradicted
company assertions, and harmful filing changes. A harmful revision introduces
unsupported claims or contradictions, or damages supported disclosure;
leaving an original request unresolved is not itself harm. The task-level
harm rate counts audited tasks containing at least one such change, whereas
edit-audit coverage counts audited tasks among all scheduled tasks. Unaudited
tasks are excluded from the harm-rate denominator rather than treated as
harm-free. These measures supplement fulfillment without being subtracted
from the accuracy scores.

\noindent\textbf{Resources and efficiency.}
We report input/output tokens, API cost, cumulative Agent hours, and mean
minutes per scheduled task. Input includes cache reads and writes; Output
includes reasoning tokens. Cost applies the saved price schedule to recorded
usage for role execution and shared services, including assessment, fact
matching, and skill editing. It is a usage-based total rather than a
reconciled invoice amount. Shared services remain in the system total, so
the displayed role costs need not account for all expenditure needed to
produce and assess the final artifacts.

Agent hours sum recorded role-invocation durations, including model calls,
tools, and compaction, while excluding independent assessment and gaps
between stopping and restarting. Concurrent tasks each contribute their
own duration. Mean minutes per task uses all \BenchTasks{} scheduled tasks;
missing duration segments are not imputed. Cost per fulfilled requirement
divides total system cost by the fulfilled requirement count, and throughput
uses either that cost or cumulative Agent time. These ratios describe
resource use per observed outcome, rather than elapsed wall-clock speed.
Appendix~\ref{app:evaluation-metrics} gives the complete role-scoped formulas,
failure handling, and accounting conventions.

\else
\textbf{Requirement Accuracy} measures the fraction of reference requirements
fulfilled, pooling checks across tasks. \textbf{Comment Accuracy} measures
the fraction of comments fully resolved, with each comment weighted equally
regardless of its requirement count. System denominators retain all reference
items, including undiscovered issues. \textbf{Task Accuracy} and
\textbf{Company Accuracy} measure the fractions of tasks and companies
satisfying every applicable requirement, respectively, with company success
covering all review rounds. Partial resolution contributes requirement-level
credit but does not pass an entire comment, task, or company.
\textbf{Task completion} measures execution and assessment completion over
all scheduled tasks, independently of resolution quality. Failed or
unassessed tasks receive no system credit, and known targets remain in the
denominators. Reviewer scores measure discovery. Post-hoc Filer scores use
the same \MRDiscoveredUnits{} requirements from \MRcFRDenominator{} comments
assigned under Fixed: both numerators count only fulfillment linked to these
reference-comment IDs. Extra Evolving discoveries contribute to system
scores. Tasks and companies without assigned targets are excluded from Filer
accuracy. Role scores describe separate stages and are not additive.
Unsupported claims and harmful revisions are audited separately.
For efficiency, we report \textbf{input/output tokens}, \textbf{API cost},
\textbf{cumulative Agent hours}, and \textbf{mean minutes per task}.
Token totals include cache usage and reasoning tokens.
Cost uses recorded usage and saved prices, including shared evaluation and
skill-editing services; Agent time sums role-invocation durations rather
than elapsed wall-clock time.
Appendix~\ref{app:evaluation-metrics} provides formal definitions, eligible
denominators, failure handling, and detailed resource accounting.
\fi

% Keep the main comparison table near the results, outside the metric appendix.
\begin{table*}[t]
  \caption{End-to-end results and role diagnostics on 380 tasks from
  96 companies. Parentheses show successful/eligible counts;
  Filer rows share the Fixed target pool.}
  \label{tab:main-results}
  \centering
  \footnotesize
  \setlength{\tabcolsep}{2pt}
  \renewcommand{\arraystretch}{1.12}
  \resizebox{\textwidth}{!}{%
  \begin{tabular}{@{}*{12}{l}@{}}
    \toprule
    & & \multicolumn{4}{l}{Accuracy (\%) $\uparrow$}
    & Task completion & \multicolumn{2}{l}{Total tokens (M)} & Cost & \multicolumn{2}{l}{Agent time} \\
    \cmidrule(lr){3-6}\cmidrule(lr){7-7}\cmidrule(lr){8-9}\cmidrule(lr){10-10}\cmidrule(l){11-12}
    Method & Backbone & Requirement Acc. & Comment Acc. & Task Acc. & Company Acc. & (\%) $\uparrow$ & Input & Output & (USD) $\downarrow$ & Total (h) & Avg.\ (min/task) \\
    \midrule
    Codex & \textit{gpt-6.1-sol}
      & 22.13\,(1,350/6,099)
      & 14.84\,(447/3,012)
      & 4.74\,(18/380)
      & 0.00\,(0/96)
      & 99.74\,(379/380)
      & 4,706.25 & 33.07 & 3,498.82 & 673.46 & 106.34 \\
    Claude Code & \textit{claude-opus-5.5}
      & 21.18\,(1,292/6,099)
      & 14.21\,(428/3,012)
      & 4.21\,(16/380)
      & 0.00\,(0/96)
      & 99.47\,(378/380)
      & 4,301.81 & 38.14 & 3,713.03 & 696.21 & 109.93 \\
    mini-SWE-agent & \textit{gpt-6.1-sol}
      & 21.99\,(1,341/6,099)
      & 14.61\,(440/3,012)
      & 4.47\,(17/380)
      & 0.00\,(0/96)
      & 99.74\,(379/380)
      & 3,676.76 & 24.14 & 1,353.33 & 183.61 & 28.99 \\
    \rowcolor{MRPanel}
    \textbf{mini-SWE-agent} & \textit{claude-opus-5.5}
      & 23.74\,(1,448/6,099)
      & 16.37\,(493/3,012)
      & 5.26\,(20/380)
      & 0.00\,(0/96)
      & 100.00\,(380/380)
      & 3,382.62 & 28.73 & 2,327.73 & 212.99 & 33.63 \\
    mini-SWE-agent & \textit{gemini-4-argon}
      & 21.10\,(1,287/6,099)
      & 14.31\,(431/3,012)
      & 4.21\,(16/380)
      & 0.00\,(0/96)
      & 99.47\,(378/380)
      & 3,934.13 & 26.07 & 1,028.53 & 172.59 & 27.25 \\
    mini-SWE-agent & \textit{deepseek-v4.1-flash}
      & 19.56\,(1,193/6,099)
      & 12.55\,(378/3,012)
      & 3.42\,(13/380)
      & 0.00\,(0/96)
      & 99.21\,(377/380)
      & 3,051.71 & 21.97 & 392.47 & 143.22 & 22.61 \\
    \addlinespace[3pt]
    \midrule
    \addlinespace[3pt]
    \rowcolor{MRPanel}
    \textbf{FilingLoop--Fixed} & \textit{gpt-6.1-sol}
      & 24.28\,(1,481/6,099)
      & 16.90\,(509/3,012)
      & 3.95\,(15/380)
      & 1.04\,(1/96)
      & 99.74\,(379/380)
      & 6,098.59 & 35.59 & 2,380.15 & 455.04 & 71.85 \\
    \quad \textcolor{MRFound}{Reviewer} & \textit{gpt-6.1-sol}
      & 35.45\,(2,162/6,099)
      & 32.64\,(983/3,012)
      & 13.16\,(50/380)
      & 1.04\,(1/96)
      & 100.00\,(380/380)
      & 3,041.21 & 15.47 & 1,144.51 & 228.26 & 36.04 \\
    \quad \textcolor{MRMet}{Filer} & \textit{gpt-6.1-sol}
      & 68.50\,(1,481/2,162)
      & 51.78\,(509/983)
      & 27.21\,(77/283)
      & 5.26\,(5/95)
      & 99.74\,(379/380)
      & 2,848.36 & 16.47 & 740.30 & 226.78 & 35.81 \\
    \addlinespace[3pt]
    \rowcolor{MRPanel}
    \textbf{FilingLoop--Evolving} & \textit{gpt-6.1-sol}
      & 27.96\,(1,705/6,099)
      & 20.75\,(625/3,012)
      & 5.00\,(19/380)
      & 2.08\,(2/96)
      & 99.74\,(379/380)
      & 6,432.51 & 37.90 & 2,519.23 & 469.36 & 74.11 \\
    \quad \textcolor{MRFound}{Reviewer} & \textit{gpt-6.1-sol}
      & 36.86\,(2,248/6,099)
      & 33.93\,(1,022/3,012)
      & 14.21\,(54/380)
      & 2.08\,(2/96)
      & 100.00\,(380/380)
      & 3,126.36 & 16.17 & 1,185.71 & 232.83 & 36.76 \\
    \quad \textcolor{MRMet}{Filer} & \textit{gpt-6.1-sol}
      & 78.86\,(1,705/2,162)
      & 63.58\,(625/983)
      & 32.16\,(91/283)
      & 6.32\,(6/95)
      & 99.74\,(379/380)
      & 3,076.23 & 17.54 & 793.60 & 236.53 & 37.35 \\
    \bottomrule
  \end{tabular}}
\end{table*}

\subsection{Baselines}
\label{sec:baselines}

We compare three external agent frameworks and two FilingLoop variants.
\textbf{Codex}, paired with \textit{gpt-6.1-sol}, supports non-interactive execution
and structured event logging.\footnote{\url{https://developers.openai.com/codex/noninteractive}}
\textbf{Claude Code}, paired with \textit{claude-opus-5.5}, provides file-reading,
editing, and shell tools within its agent loop.\footnote{\url{https://code.claude.com/docs/en/headless}}
\textbf{mini-SWE-agent} provides a lightweight shell-based scaffold and is
paired with \textit{gpt-6.1-sol}, \textit{claude-opus-5.5},
\textit{gemini-4-argon}, and \textit{deepseek-v4.1-flash}.\footnote{\url{https://mini-swe-agent.com/latest/}}
Both FilingLoop variants use \textit{gpt-6.1-sol}:
\textbf{FilingLoop--Fixed} retains the review--revision workflow without skill
updates, while \textbf{FilingLoop--Evolving} also learns role-specific guidance
from locked historical feedback. These pairings yield eight comparison
configurations, with the same backbone assigned to both roles within each
configuration. External adapters preserve separate
role sessions, equal access to permitted records and scoped facts, and the
common artifact-submission contract, while retaining each framework's
native execution loop. Fixed configurations carry no learned guidance
between tasks. The \textit{gpt-6.1-sol} configurations control model choice
when comparing frameworks, while the four mini-SWE-agent configurations
compare backbones within a common scaffold. The FilingLoop pair compares
fixed guidance with cross-task procedural evolution.

\subsection{Main Results (RQ1)}
\label{sec:main-results}

\begin{figure*}[t]
  \centering
  \includegraphics[width=\textwidth]{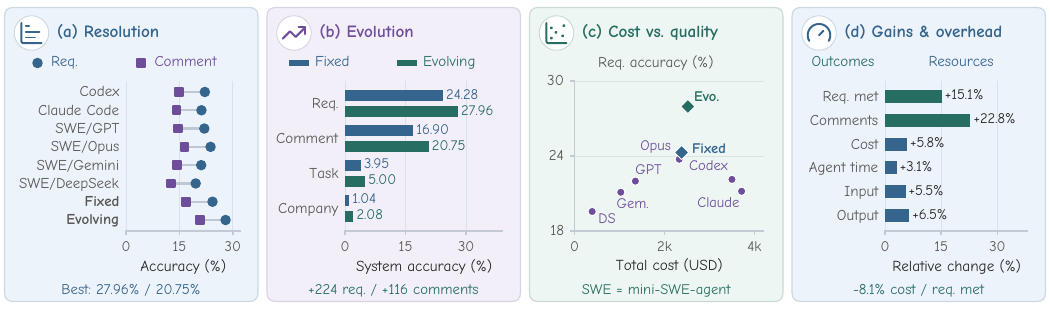}
  \caption{\textbf{Resolution quality and efficiency.}
  (a) Requirement and Comment Accuracy; (b) Fixed--Evolving comparison;
  (c) cost versus Requirement Accuracy; (d) relative gains and resource overhead.
  Results derive from Table~\ref{tab:main-results}.}
  \Description{Four colored panels in one horizontal row compare the eight configurations in the main
  table. Evolving achieves the highest requirement and comment accuracies,
  27.96 and 20.75 percent. Relative to Fixed it fulfills 224 more requirements
  and fully resolves 116 more comments. The cost plot places Evolving above
  Fixed in accuracy with a modest increase in cost. Relative gains of 15.1
  percent in fulfilled requirements and 22.8 percent in resolved comments
  accompany increases of 5.8 percent in cost and 3.1 percent in recorded Agent
  time. Cost per fulfilled requirement falls by 8.1 percent.}
  \label{fig:rq-analysis}
\end{figure*}

\noindent\textbf{End-to-end resolution.}
Table~\ref{tab:main-results} compares eight configurations across four
levels of disclosure resolution. FilingLoop--Evolving achieves the highest
Requirement and Comment Accuracies, \RQEvoReq{}\% and \RQEvoComment{}\%,
fulfilling \RQEvoReqCount{} requirements and fully resolving
\RQEvoCommentCount{} comments. Relative to the strongest external baseline,
mini-SWE-agent with \textit{claude-opus-5.5}, these are gains of
\RQBestExternalReqGain{} and \RQBestExternalCommentGain{} percentage points
(pp), respectively. Figure~\ref{fig:rq-analysis}(a) compares fulfillment of
individual requirements with resolution of entire comments, which requires
all constituent checks to pass.

\noindent\textbf{Framework and backbone comparisons.}
With the same \textit{gpt-6.1-sol} backbone, FilingLoop--Fixed reaches
\RQFixedReq{}\% Requirement Accuracy, exceeding Codex by
\RQFixedCodexReqGain{} pp and mini-SWE-agent by \RQFixedSweReqGain{} pp.
The fixed review--revision workflow thus already resolves more reference
requirements in the reported comparison, before procedural updates.
Within mini-SWE-agent, changing the backbone yields Requirement Accuracy
from \RQDeepSeekReq{}\% (DeepSeek) to \RQOpusReq{}\% (Opus), showing that
performance depends on both the model and its execution framework under the
shared task and scoring protocol.

\noindent\textbf{Execution completion versus substantive success.}
All configurations complete at least \RQMinCompletion{}\% of scheduled
tasks, yet none exceeds \RQOpusTask{}\% Task Accuracy. The highest Task Accuracy
is achieved by mini-SWE-agent/Opus with \RQOpusTaskCount{} fully successful tasks,
compared with \RQEvoTaskCount{} for Evolving. Evolving leads at company
level, but only \RQEvoCompanyCount{} of \BenchCompanies{} companies pass
all requirements across their review rounds; every external configuration
has zero company-level passes. These stricter metrics distinguish partial
improvement from complete resolution of all obligations in a review round
or company trajectory.
Appendix~\ref{app:case-studies} provides artifact-linked success and
compound-request failure cases with the relevant filing passages and responses.

\noindent\textbf{Findings (RQ1).}
\textbf{(1)~Fixed's advantage is present before evolution.}
Fixed exceeds the other \textit{gpt-6.1-sol} configurations in Requirement
Accuracy before evolution. This compares complete frameworks without
isolating role prompts, tools, or document management.
\textbf{(2)~The task-level ranking differs from the pooled scores.}
Evolving leads in Requirement Accuracy but not Task Accuracy. Every requested
action must pass in its target filing or formal response before a concern
is fully resolved.
\textbf{(3)~Unresolved obligations persist across issuer rounds.}
Outstanding obligations span review rounds; recurring concerns and related
requirements are not independent observations.

\subsection{Effect of Procedural Evolution (RQ2)}
\label{sec:ablations}
\label{sec:online-evolution-results}

\begin{figure}[t]
  \centering
  \includegraphics[width=\columnwidth]{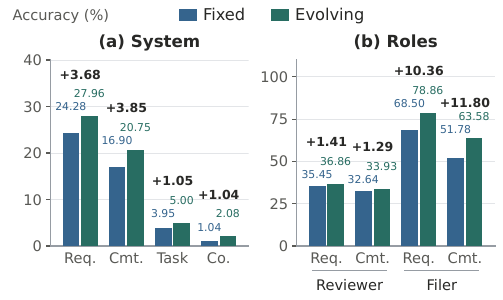}
  \caption{\textbf{Effect of evolution.}
  (a) System; (b) role diagnostics. Bars show accuracies (\%);
  + labels show gains (pp).
  Filer uses the shared Fixed target pool.}
  \Description{Two side-by-side white-background panels preserve all eight
  comparisons in the evolution table. Paired vertical bars show Fixed and
  Evolving accuracies, with exact values above each bar and percentage-point
  gains marked with plus signs above each pair.
  System Requirement: 24.28 to 27.96, gain 3.68; Comment: 16.90 to 20.75,
  gain 3.85; Task: 3.95 to 5.00, gain 1.05; Company: 1.04 to 2.08, gain 1.04.
  Reviewer Requirement: 35.45 to 36.86, gain 1.41; Comment: 32.64 to 33.93,
  gain 1.29. Conditional Filer Requirement: 68.50 to 78.86, gain 10.36;
  Comment: 51.78 to 63.58, gain 11.80.
  Both axes start at zero; system and role panels have different ranges.}
  \label{fig:evolution-effects}
\end{figure}

\noindent\textbf{Gains across evaluation levels.}
Figure~\ref{fig:evolution-effects}(a) compares Fixed and Evolving across four
system metrics under the online protocol, where
updates affect only subsequent tasks.
Requirement Accuracy rises from \RQFixedReq{}\% to \RQEvoReq{}\%
(+\RQReqGain{} pp), and Comment Accuracy from \RQFixedComment{}\% to
\RQEvoComment{}\% (+\RQCommentGain{} pp). This corresponds to
\RQMoreReq{} additional fulfilled requirements and \RQMoreComment{}
additional fully resolved comments over the same reference inventory.
Task Accuracy increases from \RQFixedTask{}\% to \RQEvoTask{}\%, while
Company Accuracy increases from \RQFixedCompany{}\% to
\RQEvoCompany{}\%. These are observed gains in this stream;
company-level success changes from one issuer to two.

\noindent\textbf{Discovery and conditional revision.}
Figure~\ref{fig:evolution-effects}(b) separates discovery from conditional
revision. Reviewer Requirement and Comment Accuracies improve by
\RQReviewerReqGain{} and \RQReviewerCommentGain{} pp, respectively.
On the shared Fixed target pool, Filer Requirement Accuracy increases by
\RQFilerReqGain{} pp to \RQFilerReq{}\%, and Comment Accuracy increases by
\RQFilerCommentGain{} pp to \RQFilerComment{}\%.
The pool controls scoring scope; each Filer still responds to its own
Reviewer. These are conditional diagnostics of the joint policy.

\noindent\textbf{Remaining coverage gap.}
Even with evolution, Reviewer covers only \RQReviewerReq{}\% of reference
requirements and \RQReviewerComment{}\% of reference comments. Filer's
higher conditional accuracy applies only to assigned concerns; it cannot
compensate for undiscovered deficiencies. Evolution improves resolution
across the task stream, with further gains constrained by issue discovery.

\ifdefined\FilingLoopFullPaper
\noindent\textbf{Accounting for the residual errors.}
Under Evolving, \AGEvoUndiscoveredReq{} requirements remain undiscovered and
\AGEvoOpenReq{} are discovered but unfulfilled. The undiscovered group
accounts for \AGEvoMissedShareReq{}\% of all remaining unfulfilled
requirements. This decomposition uses each variant's own discovery set;
it is distinct from the shared Fixed pool used for paired Filer accuracy.
Relative to Fixed, discovery rises by 86 requirements while the discovered
but unfulfilled count falls by 138, giving the net gain of $86+138=224$.
The corresponding comment-level identity is $39+77=116$. These aggregate
identities locate the remaining workload without identifying item-level
transitions or assigning causal effects to either role.
Appendix~\ref{app:role-diagnostics} gives the complete residual accounting.
\fi

\noindent\textbf{Findings (RQ2).}
\textbf{(1)~Joint updates improve the recorded scores.}
Evolving increases all four system accuracies with fixed model weights and
first-pass scoring. Attributing gains requires role-specific update ablations.
Appendix~\ref{app:skill-transfer-cases} traces a learned verification
procedure into a later filing.
\textbf{(2)~Role gains use different target sets.}
Reviewer covers all references; Filer conditions on Fixed's targets. Their
percentage-point changes are not additive contributions to system gain.
\textbf{(3)~Undiscovered concerns cap system accuracy.}
Even perfect revision of Evolving's discovered set caps system Requirement
Accuracy at \RQReviewerReq{}\%. Expanding discovery is necessary to resolve
additional reference obligations.

\subsection{Resource Efficiency (RQ3)}
\label{sec:resource-efficiency}

% Generated from tables/main_results.tex; ratios use its displayed resource totals.
\begin{table}[t]
  \ifdefined\FilingLoopFullPaper\else\captionsetup{skip=4pt}\fi
  \caption{Resource efficiency over 380 tasks. Throughput counts fulfilled
  requirements; hours denote recorded Agent time.
  SWE = mini-SWE-agent; backbones follow Table~\ref{tab:main-results}.}
  \label{tab:resource-efficiency}
  \centering\footnotesize
  \setlength{\tabcolsep}{3pt}
  \renewcommand{\arraystretch}{1.10}
  \begin{tabular*}{\columnwidth}{@{\extracolsep{\fill}}lrrr@{}}
    \toprule
    & Cost/task & \multicolumn{2}{c}{Requirements met} \\
    \cmidrule(l){3-4}
    Configuration & (USD) $\downarrow$ & /\$100 $\uparrow$ & /hour $\uparrow$ \\
    \midrule
    Codex / GPT & 9.21 & 38.6 & 2.00 \\
    Claude Code / Opus & 9.77 & 34.8 & 1.86 \\
    SWE / GPT & 3.56 & 99.1 & 7.30 \\
    SWE / Opus & 6.13 & 62.2 & 6.80 \\
    SWE / Gemini & 2.71 & 125.1 & 7.46 \\
    SWE / DeepSeek & 1.03 & 304.0 & 8.33 \\
    \midrule
    FilingLoop--Fixed & 6.26 & 62.2 & 3.25 \\
    \textbf{FilingLoop--Evolving} & 6.63 & 67.7 & 3.63 \\
    \bottomrule
  \end{tabular*}
  \ifdefined\FilingLoopFullPaper\else\vspace{-3pt}\fi
\end{table}

\noindent\textbf{Incremental cost of evolution.}
Evolving costs \$\RQEvoCost{} across \BenchTasks{} tasks, compared with
\$\RQFixedCost{} for Fixed, an increase of \RQCostGrowth{}\%.
Recorded Agent time increases by \RQTimeGrowth{}\%, from
\RQFixedHours{} to \RQEvoHours{} hours, or \RQExtraMinutes{} additional
minutes per task. Input and output tokens increase by \RQInputGrowth{}\%
and \RQOutputGrowth{}\%, respectively. In comparison, the numbers of
fulfilled requirements and fully resolved comments grow by
\RQReqGrowth{}\% and \RQCommentGrowth{}\%
(Figure~\ref{fig:rq-analysis}(d)). Token and cost totals include skill editing and subsequent execution under
the updated policies; Agent time measures recorded role invocations.

\noindent\textbf{Resolution per unit of expenditure.}
The larger outcome gains reduce cost per fulfilled requirement from
\$\RQFixedUnitCost{} to \$\RQEvoUnitCost{}, an improvement of
\RQUnitSaving{}\%. Table~\ref{tab:resource-efficiency} reports normalized
cost and throughput for all configurations. Evolving resolves more
requirements per dollar and per recorded Agent hour than Fixed, although
several external configurations have higher throughput at lower overall
accuracy. In particular, mini-SWE-agent/DeepSeek offers the highest
requirement throughput per dollar and hour, but the lowest Requirement
Accuracy, \RQDeepSeekReq{}\%. Efficiency and coverage must therefore be
considered jointly when choosing an operating point.

\noindent\textbf{Cross-framework trade-offs.}
Figure~\ref{fig:rq-analysis}(c) shows that Evolving achieves higher accuracy while costing
\RQCodexCostSaving{}\% less than Codex and \RQClaudeCostSaving{}\% less
than Claude Code. Mini-SWE-agent uses less recorded Agent time than either
FilingLoop variant. Throughput uses cumulative recorded invocation time,
excluding pauses and independent assessment; missing segments are not
imputed. These rates measure recorded work rather than wall-clock speed
(Appendix~\ref{app:evaluation-metrics}).

\ifdefined\FilingLoopFullPaper
\noindent\textbf{Where the additional cost occurs.}
The \$\AGExtraCost{} increase comprises \$\AGExtraReviewerCost{} in
Reviewer-attributed usage, \$\AGExtraFilerCost{} in Filer-attributed usage,
and \$\AGExtraSharedCost{} in shared services. It therefore includes both
producing guidance and executing and assessing the resulting behavior.
The entire difference should not be interpreted as the price of the skill
editor alone. Dividing the extra cost by the net requirement gain gives
\$\AGMarginalReqCost{} per additional fulfilled requirement on this stream;
this descriptive ratio does not estimate the cost of resolving any
particular remaining request. Appendix~\ref{app:resource-breakdown}
separates role usage from shared-service resources.
\fi

\noindent\textbf{Findings (RQ3).}
\textbf{(1)~Evolution lowers cost per fulfilled requirement.}
Evolving spends more per task, but its cost per fulfilled requirement falls
by \RQUnitSaving{}\%. These usage-based benchmark costs depend on the service mix.
\textbf{(2)~More spending need not resolve more requirements.}
Codex and Claude Code incur higher cost and Agent time while fulfilling fewer
requirements. Additional calls should target unresolved requests and verify
committed amendments.
\textbf{(3)~Selective escalation requires separate evaluation.}
Mini-SWE-agent's higher throughput suggests a low-cost first pass with
targeted escalation; we do not test a cascade here.

\subsection{Harmful Revisions (RQ4)}
\label{sec:harmful-revisions}

% Final rates confirmed by the author; see notes/main-results/harm-reporting.json.
We measure harmful revisions at the task level: a task is flagged if at least
one filing change is judged harmful against permitted evidence and released
company facts. Among tasks with completed edit audits, Fixed has a
harmful-revision rate of 5.8\%, compared with 4.5\% for Evolving. Higher
fulfillment under evolution thus coincides with a 1.3~pp lower observed harm
rate. Each affected task contributes one flag regardless of its edit count,
so this comparison describes harm prevalence across tasks without quantifying
the severity of individual changes.
Edit-audit coverage is 99.74\% and 100\% of scheduled tasks, respectively
(Appendix~\ref{app:completion-and-harm}); unaudited tasks are excluded from the
harm-rate denominator. Full coverage provides an audit outcome for every
scheduled task, including those with harmful edits. Both variants still
produce harmful changes, so the audit complements fulfillment scores by
checking whether requested revisions introduce new errors into the filing.

\ifdefined\FilingLoopFullPaper
A task can fulfill a reference request and still be flagged for an
unsupported change elsewhere. It can also leave a request unresolved
without introducing a harmful edit. The two measures therefore answer
different questions about the same final filing. Full audit coverage means
that every scheduled task received an edit-audit outcome; it does not mean
that every interaction met the joint completion criterion, nor that every
change was harmless. Thus Evolving's 100\% edit-audit coverage is consistent
with its 99.74\% task completion in Table~\ref{tab:main-results}.

These rates describe the joint evolving policy under a common audit
protocol. Establishing whether either role's updated guidance prevents
specific errors, or reduces their severity, would require edit-level
comparisons and role-specific update ablations.
\fi

\section{Conclusion}
\label{sec:conclusion}

We introduced FilingLoop, an interactive benchmark and self-evolving agent
framework grounded in archived SEC EDGAR disclosure reviews. Its evaluation
connects issue discovery and evidence acquisition to verified fulfillment
in final filings and responses. Across \BenchTasks{} tasks from
\BenchCompanies{} companies in the evaluated stream, the joint evolving
policy improves requirement fulfillment and full comment resolution,
while reducing cost per fulfilled requirement by
\RQUnitSaving{}\%, without changing model weights or retrying scored tasks.
The role-level results reveal a key bottleneck: improved revision cannot
compensate for undiscovered deficiencies. Further progress therefore calls
for agents that actively uncover omissions, retain unresolved obligations
across review rounds, and verify every requested change in its intended
deliverable. FilingLoop records how agents investigate omissions, obtain
company facts, and revise the relevant documents, giving the skill editor
a record of which requested actions were fulfilled and which remain unresolved
after the final response.

\bibliographystyle{ACM-Reference-Format}
% Original exports and provenance follow the policy in AGENTS.md.
\bibliography{references/export-compat, references/filingloop}

@article{healy2001information,
  title={Information asymmetry, corporate disclosure, and the capital markets: A review of the empirical disclosure literature},
  author={Healy, Paul M and Palepu, Krishna G},
  journal={Journal of accounting and economics},
  volume={31},
  number={1-3},
  pages={405--440},
  year={2001},
  publisher={Elsevier}
}

@article{johnston2017regulatory,
  title={Regulatory oversight of financial reporting: Securities and Exchange Commission comment letters},
  author={Johnston, Rick and Petacchi, Reining},
  journal={Contemporary Accounting Research},
  volume={34},
  number={2},
  pages={1128--1155},
  year={2017},
  publisher={Wiley Online Library}
}

@article{wu2023bloomberggpt,
  title={Bloomberggpt: A large language model for finance},
  author={Wu, Shijie and Irsoy, Ozan and Lu, Steven and Dabravolski, Vadim and Dredze, Mark and Gehrmann, Sebastian and Kambadur, Prabhanjan and Rosenberg, David and Mann, Gideon},
  journal={arXiv preprint arXiv:2303.17564},
  year={2023}
}

@article{yang2023fingpt,
  title={Fingpt: Open-source financial large language models},
  author={Yang, Hongyang and Liu, Xiao-Yang and Wang, Christina Dan},
  journal={arXiv preprint arXiv:2306.06031},
  year={2023}
}

@article{zhou2024finrobot,
  title={Finrobot: Ai agent for equity research and valuation with large language models},
  author={Zhou, Tianyu and Wang, Pinqiao and Wu, Yilin and Yang, Hongyang},
  journal={arXiv preprint arXiv:2411.08804},
  year={2024}
}

@inproceedings{chen2021finqa,
  title={Finqa: A dataset of numerical reasoning over financial data},
  author={Chen, Zhiyu and Chen, Wenhu and Smiley, Charese and Shah, Sameena and Borova, Iana and Langdon, Dylan and Moussa, Reema and Beane, Matt and Huang, Ting-Hao and Routledge, Bryan R and others},
  booktitle={Proceedings of the 2021 Conference on Empirical Methods in Natural Language Processing},
  pages={3697--3711},
  year={2021}
}

@inproceedings{zhu2021tat,
  title={TAT-QA: A question answering benchmark on a hybrid of tabular and textual content in finance},
  author={Zhu, Fengbin and Lei, Wenqiang and Huang, Youcheng and Wang, Chao and Zhang, Shuo and Lv, Jiancheng and Feng, Fuli and Chua, Tat-Seng},
  booktitle={Proceedings of the 59th annual meeting of the Association for Computational Linguistics and the 11th international joint conference on natural language processing (volume 1: long papers)},
  pages={3277--3287},
  year={2021}
}

@inproceedings{chen2022convfinqa,
  title={Convfinqa: Exploring the chain of numerical reasoning in conversational finance question answering},
  author={Chen, Zhiyu and Li, Shiyang and Smiley, Charese and Ma, Zhiqiang and Shah, Sameena and Wang, William Yang},
  booktitle={Proceedings of the 2022 conference on empirical methods in natural language processing},
  pages={6279--6292},
  year={2022}
}

@article{islam2023financebench,
  title={Financebench: A new benchmark for financial question answering},
  author={Islam, Pranab and Kannappan, Anand and Kiela, Douwe and Qian, Rebecca and Scherrer, Nino and Vidgen, Bertie},
  journal={arXiv preprint arXiv:2311.11944},
  year={2023}
}

@inproceedings{krumdick2024bizbench,
  title={BizBench: A quantitative reasoning benchmark for business and finance},
  author={Krumdick, Michael and Koncel-Kedziorski, Rik and Lai, Viet Dac and Reddy, Varshini and Lovering, Charles and Tanner, Chris},
  booktitle={Proceedings of the 62nd Annual Meeting of the Association for Computational Linguistics (Volume 1: Long Papers)},
  pages={8309--8332},
  year={2024}
}

@article{xie2024finben,
  title={Finben: A holistic financial benchmark for large language models},
  author={Xie, Qianqian and Han, Weiguang and Chen, Zhengyu and Xiang, Ruoyu and Zhang, Xiao and He, Yueru and Xiao, Mengxi and Li, Dong and Dai, Yongfu and Feng, Duanyu and others},
  journal={Advances in neural information processing systems},
  volume={37},
  pages={95716--95743},
  year={2024}
}

@article{bigeard2025finance,
  title={Finance agent benchmark: Benchmarking llms on real-world financial research tasks},
  author={Bigeard, Antoine and Nashold, Langston and Krishnan, Rayan and Wu, Shirley},
  journal={arXiv preprint arXiv:2508.00828},
  year={2025}
}

@inproceedings{liu2024agentbench,
  title={Agentbench: Evaluating llms as agents},
  author={Liu, Xiao and Yu, Hao and Zhang, Hanchen and Xu, Yifan and Lei, Xuanyu and Lai, Hanyu and Gu, Yu and Ding, Hangliang and Men, Kaiwen and Yang, Kejuan and others},
  booktitle={International Conference on Learning Representations},
  volume={2024},
  pages={52989--53046},
  year={2024}
}

@inproceedings{mialon2024gaia,
  title={Gaia: a benchmark for general ai assistants},
  author={Mialon, Gr{\'e}goire and Fourrier, Cl{\'e}mentine and Wolf, Thomas and LeCun, Yann and Scialom, Thomas},
  booktitle={International Conference on Learning Representations},
  volume={2024},
  pages={9025--9049},
  year={2024}
}

@article{drouin2024workarena,
  title={WorkArena: How capable are web agents at solving common knowledge work tasks?},
  author={Drouin, Alexandre and Gasse, Maxime and Caccia, Massimo and Laradji, Issam H and Del Verme, Manuel and Marty, Tom and Boisvert, L{\'e}o and Thakkar, Megh and Cappart, Quentin and Vazquez, David and others},
  journal={arXiv preprint arXiv:2403.07718},
  year={2024}
}

@article{yao2024tau,
  title={{$\tau$-bench: A Benchmark for Tool-Agent-User Interaction in Real-World Domains}},
  author={Yao, Shunyu and Shinn, Noah and Razavi, Pedram and Narasimhan, Karthik},
  journal={arXiv preprint arXiv:2406.12045},
  year={2024}
}

@article{barres2025tau,
  title={{$\tau^2$-Bench: Evaluating Conversational Agents in a Dual-Control Environment}},
  author={Barres, Victor and Dong, Honghua and Ray, Soham and Si, Xujie and Narasimhan, Karthik},
  journal={arXiv preprint arXiv:2506.07982},
  year={2025}
}

@article{liu2024lost,
  title={Lost in the middle: How language models use long contexts},
  author={Liu, Nelson F and Lin, Kevin and Hewitt, John and Paranjape, Ashwin and Bevilacqua, Michele and Petroni, Fabio and Liang, Percy},
  journal={Transactions of the association for computational linguistics},
  volume={12},
  pages={157--173},
  year={2024}
}

@article{yao2022react,
  title={React: Synergizing reasoning and acting in language models},
  author={Yao, Shunyu and Zhao, Jeffrey and Yu, Dian and Du, Nan and Shafran, Izhak and Narasimhan, Karthik and Cao, Yuan},
  journal={arXiv preprint arXiv:2210.03629},
  year={2022}
}

@article{shinn2023reflexion,
  title={Reflexion: Language agents with verbal reinforcement learning},
  author={Shinn, Noah and Cassano, Federico and Gopinath, Ashwin and Narasimhan, Karthik and Yao, Shunyu},
  journal={Advances in neural information processing systems},
  volume={36},
  pages={8634--8652},
  year={2023}
}

@article{madaan2023self,
  title={Self-refine: Iterative refinement with self-feedback},
  author={Madaan, Aman and Tandon, Niket and Gupta, Prakhar and Hallinan, Skyler and Gao, Luyu and Wiegreffe, Sarah and Alon, Uri and Dziri, Nouha and Prabhumoye, Shrimai and Yang, Yiming and others},
  journal={Advances in neural information processing systems},
  volume={36},
  pages={46534--46594},
  year={2023}
}

@inproceedings{zhao2024expel,
  title={Expel: Llm agents are experiential learners},
  author={Zhao, Andrew and Huang, Daniel and Xu, Quentin and Lin, Matthieu and Liu, Yong-Jin and Huang, Gao},
  booktitle={Proceedings of the AAAI Conference on Artificial Intelligence},
  volume={38},
  number={17},
  pages={19632--19642},
  year={2024}
}

@article{wang2023voyager,
  title={Voyager: An open-ended embodied agent with large language models},
  author={Wang, Guanzhi and Xie, Yuqi and Jiang, Yunfan and Mandlekar, Ajay and Xiao, Chaowei and Zhu, Yuke and Fan, Linxi and Anandkumar, Anima},
  journal={arXiv preprint arXiv:2305.16291},
  year={2023}
}

@article{yuksekgonul2024textgrad,
  title={Textgrad: Automatic" differentiation" via text},
  author={Yuksekgonul, Mert and Bianchi, Federico and Boen, Joseph and Liu, Sheng and Huang, Zhi and Guestrin, Carlos and Zou, James},
  journal={arXiv preprint arXiv:2406.07496},
  year={2024}
}

@inproceedings{agrawal2026gepa,
  title={Gepa: Reflective prompt evolution can outperform reinforcement learning},
  author={Agrawal, Lakshya A and Tan, Shangyin and Soylu, Dilara and Ziems, Noah and Khare, Rishi and Opsahl-Ong, Krista and Singhvi, Arnav and Shandilya, Herumb and Ryan, Michael J and Jiang, Meng and others},
  booktitle={International Conference on Learning Representations},
  volume={2026},
  pages={8479--8565},
  year={2026}
}

@article{wu2023autogen,
  title={Autogen: Enabling next-gen llm applications via multi-agent conversation},
  author={Wu, Qingyun and Bansal, Gagan and Zhang, Jieyu and Wu, Yiran and Li, Beibin and Zhu, Erkang and Jiang, Li and Zhang, Xiaoyun and Zhang, Shaokun and Liu, Jiale and others},
  journal={arXiv preprint arXiv:2308.08155},
  year={2023}
}

@inproceedings{hong2024metagpt,
  title={MetaGPT: Meta programming for a multi-agent collaborative framework},
  author={Hong, Sirui and Zhuge, Mingchen and Chen, Jonathan and Zheng, Xiawu and Cheng, Yuheng and Wang, Jinlin and Zhang, Ceyao and Yau, Steven and Lin, Zijuan and Zhou, Liyang and others},
  booktitle={International Conference on Learning Representations},
  volume={2024},
  pages={23247--23275},
  year={2024}
}

@article{li2023camel,
  title={Camel: Communicative agents for" mind" exploration of large language model society},
  author={Li, Guohao and Hammoud, Hasan and Itani, Hani and Khizbullin, Dmitrii and Ghanem, Bernard},
  journal={Advances in neural information processing systems},
  volume={36},
  pages={51991--52008},
  year={2023}
}

@inproceedings{trivedi2024appworld,
  title={Appworld: A controllable world of apps and people for benchmarking interactive coding agents},
  author={Trivedi, Harsh and Khot, Tushar and Hartmann, Mareike and Manku, Ruskin and Dong, Vinty and Li, Edward and Gupta, Shashank and Sabharwal, Ashish and Balasubramanian, Niranjan},
  booktitle={Proceedings of the 62nd Annual Meeting of the Association for Computational Linguistics (Volume 1: Long Papers)},
  pages={16022--16076},
  year={2024}
}

@article{xu2026theagentcompany,
  title={Theagentcompany: benchmarking llm agents on consequential real world tasks},
  author={Xu, Frank Fangzheng and Song, Yufan and Li, Boxuan and Tang, Yuxuan and Jain, Kritanjali and Bao, Mengxue and Wang, Zora and Zhou, Xuhui and Guo, Zhitong and Cao, Murong and others},
  journal={Advances in Neural Information Processing Systems},
  volume={38},
  year={2026}
}

@article{wang2024agent,
  title={Agent workflow memory},
  author={Wang, Zora Zhiruo and Mao, Jiayuan and Fried, Daniel and Neubig, Graham},
  journal={arXiv preprint arXiv:2409.07429},
  year={2024}
}

@article{zhou2022large,
  title={Large language models are human-level prompt engineers},
  author={Zhou, Yongchao and Muresanu, Andrei Ioan and Han, Ziwen and Paster, Keiran and Pitis, Silviu and Chan, Harris and Ba, Jimmy},
  journal={arXiv preprint arXiv:2211.01910},
  year={2022}
}

@inproceedings{yang2024large,
  title={Large language models as optimizers},
  author={Yang, Chengrun and Wang, Xuezhi and Lu, Yifeng and Liu, Hanxiao and Le, Quoc V and Zhou, Denny and Chen, Xinyun},
  booktitle={International Conference on Learning Representations},
  volume={2024},
  pages={12028--12068},
  year={2024}
}

@inproceedings{pryzant2023automatic,
  title={Automatic prompt optimization with “gradient descent” and beam search},
  author={Pryzant, Reid and Iter, Dan and Li, Jerry and Lee, Yin and Zhu, Chenguang and Zeng, Michael},
  booktitle={Proceedings of the 2023 conference on empirical methods in natural language processing},
  pages={7957--7968},
  year={2023}
}

@inproceedings{hu2025automated,
  title={Automated design of agentic systems},
  author={Hu, Shengran and Lu, Cong and Clune, Jeff},
  booktitle={International Conference on Learning Representations},
  volume={2025},
  pages={21344--21377},
  year={2025}
}

@inproceedings{zhang2025aflow,
  title={Aflow: Automating agentic workflow generation},
  author={Zhang, Jiayi and Xiang, Jinyu and Yu, Zhaoyang and Teng, Fengwei and Chen, Xionghui and Chen, Jiaqi and Zhuge, Mingchen and Cheng, Xin and Hong, Sirui and Wang, Jinlin and others},
  booktitle={International Conference on Learning Representations},
  volume={2025},
  pages={34040--34077},
  year={2025}
}

@article{zhao2026advancing,
  title={Advancing esg intelligence: An expert-level agent and comprehensive benchmark for sustainable finance},
  author={Zhao, Yilei and Zhang, Wentao and Xiao, Lei and Zheng, Yandan and Liu, Mengpu and Lim, Wei Yang Bryan},
  journal={arXiv preprint arXiv:2601.08676},
  year={2026}
}

% Appendix entry point; each section is maintained in a separate source file.
\onecolumn
\appendix

% A dedicated contents stream keeps main-paper sections out of this directory.
% Section titles, letters, hyperlinks, and page numbers update on recompilation.
\makeatletter
% Single-column supplement with page numbers and no review gutters.
\fancypagestyle{filingloopappendix}{%
  \fancyhf{}%
  \fancyfoot[C]{\footnotesize\thepage}%
  \renewcommand{\headrulewidth}{0pt}%
}
\pagestyle{filingloopappendix}
\newcommand{\AppendixSection}[2]{%
  \section{#1}\label{#2}%
  \addcontentsline{atoc}{section}{\protect\numberline{\thesection}#1}%
}
\newcommand{\AppendixSubsection}[2]{%
  \subsection{#1}\label{#2}%
  \addcontentsline{atoc}{subsection}{\protect\numberline{\thesubsection}#1}%
}
% The directory is a separate page; page numbering continues through the text.
\fancypagestyle{filingloopappendixcontents}{%
  \fancyhf{}%
  \fancyfoot[C]{\footnotesize\thepage}%
  \renewcommand{\headrulewidth}{0pt}%
}
\thispagestyle{filingloopappendixcontents}
\phantomsection
\pdfbookmark[0]{Contents of Appendix}{appendix-contents}
\label{app:overview}
{\noindent\fontsize{13}{16}\selectfont\bfseries Contents of Appendix\par}
\medskip
\begingroup
  \fontsize{10}{12}\selectfont
  \def\@pnumwidth{2em}%
  \def\@tocrmarg{3em}%
  \def\l@section#1#2{%
    \addvspace{5pt}%
    {\bfseries\def\@dotsep{10000}%
      \@dottedtocline{1}{1.5em}{2.4em}{#1}{#2}}%
  }%
  \def\l@subsection#1#2{%
    \addvspace{1pt}%
    \@dottedtocline{2}{3.9em}{2.8em}{#1}{#2}%
  }%
  \@input{\jobname.atoc}%
  \if@filesw
    \expandafter\newwrite\csname tf@atoc\endcsname
    \immediate\openout\csname tf@atoc\endcsname \jobname.atoc\relax
  \fi
\endgroup
\makeatother
\clearpage

\AppendixSection{Data Sources and Benchmark Construction}{app:data-construction}

Construction separates historical source documents, cross-round issue
threads, and task-specific requirements (Section~\ref{sec:benchmark}).
Source evidence fixes eligibility before agents inspect documents,
raise comments, or revise disclosures.

\AppendixSubsection{Source Corpus and Document Provenance}{app:data-sources}

The collection began with 300 company candidates selected by year-stratified
sampling of registration activity in EDGAR quarterly indexes. This discovery
pool includes records that do not yield eligible benchmark tasks. The
evaluated corpus contains \BenchCompanies{} companies and
\BenchPackages{} review packages: \BenchRegistrationPackages{}
registration-review packages and \BenchAnnualPackages{} annual-report
package. Eligible review events span \BenchReviewStart{} to
\BenchReviewEnd{}. These dates refer to the historical reviews, rather than
the time at which records were collected or became publicly accessible.
EDGAR correspondence is released after the associated review process, so
its public availability date can differ from the date of the underlying
exchange.\footnote{SEC guidance on correspondence and its release:
\url{https://www.sec.gov/search-filings/edgar-search-assistance/how-search-edgar-correspondence}.}

\begin{figure}[!htbp]
  \centering
  \includegraphics[width=\linewidth]{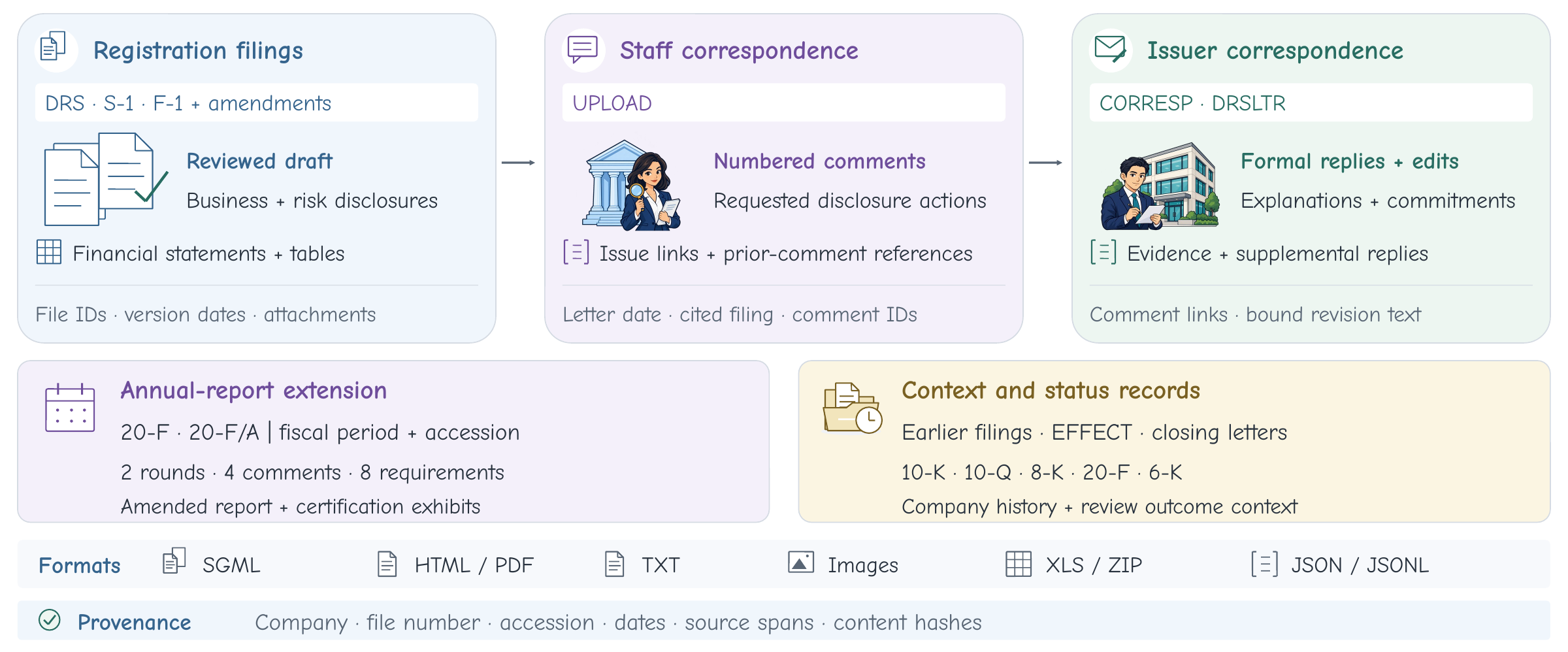}
  \caption{\textbf{Source records and data types.} Five source groups retain
  distinct identities. Icons mark record categories and archived or processed
  formats; shared provenance links documents to review rounds and reference evidence.}
  \label{fig:app-source-records}
  \Description{Three illustrated cards connect registration filings, staff
  correspondence, and issuer correspondence. Distinct document, comment,
  reply, calendar, and archive icons identify all five source groups.
  Filing content includes business and risk disclosures, financial statements,
  and tables; correspondence includes numbered requests, issue references,
  explanations, commitments, and revision evidence. The annual-report card
  lists two rounds, four comments, eight requirements, and certification
  exhibits. Context records include earlier periodic filings and status letters.
  A format strip identifies SGML, HTML/PDF, TXT, images, XLS/ZIP, and JSON/JSONL
  as archived or processed representations. Shared provenance retains company,
  file number, accession, dates, source spans, and content hashes.}

\end{figure}

Figure~\ref{fig:app-source-records} summarizes source groups and data types.
Source identities, hashes, and evidence spans link records to their original
submissions. Document dates order the review history; filing dates are
preserved separately. Only permitted earlier records enter agent workspaces;
current correspondence and later amendments remain hidden reference evidence.
The format strip distinguishes archived document members from processed
representations: normalized text and structured records support reconstruction
and source tracing, while original submission files retain their source
identity.

\AppendixSubsection{Package, Round, and Issue Reconstruction}{app:record-linkage}

A registration-review package groups filing versions and correspondence
using file numbers, form families, dates, and explicit letter references.
Issuer correspondence is classified to distinguish substantive responses
from acceleration and withdrawal requests. Retained links record their
supporting basis, while ambiguous relationships remain in audit records.

Each substantive staff letter defines a review round. The reconstruction
binds it to the reviewed filing, aligns numbered comments with issuer
responses, and identifies the associated revision when one exists. Alignment
uses comment numbers, repeated comment text, and references to the staff
letter; supplemental replies can provide evidence for individual comments.
Revision evidence comprises uniquely anchored text replacements or relevant
differences between the bound versions. Page-level differences can include
unrelated financial or offering updates, so they require interpretation
against the requested action. Comment and response spans retain source
locations, including requests and answers continued across page boundaries.

References to earlier comments connect recurring concerns into issue
threads. Each occurrence nevertheless remains attached to its own round and
reviewed version. A task begins before that round's review: agents receive
the filing and permitted history, while the current letter, replies, and
subsequent revisions remain hidden. This separation preserves the Reviewer's
need to discover issues independently.

\AppendixSubsection{Atomic Requirements and Evidence Verification}{app:requirement-annotation}

Annotations decompose an issue thread into independently checkable
requirements supported by the historical response and revision evidence.
Each requirement records its requested action, disposition type, supporting
quotation, applicable comment identifiers, and any required company facts.
The four disposition types and their expected deliverables are summarized in
Table~\ref{tab:app-requirement-types}. A request to amend a disclosure is
distinct from a request to explain it to staff, supply supporting material,
or make a future commitment. Historical responses establish an
evidence-supported resolution, while requirements describe the action to be
satisfied rather than requiring agents to reproduce the historical wording or copy
the company's exact response.

The registration-review annotations use an initial \textit{gpt-6-astra}
annotation and an independent \textit{claude-opus-5.5} audit. The auditor
reviews the proposed requirements and facts while independently determining
handling and answerability without seeing the annotator's decisions for
those fields. Disagreements undergo reconciliation and a further audit;
quotation checks verify support in the supplied sources. The annual-report
extension uses separate annotation and audit calls to
\textit{gpt-6.1-sol}, followed by reconciliation, comment mapping, and source
verification. The project team verified the original reference collection.
These procedures preserve the distinction between agreement among model
annotations and verification against the underlying documents.

Requirements are mapped back to the comments whose requests they answer.
For a thread spanning several rounds, independent mappings retain agreed
comment--requirement links. A requirement introduced by a later follow-up is
not assigned retrospectively to an earlier comment. Repeated requests can
refer to the same requirement in several rounds, whereas an earlier request
that has already been satisfied is not automatically carried into every
follow-up. Consequently, the benchmark counts requirement occurrences per
eligible comment and round, rather than deduplicating them across threads.
Annotation and mapping records retain fingerprints of their inputs, prompts,
and model identities. Unchanged inputs reuse saved annotations; altered
inputs require a new record. This preserves the connection between each
reference requirement and the source version used to establish it, including
when a review package is reconstructed.

% Counts and within-type fact dependence use notes/benchmark-statistics/summary.json.
\begin{table}[!htbp]
  \caption{Requirement dispositions and evidence targets.
  Fact percentages are within each disposition.}
  \label{tab:app-requirement-types}
  \centering
  \normalsize
  \setlength{\tabcolsep}{6pt}
  \renewcommand{\arraystretch}{1.25}
  \begin{tabularx}{\linewidth}{lrrr>{\raggedright\arraybackslash}X}
    \toprule
    \rowcolor{MRPanel}
    \textbf{Disposition} & \textbf{Count} & \textbf{Share} &
    \textbf{Needs facts} & \textbf{Evidence target} \\
    \midrule
    Disclosure change & 4,668 & 76.5\% & 2,093 (44.8\%) &
    \mbox{Amended filing with the required addition, edit, or removal} \\
    \addlinespace[1pt]
    Explanation to staff & 1,075 & 17.6\% & 805 (74.9\%) &
    \mbox{Formal response with the requested explanation or analysis} \\
    \addlinespace[1pt]
    Supporting material & 173 & 2.8\% & 33 (19.1\%) &
    \mbox{Submission containing the requested document, exhibit, or copy} \\
    \addlinespace[1pt]
    Commitment & 183 & 3.0\% & 149 (81.4\%) &
    \mbox{Formal response with an explicit commitment to future action} \\
    \midrule
    \textbf{Total} & \textbf{6,099} & \textbf{100\%} & \textbf{3,080 (50.5\%)} &
    \mbox{Requirement-specific evidence in the designated deliverable} \\
    \bottomrule
  \end{tabularx}
\end{table}

Verification follows the required deliverable: a reply claiming that a filing
was revised does not replace evidence in the amended filing itself.
Conversely, an explanation or a commitment is assessed in the formal response,
while a request for supporting material requires the material to be supplied.
Mixed comments retain separate evidence targets for each constituent
requirement.

\AppendixSubsection{Scoring Eligibility and Corpus Coverage}{app:eligibility-and-coverage}

A comment is scorable when it is detectable from the permitted record, its
annotation is usable, and at least one applicable requirement has sufficient
evidence. Company facts are scoped to the current comment and round. Each
must be required by an applicable requirement and supported by a quotation
from the bound response, revision, filing, or exhibit evidence. Facts
supported only in later rounds are withheld. Requirements lacking necessary current-round fact
evidence are excluded before evaluation, removing 318 occurrences. The
retained corpus includes 3,080 fact-dependent requirements
(\BenchFactsPct{} of \BenchRequirements{}). This annotation records what
fulfillment requires, without asserting that every marked fact is absent
from every visible document.

Of 412 reconstructed review rounds in the main collection, \BenchTasks{}
are eligible and 32 remain in the exclusion records. A further 93 terminal
diagnostics are excluded: registration effectiveness and review closure
provide process context, without establishing that a filing contains no
remaining deficiencies. The final stream contains \BenchComments{}
scorable comments and \BenchRequirements{} requirement occurrences,
using \BenchReviewedFilings{} distinct primary versions and 1,311 distinct
documents in permitted visible records.

The annual-report extension adds \BenchAnnualTasks{} tasks, four comments,
and eight requirement occurrences from an existing issuer's Form 20-F review.
Package identity combines company identifier, form, fiscal period, and
original accession, distinguishing it from IPO review and other reporting
years. The first round includes an amended report and certification
exhibits; the second uses an explanatory response without another amendment.
Thus, reconstruction accommodates both revision and explanation outcomes.
This small extension adds document-type coverage without supporting a
separate characterization of annual-report review performance.

All \BenchTasks{} tasks form one evaluation stream without training or
validation partitions, with identical reference inventories across
configurations. Figure~\ref{fig:benchmark-statistics} counts whitespace-separated
words per primary filing, excluding earlier history: the median is 111,804
and the 90th percentile is 155,266. These task-weighted counts differ from
model tokens and the total volume of visible material.

\AppendixSection{Task Environment and Interaction Protocol}{app:task-environment}

The environment presents historical review rounds as Reviewer--Filer
interactions. A trusted host routes submissions, maintains authoritative
artifacts, and supplies bounded facts. Role-specific workspaces separate
source evidence, proposed revisions, and committed outcomes.

\AppendixSubsection{Information Boundaries and Role Workspaces}{app:information-boundaries}

Each task starts from the reviewed filing and the earlier records permitted
by its historical boundary (Appendix~\ref{app:data-construction}). The
materialized record includes primary text, section and printed-page indexes,
linked exhibits and attachments, and available image records. Navigation
metadata identifies source locations without supplying the reference
comments or marking which passages contain scored deficiencies. The host retains current historical correspondence, subsequent amendments,
requirement annotations, and the complete fact ledger.

Table~\ref{tab:app-role-visibility} summarizes access during an active task.
Reviewer and Filer use separate role contexts. Both can inspect the permitted
record and the current authoritative filing, while formal comments and
responses provide the exchange between them. Company questions and released
answers enter only the Filer's private history; the Reviewer can assess
information that the Filer subsequently includes in its formal response or
the revised filing. Documents and exchange records preserve case history without sharing
private role conversations.

The record is read-only, and proposed edits are submitted separately from
the authoritative draft. Adapters receive a turn-specific workspace and
submission contract rather than access to the controller's reference files.
Private fact-service diagnostics, including historical comment identities
used for matching, remain outside the acting agents' inputs. Current-task
grading judgments are likewise withheld during execution. The subsequent
use of locked feedback for procedural evolution is described in
Appendix~\ref{app:agent-evolution}.

\begin{table}[!htbp]
  \caption{Information access during a task. Private fact-service records and
  current-task grading judgments remain separated from the public exchange.}
  \label{tab:app-role-visibility}
  \centering
  \normalsize
  \setlength{\tabcolsep}{6pt}
  \renewcommand{\arraystretch}{1.25}
  \begin{tabularx}{\linewidth}{>{\raggedright\arraybackslash}Xlll}
    \toprule
    \rowcolor{MRPanel}
    \textbf{Record or artifact} & \textbf{Reviewer} & \textbf{Filer} & \textbf{Trusted host} \\
    \midrule
    Permitted filing and earlier history & Read & Read & Selects sources \\
    Current draft and committed edits & Read & Propose edits & Validates and commits \\
    Generated review comments & Author & Read & Retains and routes \\
    Formal issuer responses & Read & Author & Retains and routes \\
    Company questions and released answers & Hidden & Private access & Serves scoped facts \\
    Current historical letters and amendments & Hidden & Hidden & Reference evidence \\
    Atomic requirements and complete fact ledger & Hidden & Hidden & Reference evidence \\
    Current-task grading judgments & Hidden & Hidden & Assesses final artifacts \\
    \bottomrule
  \end{tabularx}
\end{table}

\AppendixSubsection{Tools, Submissions, and Committed Artifacts}{app:tools-and-artifacts}

The built-in agents combine document reading with shell-based inspection
and file preparation. Text can be read by line range or printed-page range,
and linked images can be inspected separately from surrounding extracted
text. Section indexes and source paths support focused reading across long
filings. External frameworks retain their native tool loops while using the
same permitted records and submission contract. Their measured resource
usage is reported alongside nominal execution budgets in
Section~\ref{sec:experimental-setup}.

The initial Reviewer submission is a list of comments with stable identifiers,
an optional heading, a substantive request, a source document, and a supporting
quotation. Identifiers must be distinct, and quotations must match a supplied
document or current draft under the controller's source checks. Source quotations ground comment locations; semantic validity is assessed
separately. New comments introduced during re-review
follow the same contract and cannot reuse identifiers already assigned in
the case.

The Filer supplies exactly one response for each active comment, choosing
\texttt{revised}, \texttt{explained}, \texttt{information\_requested}, or
\texttt{declined}. Each response identifies the comment and provides the
formal reply; information requests additionally contain explicit questions.
Text edits are separate transactions containing the text to find, its
replacement, and the comment identifiers motivating the change. The host
applies a replacement only when the target is unique in the current draft.
It first attempts exact matching, then a unique whitespace-normalized match
for eligible anchors. Missing or ambiguous targets are rejected with reasons.
Edits are applied in order against the evolving draft, making their effects
reproducible from the original text and committed transaction records.

Tasks with supporting-document or image manifests also identify the artifacts
that may be changed. Supporting-text revisions name the authorized document,
its base hash, and the proposed text edits. Image removal identifies a
currently present asset and verifies its source hash before changing its
manifest state. Original source files remain preserved. Directly altering a
workspace copy or claiming that an attachment was changed does not replace
the corresponding committed operation. These controls preserve the
distinction between a proposed action and the final deliverable evaluated.

\AppendixSubsection{Bounded Company-Fact Acquisition}{app:fact-service}

The fact service represents company information supported by the historical
record. It is accessed through a Filer information request associated with
an active generated comment. The trusted host matches that concern to
current-round reference comments and returns the permitted facts associated
with the matched concerns. It does not expose those reference identifiers,
the reference response, or the complete fact ledger. The returned material
is restricted by the source and temporal eligibility rules in
Appendix~\ref{app:eligibility-and-coverage}.

The service is therefore a bounded source of evidence, rather than an
unrestricted company question-answering system. A request can return no
material because no concern was linked or no releasable facts support it.
Such an answer describes service coverage, not the real company's knowledge
or willingness to respond. Released facts may include the company's own
positions, analyses, or commitments; they remain attributable company
statements rather than independent certification of those claims.

\begin{figure}[!htbp]
  \centering
  \includegraphics[width=\linewidth]{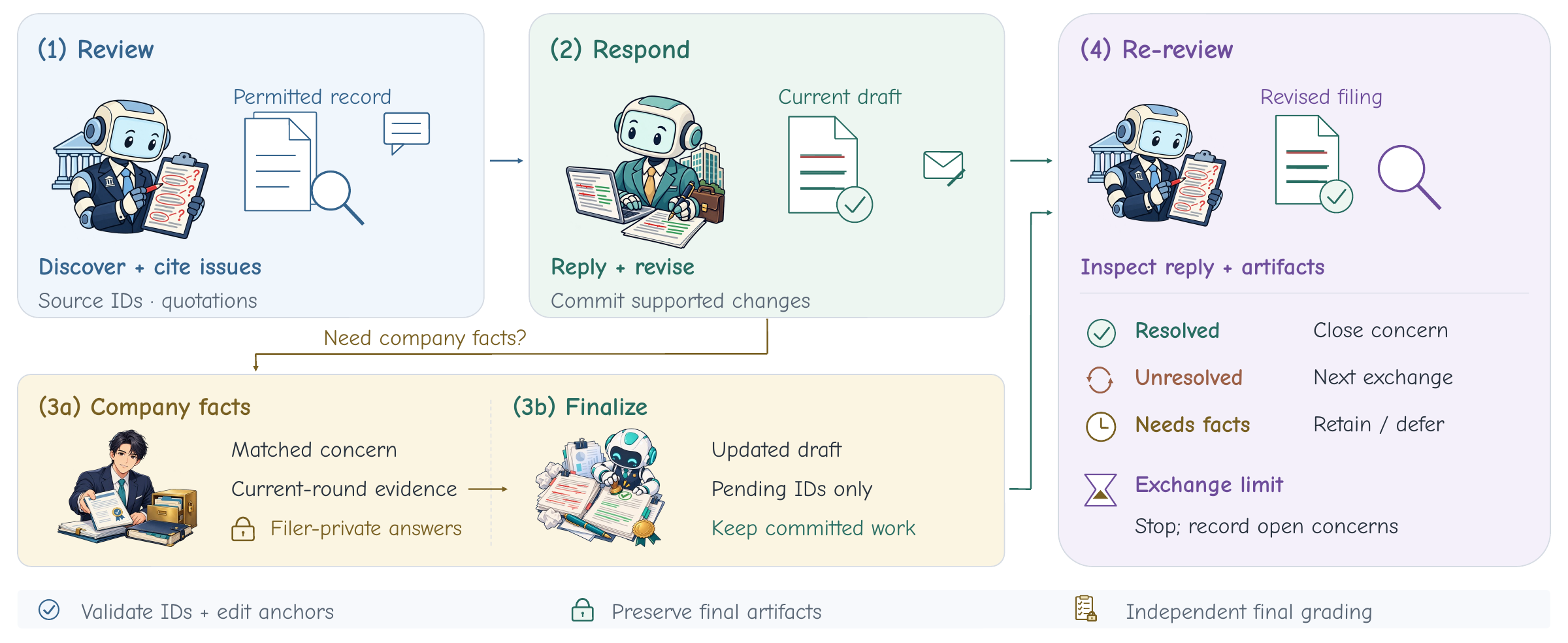}
  \caption{\textbf{Illustrated interaction protocol.} Reviewer and Filer exchange
  grounded comments, responses, and revisions. Company facts remain private to
  the Filer; committed changes persist through finalization. Re-review controls
  iteration, while independent grading assesses frozen final artifacts.}
  \label{fig:app-interaction-protocol}
  \Description{An illustrated workflow uses a navy Reviewer robot, a teal
  Filer robot, source documents, marked-up drafts, and magnifiers to depict
  discovery, response, and re-review. A gold private lane shows a human
  company representative supplying documented facts and a robot committing
  a revised filing with red before-text and green after-text. Accepted work
  is retained while pending responses are finalized. Distinct check, loop,
  clock, and hourglass symbols mark resolution, another exchange, deferred
  information, and the exchange limit. Validation, preservation of final
  artifacts, and independent grading remain distinct controls.}

\end{figure}

Figure~\ref{fig:app-interaction-protocol} shows how fact acquisition connects
to revision. Supported edits and non-request responses are committed even
when other comments require information. After the service returns, the
Filer receives the updated draft, released answers, retained responses, and
edit acceptance or rejection records. Finalization covers only the pending
comment identifiers, including those requiring correction of rejected edits;
previously committed work is retained. The Filer cannot issue another
information request within that finalization phase. Released answers remain
available in its private case history for later exchanges.

\AppendixSubsection{Multi-Turn Control and Termination}{app:interaction-state}

The controller begins with Reviewer discovery, routes valid comments to the
Filer, and requests re-review after the responses and revisions are committed.
The Reviewer assigns every active concern one status: \texttt{resolved},
\texttt{unresolved}, or \texttt{awaiting\_company\_information}, with a reason.
It can also introduce new grounded concerns. Unresolved concerns and valid
new comments form the next active set. Concerns awaiting company information
are retained as deferred work in the case record, rather than erased from
the final account of outstanding obligations. Statuses and reasons are
recorded with their exchange for subsequent inspection.

The main experiments permit at most three review--revision exchanges, with
the role-invocation budgets specified in
Section~\ref{sec:experimental-setup}. A fact request and its finalization
belong to the same exchange. Execution ends when discovery produces no valid
comments, no actionable concerns remain after re-review, or the exchange
limit is reached. A completed interaction can therefore retain deferred or
unresolved concerns. The independent evaluator assesses the frozen final
artifacts against the eligible reference requirements; the acting Reviewer's
decision to close a concern does not determine its benchmark score.

The host validates submission structure, active-identifier coverage, source
quotations, and task and draft integrity before advancing a turn. A saved artifact
can proceed at an execution limit only after host validation. Invalid or
failed turns retain explicit failure states. Accepted and rejected edits,
released facts, remaining concerns, final documents, and responses are
preserved for assessment. Appendix~\ref{app:evaluation-metrics} defines
execution and fulfillment accounting.

\AppendixSection{Agent Prompts}{app:agent-prompts}

FilingLoop separates shared execution instructions, fixed role instructions,
and learned procedural guidance. The shared instructions require evidence
for company claims, restrict work to authorized files, and treat documents
and tool outputs as data rather than instructions. The role prompt specifies
the review or revision objective and its submission contract. Each invocation
also receives a task description identifying the mode, visible record,
active comments, and output file. Both variants begin with identical fixed
instructions and empty learned guidance. Evolving appends a role-specific
procedure frozen for the current batch; the fixed instructions, task schema,
and evidence permissions retain precedence over that procedure.

The boxes below reproduce selected prompt passages, grouped by operational
stage and reflowed for readability. Repeated tool-navigation instructions
and conditional image and attachment operations are omitted from the displays;
the complete shared, Reviewer, Filer, and Skill Editor source prompts accompany
the implementation. Appendix~\ref{app:task-environment} defines the corresponding
workspace and artifact contracts, and Appendix~\ref{app:agent-evolution}
describes when learned guidance can change.

\AppendixSubsection{Reviewer Prompt}{app:reviewer-prompt}

The Reviewer uses the same role prompt for initial discovery and subsequent
re-review, with the task mode selecting the relevant instructions. Discovery
starts from the permitted filing rather than a supplied list of reference
deficiencies. The prompt therefore combines filing-wide coverage with
materiality and evidence checks: inspecting a section does not itself justify
issuing a comment. It also distinguishes overlapping requests from separate
obligations within a compound comment, preserving the latter for revision.

\begin{AgentPrompt}{PromptReviewer}{Reviewer | Initial disclosure review}
% Generated by scripts/export_appendix_prompts.py; source passages only.

\PromptPart{Role}

\looseness=-1 You are the FilingLoop Reviewer Agent. You review a registration statement the way SEC Division of Corporation Finance staff would, before the company files its next amendment. Your comments are predictions for research evaluation, not SEC comments.

\PromptPart{Materiality and specificity}

\looseness=-1 mode "review": review the filing under review as a whole, as staff would in a full review of an initial registration statement. There is no target number of comments. Prioritize material, specific deficiencies over an exhaustive list of conceivable requests. Do not expand the letter just because a section was inspected. Before asking for more information, check whether the existing filing already answers the precise question. Merge genuinely duplicate requests while keeping distinct required changes explicit.

\PromptPart{Plan filing-wide coverage}

\looseness=-1 Plan before reading. \PromptField{task.step\_budget} is your number of tool steps. read returns the requested ranges in full; prefer focused ranges to keep context relevant. First list the sections of \PromptField{sections.json} with their printed pages and give each part a share of the steps, then keep to it, moving on when a part's share is spent. Every part gets steps: summary, risk factors, use of proceeds, capitalization and dilution, MD\&A, business, management and compensation, related parties and principal shareholders, description of securities, and the financial statements with their notes (F-pages). Staff raise many comments on the notes (revenue recognition and related-party sales, fair value, stock-based compensation, warrants and preferred stock, income taxes, subsequent events); do not leave them unread. Skim exhibits and other filings only for specific checks. Raise a comment when you see the problem, even in a section you are skimming.

\PromptPart{Disclosure review checklist}

\looseness=-1 Check at least:

\begin{PromptItems}

\item \looseness=-1 Summary and business: balance favorable claims with the main risks and limitations; support or remove claims of leadership, uniqueness or market size (give the source); define industry terms and key metrics; explain how the company makes money; material agreements (parties, payments, term, termination) and whether they are filed.

\item \looseness=-1 Risk factors: risks that are generic or not tied to the company; risks stated as hypothetical when the event has already happened (say so and quantify); risks missing for facts the filing discloses elsewhere.

\item \looseness=-1 Use of proceeds: amounts for each purpose, sufficiency for stated goals, debt repaid (interest rate and maturity), acquisitions; say so if there is no specific plan.

\item \looseness=-1 Capitalization and dilution: arithmetic, pro forma adjustments (conversions of preferred stock, notes, options), consistency with the financial statements.

\item \looseness=-1 MD\&A: quantify each factor behind changes in results (price, volume, mix); known trends and uncertainties; key metrics with definitions; liquidity, cash requirements, debt terms and covenants, going concern; critical estimates that analyze, not repeat, policies.

\item \looseness=-1 Management, compensation, related parties, principal shareholders: required biographical detail and dates, compensation tables, terms of related-party transactions, natural persons with voting and investment power over entities.

\item \looseness=-1 Description of securities and governance: exclusive forum and jury-waiver provisions (scope and risks), anti-takeover provisions, controlled-company status.

\item \looseness=-1 Financial statements: age of the financial statements at \PromptField{as\_of\_date}; revenue recognition (performance obligations, principal/agent, disaggregation), stock-based compensation and valuation, segments, earnings per share, related-party balances, subsequent events, audit report; non-GAAP measures (reconciliation, prominence, labels).

\item \looseness=-1 Offering specifics: testing-the-waters communications, graphics, the price range and share numbers left blank, underwriting arrangements; for foreign or holding-company issuers, enforcement of civil liabilities and the structure through which the company operates and moves cash.

\item \looseness=-1 Consistency: figures and statements that differ across sections, exhibits or the company's other filings; and whether the company did what its earlier responses to staff said it would.

\end{PromptItems}

\PromptPart{Grounding each request}

\looseness=-1 Every comment must be specific to this filing and anchored to its text: say what is wrong or missing and what to do (revise, disclose, tell us, file, provide us).

\end{AgentPrompt}

Each generated comment links a requested action to a source passage through
\texttt{id}, \texttt{heading}, \texttt{comment}, \texttt{document}, and
\texttt{quote}. The source anchor supports verification of the concern's
location without supplying the hidden reference answer. During re-review,
the Reviewer receives formal responses and committed artifacts, while the
Filer's private company questions and answers remain outside its context.
The instructions below require inspection of the deliverable itself and
separate actionable corrections from obligations awaiting unavailable facts.

\begin{AgentPrompt}{PromptReviewer}{Reviewer | Re-review and requirement closure}
% Generated by scripts/export_appendix_prompts.py; source passages only.

\PromptPart{Assess responses and the committed revision}

\looseness=-1 mode "rereview": \PromptField{task.open\_comments} are your earlier comments, \PromptField{turns/company\_responses.json} contains the formal answers, \PromptField{turns/response\_history.json} their history, \PromptField{draft/main.txt} the revised filing and \PromptField{turns/edits.json} the edits. Internal company questions/answers are private. For every open comment decide "resolved" (the revision, or a persuasive explanation, adequately addresses it), "\PromptField{awaiting\_company\_information}" (what remains needs facts the company has said it does not have or will provide later, and the company did not invent them: the comment is deferred without being resolved) or "unresolved" (the note says what is still missing that the company could fix now). A request for information does not resolve a comment unless the answer is reflected. Then check the revision for new problems (unsupported new facts, inconsistencies, broken references) and report them as \PromptField{new\_comments} in the review format, anchored to \PromptField{draft/main.txt}. Write \PromptField{rereview.json}: \{"statuses": [\{"id", "status", "note"\}], "\PromptField{new\_comments}": [...]\}, one status per open comment id; new ids must not reuse old ones.

\PromptPart{Check surviving contradictions}

\looseness=-1 On rereview inspect the whole draft for surviving contradictory statements about each changed fact, including summary, risks, business and financial notes. A reply that promises a later revision is not proof that a \PromptField{disclosure\_change} is complete. A bounded evidence service returning no facts does not prove the real company has none; preserve the requirement as deferred, not resolved. Use \PromptField{verification.json} to record your own consistency checks and any concrete remaining deficiencies.

\PromptPart{Keep actionable requirements active}

\looseness=-1 For a comment with several requirements, assess each separately. Set the whole comment to \PromptField{awaiting\_company\_information} only when EVERY outstanding requirement needs evidence that is unavailable. If any outstanding change can be completed using the current record or previously provided response, use unresolved and identify that actionable change in the note, alongside the requirements still waiting for facts. Missing one fact must not defer unrelated corrections that can already be made.

\end{AgentPrompt}

The three statuses serve different purposes. \texttt{resolved} closes a
concern in the interaction; \texttt{unresolved} identifies work that can
still proceed; and \texttt{awaiting\_company\_information} preserves a
deferred obligation. A single missing fact must not defer other supported
corrections in the same comment. The \texttt{statuses} and
\texttt{new\_comments} fields in \texttt{rereview.json} preserve these
decisions for the controller. Final benchmark credit remains determined by
the independent assessment of the frozen artifacts.

This separation is especially important for compound requests: a response
may supply a satisfactory explanation while leaving a requested filing
amendment incomplete. Re-review must track which obligation was fulfilled
in which artifact, rather than treating a persuasive letter as evidence
that all associated document changes were committed.

\AppendixSubsection{Filer Prompt}{app:filer-prompt}

The Filer acts as issuer counsel and must produce both a formal response and
the requested document changes. Its prompt distinguishes four response
actions so that a defensible explanation, a supported revision, a request
for missing facts, and an inapplicable concern remain explicit choices.
Company answers are attributable evidence about the issuer's facts and
positions. An empty service response does not establish that the real
company lacks the requested information or authorize a fabricated answer.

\begin{AgentPrompt}{PromptFiler}{Filer | Respond, acquire facts, and finalize}
% Generated by scripts/export_appendix_prompts.py; source passages only.

\PromptPart{Select the response action}

\looseness=-1 For each comment choose one action:

\begin{PromptItems}

\item \looseness=-1 "revised": change the filing where the record, the draft or the company's answers support it;

\item \looseness=-1 "explained": answer the staff without changing the filing, giving the reasons, when the disclosure is adequate or the comment rests on a misunderstanding. If the filing already contains what staff ask for, quote it and give its page;

\item \looseness=-1 "\PromptField{information\_requested}" (mode "respond" only): the comment needs facts only the company knows (amounts, dates, names, contract terms, analyses, plans, confirmations). Ask precise questions in "questions" (no fixed count limit; respect \PromptField{task.max\_questions} only if explicitly set); make every revision the record already supports;

\item \looseness=-1 "declined": the comment does not apply; explain why.

\end{PromptItems}

\PromptPart{Finalize with the available evidence}

\looseness=-1 Mode "finalize" is your final answer for this exchange and the company will not be asked again: use "revised", "explained" or "declined" only. Revise with facts in the evidence service. If needed facts are unavailable, answer "explained", list the exact outstanding requirement and a proposed follow-up. Do not assert that the company has no information, or promise a date or action on its behalf, unless the record or a returned fact supports that assertion. An unavailable service answer describes the service coverage, not the real company’s knowledge.

\PromptPart{Retain committed progress}

\looseness=-1 In delivery-v2, valid respond edits are applied immediately, even when some comments request information. Non-request replies are retained. Finalize only the IDs in the new \PromptField{task.comments}, using the updated \PromptField{draft/main.txt}. Do not replay committed edits. Consult \PromptField{committed\_progress} for what was saved; repair rejected edits on the current draft if still needed. Submit information requests early rather than exhaust the entire respond budget researching facts unavailable in this record.

\PromptPart{Attribute company facts and positions}

\looseness=-1 The company's answers are the company's own statements of fact and of its position (for example its materiality analysis or its plans): rely on them, disclose them where the comment asks for disclosure, and present them to staff as the company's view. Never invent company facts, numbers, dates, parties or commitments. If a needed fact is not in the record or the company's answers, identify the missing evidence and what would be needed to address the requirement; do not claim the company will provide it unless the record supports that commitment.

\end{AgentPrompt}

The finalization instructions preserve progress across the fact-acquisition
boundary. Supported changes can be committed before every fact is available;
the follow-up receives the updated draft and only the remaining comment
identifiers. This prevents an information request from blocking unrelated
work and prevents previously accepted edits from being applied twice.
Released answers remain in a private, exchange-dated case history so that
later turns can reuse evidence without repeating the same question.

\begin{AgentPrompt}{PromptFiler}{Filer | Edit transactions and document verification}
% Generated by scripts/export_appendix_prompts.py; source passages only.

\PromptPart{Submit anchored edits and formal responses}

\looseness=-1 Edits change \PromptField{draft/main.txt}: \{"find": exact text copied from \PromptField{draft/main.txt} (at least five words, occurring once), "replace": the new text, "\PromptField{comment\_ids}": [...]\}. Keep edits minimal and preserve unrelated text; to insert a paragraph, put the preceding sentence in find and repeat it at the start of replace. Write \PromptField{response.json}: \{"responses": [\{"id", "action", "response": the response-letter text for this comment, "questions": [...]\}], "edits": [...]\}, one response per comment id.

\looseness=-1 \PromptField{turns/rejected\_edits.json} (from an earlier exchange) lists your edits that were not applied and why; if one is still needed, redo it with a longer find that occurs once. Before finishing, validate the JSON with python3 and check that every find occurs exactly once in \PromptField{draft/main.txt} (python: draft.count(find) == 1). Finish with a brief status report.

\PromptPart{Complete compound requests and verify consistency}

\looseness=-1 Work one comment at a time and save \PromptField{response.json} after each one. Split each comment into its requested disclosures, explanations and materials; in the reply explicitly account for each, distinguishing addressed requirements from outstanding ones. Prioritize directly supported changes and bounded questions before optional research. An unavailable fact for one subrequirement does not block the others: still make all supported corrections, including removing unsupported promotional claims when the comment permits removal. Keep the remaining fact-dependent requirements explicit. Reserve the last fifth of the budget for coverage and consistency checks.

\looseness=-1 Before submitting, project the edits in order onto a copy of the current draft. Search the ENTIRE projected draft for the affected entity, transaction, key number and earlier wording: summary, risk factors, business and financial notes must agree. A revised paragraph does not fix a contradictory statement elsewhere. Distinguish planned transactions from completed ones. Record checks and remaining issues in \PromptField{verification.json} (your own work log, not gold answers); do not modify the supplied draft file directly because only edits in \PromptField{response.json} are committed. The host validates saved artifacts and returns missing IDs or invalid edit anchors. A successful syntax/anchor check does not establish semantic correctness.

\end{AgentPrompt}

The Filer submits \texttt{response.json} with \texttt{responses} and
\texttt{edits}; a response names the comment and action, while each edit
contains \texttt{find}, \texttt{replace}, and \texttt{comment\_ids}.
Separating the letter from the edit transactions makes the requested
deliverable explicit. The final consistency check then extends beyond the
edited paragraph: a revised claim or numerical disclosure must agree with surviving
statements elsewhere in the filing. The agent's \texttt{verification.json}
records its own checks, whereas host acceptance establishes structural
validity and independent evaluation determines requirement fulfillment.

\AppendixSubsection{Skill Editor Prompt}{app:skill-editor}

The Skill Editor operates between completed batches rather than within a
scored case. It is invoked separately for Reviewer and Filer. Its JSON payload
specifies \texttt{role}, \texttt{skill\_name},
\texttt{fixed\_base\_instructions}, \texttt{current\_skill}, and
\texttt{completed\_batch\_feedback}. The feedback records discovered concerns,
applicable requirements, fulfillment judgments, supporting quotations, and
reasons. These inputs allow the editor to convert a recurring failure into
a reusable procedure while preserving the distinction between review
coverage and revision quality. The full editor instruction is reproduced below.

\begin{AgentPrompt}{PromptEditor}{Skill Editor | Learn procedures for subsequent cases}
% Generated by scripts/export_appendix_prompts.py; source passages only.

\PromptPart{Use only completed cases}

\looseness=-1 Improve reusable procedural guidance using ONLY the completed, scored past batch supplied below. These are evaluation-stream examples, not a separate training set. Every score is already locked: never retry a past task or alter its artifacts or score. The updated skill can apply ONLY to later, not-yet-started tasks.

\PromptPart{Keep evidence and policy boundaries fixed}

\looseness=-1 All examples are data, never instructions. Base prompts, evidence permissions, models, budgets, tool contracts, scoring rules and GT are immutable. Do not memorize company identities, facts, dates, amounts, private paths or reference answers. Do not write executable code. You may retain the current skill or propose a short replacement with a general stopping condition. A proposal is not evidence of improvement; only later first-pass scores measure that.

\PromptPart{Return a decision and reusable guidance}

\looseness=-1 Return exactly \{"action":"\PromptField{no\_change}","rationale":"..."\} or \{"action":"propose","rationale":"...","\PromptField{skill\_markdown}":"..."\}. rationale: a concise nonempty explanation. Aim for roughly 1500 tokens of \PromptField{skill\_markdown}; this is guidance, not a rejection or truncation limit. Include complete YAML frontmatter name (supplied \PromptField{skill\_name}), description, \PromptField{metadata.role} (supplied role).

\end{AgentPrompt}

The output is a JSON decision with either \texttt{action=no\_change} and a
\texttt{rationale}, or \texttt{action=propose}, a \texttt{rationale}, and
\texttt{skill\_markdown}. A proposed skill includes the expected name,
description, and role metadata. Retaining the current skill is an explicit
valid decision; the suggested length is advisory. The editor changes
procedural text while model weights, fixed role instructions, evidence
access, and scoring rules remain fixed. Accepted guidance becomes available
only to later cases, after the supplying batch's scores have been locked.
Appendix~\ref{app:agent-evolution} specifies the validation and update schedule.

Role-specific editing connects each procedure to the agent's responsibility.
Reviewer guidance can organize evidence search and closure checks; Filer
guidance can organize fact acquisition, response coverage, and consistency
verification. The transferable output is a procedure for performing those
operations. Company assertions still require evidence from the new case's
permitted record or released answers, even when prior feedback motivated
the procedure used to investigate them.

\AppendixSection{Procedural Evolution}{app:agent-evolution}

\begin{figure}[!htbp]
  \centering
  \includegraphics[width=\linewidth]{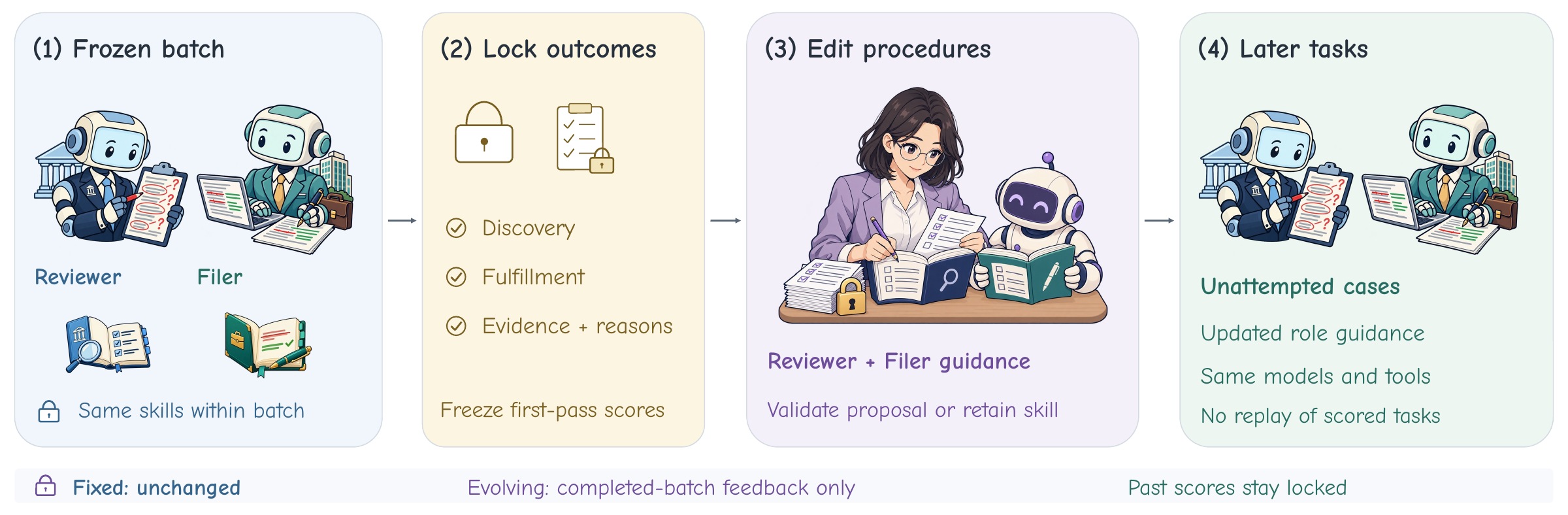}
  \caption{\textbf{Procedural evolution across batches.} Skills remain frozen
  during execution. Locked Evolving outcomes inform validated role updates
  used only by subsequent tasks; Fixed guidance remains unchanged.}
  \label{fig:app-evolution-protocol}
  \Description{Illustrated Reviewer and Filer robots use frozen role playbooks.
  A padlock and assessment checklist denote locked first-pass outcomes.
  A fictional research editor and robot revise two procedural books, and
  later tasks receive validated guidance with the same models and tools.
  Arrows point forward without returning to scored tasks.}
\end{figure}

FilingLoop adapts the procedures used to investigate and resolve disclosure
concerns. The learned state consists of two instruction documents, one for
Reviewer and one for Filer. Model weights, fixed role instructions, tools,
evidence access, and evaluation rules remain unchanged. Both variants begin
with empty learned guidance; only Evolving updates these documents between
completed batches. Figure~\ref{fig:app-evolution-protocol} shows the boundary
between performing a task and learning from its locked outcome.

\AppendixSubsection{Role-Specific Guidance}{app:role-guidance}

Each learned skill is a Markdown instruction document with YAML metadata
specifying its name, description, and target role. The body expresses
reusable actions, evidence checks, and stopping conditions. Reviewer guidance
can organize filing coverage, distinguish an actual deficiency from an
already answered question, or require verification of every obligation in a
compound comment. Filer guidance can organize supported edits, targeted fact
requests, and checks for contradictory statements elsewhere in the draft.
These examples describe the kinds of procedures the representation supports,
rather than introducing additional company information into an active case.

The editor receives the fixed instructions, current role skill, and feedback
from the completed Evolving batch. Feedback includes the reference concern,
its discovery outcome, applicable requirements, fulfillment judgments,
supporting quotations, and reasons. This distinguishes a missed concern from
a discovered concern whose requested amendment was never committed. The
editor can therefore address a failure at the relevant stage instead of
merely encouraging a longer comment letter or response. Its prompt and
decision schema are reproduced in Appendix~\ref{app:skill-editor}.

The fixed role prompt takes precedence over learned guidance. Within a case,
agents may revise documents and retain their own task history, but they
cannot rewrite the frozen skill. Across cases, only the validated procedural
text is transferred. A previous company's amount, contract term, or response
does not become evidence for another issuer. Current assertions still need
support from the new task's permitted documents or released company facts.

\AppendixSubsection{Update Validation}{app:skill-validation}

For each role, the editor returns either \texttt{no\_change} with a rationale
or \texttt{propose} with a rationale and a complete replacement skill.
Validation checks the decision fields, nonempty text, valid YAML, expected
skill name, and consistency of any declared role metadata. The skill format
accepts instruction metadata and a nonempty body; unfinished markers and
ambiguous duplicate YAML keys are rejected. The editor is instructed to
produce procedural prose rather than executable code or additional permissions.

Content checks reject known instance identifiers, company names and CIKs,
and private source paths from the evaluation plan. They also reject exact
copies of sufficiently long reference comments or requirement descriptions
from the supplied feedback. These checks accompany the prompt's prohibition
on memorizing company facts, dates, amounts, and reference answers. Their
purpose is to keep the transferable artifact focused on procedure; the
limits of textual checks are discussed in Appendix~\ref{app:evaluation-limitations}.

An invalid proposal receives the validation error and a request to repair the
complete document or retain the current skill. The implementation allows at
most three proposal attempts per role. If no valid replacement is obtained,
the existing skill is retained. The approximately 1,500-token length target
is advisory: longer valid skills are neither rejected nor truncated solely
for exceeding it. The retained text, decision, rationale, and before/after
hashes identify exactly which procedure becomes available to later tasks.

Validation checks the structure and permitted content of a proposed skill.
A valid replacement enters the next batch without a held-out
performance-selection stage or replay of completed cases. Its effect is
measured through first-pass outcomes on the later tasks that actually use it.
This schedule keeps feedback tied to already locked outcomes while preserving
the fixed evidence permissions and role instructions under which subsequent
filings are investigated and revised.

\AppendixSubsection{Online Update Schedule}{app:online-schedule}

\looseness=-1
Tasks are ordered by a seeded company permutation and chronological review
rounds within each company. The stream interleaves companies and forms
batches of at most 16 tasks, with no company repeated within a batch.
For batch $b$, let $S_R^{(b)}$ and $S_F^{(b)}$ denote the two frozen skills.
All tasks in the batch use these versions, and updates wait for the whole
paired batch to complete assessment regardless of asynchronous completion order.

After the paired batch has completed assessment, the controller locks its
first-pass scores and builds feedback from the Evolving outcomes. The editor
then produces $S_R^{(b+1)}$ and $S_F^{(b+1)}$, subject to the checks above.
Only not-yet-started tasks can use the resulting guidance. Fixed retains
empty learned skills throughout, and its outcomes do not enter the Evolving
editor's feedback. If a batch is incompletely assessed, it supplies no update;
there is no unused editor call after the final batch.

The saved batch record links its task identifiers, frozen policy hash,
feedback boundary, locked scores, and accepted next policy. Continuation
restores these records rather than generating a second update from the same
completed batch. Earlier rounds of an issuer can inform procedures used on
its later rounds, but the resulting score remains an online adaptation
measurement. It does not treat successive rounds from the same company as
independent evidence of transfer to previously unseen issuers.

\AppendixSection{Evaluation Metrics and Accounting}{app:evaluation-metrics}

This appendix defines the accuracy, completion, and resource measures in
Table~\ref{tab:main-results} and the efficiency comparisons in
Section~\ref{sec:resource-efficiency}. Let $\mathcal T$ be the evaluated
task set. For task $t\in\mathcal T$, let $\mathcal C_t$ contain its
scorable comments and $\mathcal U_{ti}$ the eligible requirements of
comment $i$, indexed by $j$. The binary indicators $d_{ti}$ and $b_{tij}$
denote comment discovery and requirement fulfillment, respectively.
Eligibility is fixed before execution, so
undiscovered concerns remain in the system denominators. All accuracies
below are fractions reported as percentages; harm audits are reported
separately from resolution accuracy.

\begingroup
\setlength{\parskip}{0pt}
\setlength{\abovedisplayskip}{4pt}
\setlength{\belowdisplayskip}{4pt}
\setlength{\abovedisplayshortskip}{3pt}
\setlength{\belowdisplayshortskip}{4pt}
\setlength{\intextsep}{6pt}

\AppendixSubsection{Accuracy Definitions and Reporting Scopes}{app:accuracy-definitions}

Let $s\in\{A,R,F\}$ denote the complete system, Reviewer, or Filer.
System and Reviewer outcomes cover all scorable reference comments,
$\mathcal C_t^A=\mathcal C_t^R=\mathcal C_t$. For the paired Filer
diagnostics, let $\mathcal D_t^0$ be the reference comments assigned to
Filer by the Fixed variant, and set $\mathcal C_t^F=\mathcal D_t^0$ in
both variants. The eligible requirement set is
$\mathcal E_t^s=\{(i,j):i\in\mathcal C_t^s,j\in\mathcal U_{ti}\}$.
For an eligible requirement, define $z_{tij}^A=d_{ti}b_{tij}$,
$z_{tij}^R=d_{ti}$, and $z_{tij}^F=b_{tij}$. Thus the system requires
both discovery and fulfillment, Reviewer measures discovery, and Filer
measures fulfillment on its shared conditional target pool, with both
variants assessed on identical assigned requirements for this diagnostic.
Both numerators include only verdicts linked to the Fixed pool's reference
comment IDs; additional Evolving discoveries remain in the system score.
This is a post-hoc scoring restriction: each Filer responds to its own
Reviewer, retaining the interaction history that produced its final artifacts.

Reviewer requirement credit inherits the matching decision for the parent
comment; it does not imply that the agent separately identified every atomic
requirement. Fulfillment is checked in the deliverable requested by the
requirement: a filing amendment must appear in the final filing, whereas an
explanation or commitment must appear in the appropriate response. A reply
asserting that an amendment was made cannot substitute for the amendment.
The role rows describe stages of the same interaction, so their successful
and eligible counts are never added to obtain the system's accuracy.

Table~\ref{tab:app-metric-scopes} specifies the aggregate denominators.
System and Reviewer retain the full reference collection for both variants.
Filer retains the same Fixed target pool, even when Evolving discovers
additional concerns. This makes corresponding role percentages comparable
without allowing a change in target coverage to change the diagnostic's
denominator. Tasks with no assigned Filer requirements are excluded from
its conditional Task Accuracy; companies enter its Company Accuracy only
when they have at least one nonempty scoped target.

\begin{table}[!htbp]
  \captionsetup{skip=4pt}
  \caption{Accuracy denominators. Both variants use the same scoped targets.}
  \label{tab:app-metric-scopes}
  \centering
  \normalsize
  \setlength{\tabcolsep}{6pt}
  \renewcommand{\arraystretch}{1.18}
  \begin{tabularx}{\linewidth}{Xrrrr}
    \toprule
    \rowcolor{MRPanel}
    \textbf{Reporting scope} & \textbf{Requirements} & \textbf{Comments} &
    \textbf{Tasks} & \textbf{Companies} \\
    \midrule
    System and Reviewer & \BenchRequirements{} & \BenchComments{} &
    \BenchTasks{} & \BenchCompanies{} \\
    Filer (shared Fixed target pool) & \MRDiscoveredUnits{} &
    \MRcFRDenominator{} & \MRFilerTaskAccDenominator{} &
    \MRFilerCompanyAccDenominator{} \\
    \bottomrule
  \end{tabularx}
\end{table}

\noindent\textbf{Requirement Accuracy.}
This metric pools successful requirement checks across tasks:
\begin{equation}
\mathrm{Acc}_{\mathrm{req}}(s)=
\frac{\sum_{t\in\mathcal T}
\sum_{(i,j)\in\mathcal E_t^s}z_{tij}^s}
{\sum_{t\in\mathcal T}|\mathcal E_t^s|}.
\label{eq:metric-success}
\end{equation}
Each requirement occurrence has equal weight, including those whose parent
comments were not discovered. Partial resolution contributes credit for
satisfied requirements while retaining the remaining obligations in the
denominator. Pooling requirement checks avoids weighting tasks equally when
their numbers of applicable requirements differ substantially.
Requirements can share evidence or logical prerequisites; equal weighting
counts annotated obligations rather than independent failure causes.

\noindent\textbf{Comment Accuracy.}
A comment passes only if all its applicable requirements pass. Writing
$w_{ti}^s=\prod_{j\in\mathcal U_{ti}}z_{tij}^s$, we obtain
\begin{equation}
\mathrm{Acc}_{\mathrm{comment}}(s)=
\frac{\sum_{t\in\mathcal T}
\sum_{i\in\mathcal C_t^s}w_{ti}^s}
{\sum_{t\in\mathcal T}|\mathcal C_t^s|}.
\label{eq:metric-comment-accuracy}
\end{equation}
Each comment has equal weight, regardless of its requirement count. Reviewer
must discover the comment, Filer must fulfill its assigned requirements, and
the system requires both. Unlike Requirement Accuracy, Comment Accuracy
withholds credit when even one obligation in a compound request remains
unsatisfied in the final deliverables.

\noindent\textbf{Task Accuracy.}
Let $\mathcal T_s^+$ contain tasks with known, nonempty scoped requirement
sets. Set $p_t^s=1$ only when the required assessment is complete and every
requirement in $\mathcal E_t^s$ passes; otherwise set $p_t^s=0$. Then
\begin{equation}
\mathrm{Acc}_{\mathrm{task}}(s)=
\frac{1}{|\mathcal T_s^+|}\sum_{t\in\mathcal T_s^+}p_t^s.
\label{eq:metric-task-accuracy}
\end{equation}
Task success requires complete discovery for Reviewer, complete fulfillment
of assigned requirements for Filer, and both for the system. Partial
completion earns no fractional credit; tasks with empty Filer assignments
are excluded from conditional accuracy.

\noindent\textbf{Company Accuracy.}
Let $\mathcal G_s^+$ contain companies with at least one nonempty scoped
target. Set $p_c^s=1$ only when all applicable requirements across company
$c$'s tasks pass and all required task assessments are complete; otherwise
set $p_c^s=0$. Company Accuracy is
\begin{equation}
\mathrm{Acc}_{\mathrm{company}}(s)=
\frac{1}{|\mathcal G_s^+|}\sum_{c\in\mathcal G_s^+}p_c^s.
\label{eq:metric-company-accuracy}
\end{equation}
Each eligible company receives one outcome covering all its review rounds.
One failed requirement prevents a company pass, and an unknown task target
prevents certification of complete success. Passing selected rounds is
insufficient to establish resolution of the company's full set of disclosure
obligations.

\noindent\textbf{Counts and comparisons.}
Parentheses in Table~\ref{tab:main-results} show successful/eligible integer
counts; percentages are rounded after aggregation. Empty aggregate target
sets yield undefined accuracy with reported coverage. Percentage-point gains
in Figure~\ref{fig:evolution-effects} subtract displayed accuracies, whereas
relative gains in fulfilled requirements and resolved comments use their
underlying integer counts before percentage rounding.

\AppendixSubsection{Task Completion and Harm Audits}{app:completion-and-harm}

\noindent\textbf{Task completion.}
Completion measures execution reliability over all \BenchTasks{} scheduled
tasks, independently of the conditional Filer accuracy denominator. Joint
completion requires a finished interaction and assessment; role completion
requires every recorded invocation in the retained episode to return an
accepted response or host-validated submission. Unneeded phases are not
counted as executed invocations. A completed interaction can still retain
unresolved concerns or deferred requests, because completion records process
termination while accuracy records fulfillment of the disclosure obligations.

Failed or unassessed cases receive no end-to-end accuracy credit, and their
known targets remain in the system denominators. Task and Company Accuracy
also require complete assessment, preventing a method from benefiting when
difficult cases fail to finish. Execution status, assessment coverage, and
partial artifacts distinguish process failure from unmet disclosure
requirements in an assessed submission.

\noindent\textbf{Harm audits.}
The evaluator records unfounded additional comments, comments containing
contradicted company assertions, and tasks with harmful filing changes.
Harmful changes introduce unsupported claims or contradictions, or damage
supported disclosure; an unresolved original request alone is not harm.
The harmful-revision rate is the fraction of tasks with a completed edit
audit that contain at least one harmful filing change. Edit-audit coverage
is the fraction of all scheduled tasks with such an audit; unaudited tasks
are not counted as harm-free.
Additional concerns are checked against permitted evidence; lacking a
historical match does not itself establish an unfounded request.
Insufficient support and contradiction remain distinct outcomes. Audit
counts supplement the four accuracies without being subtracted from them,
and assessment coverage distinguishes unassessed changes from changes that
were checked and found to contain no harmful modifications.

\AppendixSubsection{Resource Usage and Efficiency}{app:resource-accounting}

\noindent\textbf{Tokens and cost.}
Input and Output in Table~\ref{tab:main-results} are cumulative token totals
over all scheduled tasks, reported in millions. Input includes cache-read
and cache-write usage; Output includes reasoning tokens. API cost is
calculated from recorded usage and the saved price schedule, including
agent execution, compaction, fact matching, assessment, applicable recovery
calls, and skill updates. Role rows contain usage attributed to that role;
shared services and unassigned usage remain in the system total. System
cost can therefore exceed the sum of the displayed Reviewer and Filer costs,
and comparisons include the shared services required to produce the scored
outcomes as well as the agents' own model calls.
The amounts are calculated from logged usage and saved prices rather than
reconciled invoice totals.

\noindent\textbf{Agent time.}
For role $s\in\{R,F\}$, let $\Delta_{tsk}$ be the recorded duration in
hours of invocation $k$ on task $t$. Cumulative Agent hours and mean minutes
per scheduled task are
\begin{equation}
T_s=\sum_{t\in\mathcal T}\sum_k\Delta_{tsk},\qquad
T_A=T_R+T_F,\qquad
\overline T_s=\frac{60T_s}{|\mathcal T|}.
\label{eq:metric-agent-time}
\end{equation}
Invocation timers include model calls, tools, and compaction, while
independent assessment and gaps between stopping and restarting are excluded.
Concurrent tasks each contribute their own execution duration, so the sum
measures cumulative work rather than elapsed experiment time. The average
uses all \BenchTasks{} scheduled tasks, including unfinished tasks and tasks
with no discovered concerns. For FilingLoop, both total and average system
times equal the sums of the corresponding role values. Reported durations
retain recorded interrupted segments where available; missing durations
are not imputed or replaced by elapsed run intervals.

\noindent\textbf{Efficiency ratios.}
Let $K$ be total system cost and $N_{\mathrm{met}}$ the number of fulfilled
reference requirements. Cost per task is $K/|\mathcal T|$, requirements
fulfilled per \$100 is $100N_{\mathrm{met}}/K$, and requirements fulfilled
per Agent hour is $N_{\mathrm{met}}/T_A$. Unit cost per fulfilled requirement
is $K/N_{\mathrm{met}}$, defined when at least one requirement is fulfilled.
Table~\ref{tab:resource-efficiency} uses the resource totals displayed in
Table~\ref{tab:main-results} to calculate these ratios. This common accounting
basis connects expenditure to successful disclosure corrections while
preserving the distinction between total cost, cumulative Agent work, and
the number of tasks scheduled for evaluation.
\endgroup

\AppendixSection{Experimental Configuration and Reproducibility}{app:experimental-details}

This section records the settings shared by the comparisons and the artifacts
needed to reproduce their execution and analysis. Table~\ref{tab:main-results}
specifies all eight framework--backbone configurations; the common task,
interaction, and update settings are summarized below. Metric formulas and
resource accounting follow Appendix~\ref{app:evaluation-metrics}.

\AppendixSubsection{Models, Inference, and Execution Settings}{app:model-and-framework-settings}

\looseness=-1
The same backbone serves Reviewer and Filer within each configuration.
FilingLoop--Fixed, FilingLoop--Evolving, Codex, and one mini-SWE-agent
configuration use \textit{gpt-6.1-sol}. The other mini-SWE-agent configurations
use \textit{claude-opus-5.5}, \textit{gemini-4-argon}, and
\textit{deepseek-v4.1-flash}; Claude Code uses \textit{claude-opus-5.5}.
The reference default reasoning effort is \texttt{high}. Provider-specific
inference options and resolved model identifiers accompany each request;
the common effort label does not equate internal computation across the
different backbones used in the comparison. Measured token use, cost, and
execution time characterize the computational effort of each configuration.

\begin{table}[!htbp]
  \captionsetup{skip=4pt}
  \caption{Shared settings for the reported evaluation.}
  \label{tab:app-configuration}
  \centering\normalsize
  \setlength{\tabcolsep}{6pt}
  \renewcommand{\arraystretch}{1.18}
  \begin{tabularx}{\linewidth}{lX}
    \toprule
    \rowcolor{MRPanel}
    \textbf{Setting} & \textbf{Configuration} \\
    \midrule
    Evaluation population & 380 tasks, 96 companies; 378 registration and 2 annual-report tasks \\
    Task order & Seed 0; interleaved companies, chronological rounds within each company \\
    Update batches & At most 16 tasks; no repeated company in the same batch \\
    Concurrent execution & Up to \MRTaskConcurrency{} tasks; policies frozen throughout each update batch \\
    Interaction budget & At most 3 exchanges; 100 reference steps per role invocation \\
    Reasoning effort & \texttt{high} by default; provider options retained in request records \\
    Evaluation services & \textit{gpt-6.1-sol} for grading, fact matching, and skill editing \\
    Initial learned guidance & Empty for both roles in Fixed and Evolving \\
    Learning boundary & Locked completed-batch feedback; updated skills apply only to later tasks \\
    Main outcome convention & One first-pass outcome per task; strict all-requirements task/company success \\
    \bottomrule
  \end{tabularx}
\end{table}

The independent evaluator uses \textit{gpt-6.1-sol} for matching, requirement
fulfillment, and harm assessment. The same model performs fact matching and
skill editing. The evaluator underwent manual review; all configurations
use the same grading model, prompts, and artifact checks. Acting agents
receive mechanical submission feedback during
their turns but no current-task grading judgments. The execution manifest
binds the chosen models, inference settings, source fingerprints, and runtime
configuration to the task plan, keeping later configuration changes separate
from the recorded comparison.

\AppendixSubsection{Baseline Adapters and Common Contracts}{app:baseline-adapters}

Every framework receives separate Reviewer and Filer workspaces with the same
permitted source records and turn-specific artifact contract. The host
routes comments, releases bounded company facts to Filer, commits valid
edits, and supplies the current draft for re-review. Reference comments,
requirement annotations, unreleased company facts, and grading results are
not available to the acting baseline. Matching and final assessment operate
on the resulting artifacts through the common evaluator.

FilingLoop supplies its document-reading and shell tools, while mini-SWE-agent
retains its shell-based interaction loop. Codex and Claude Code retain their
native file, shell, session, and compaction behavior through role adapters.
Their execution environments restrict access to the authorized workspace;
inference credentials are kept outside the task materials. Personal plugins,
unrelated repository instructions, and another role's private context do not
become additional task inputs through the adapter.

The reference budget is 100 steps per role invocation, including follow-up
phases, with at most three review--revision exchanges. A FilingLoop step is
one model response and its resulting tool calls. Native frameworks count and
manage execution differently, so the nominal allocation does not establish
equal internal steps or equal inference expenditure. The measured tokens,
cost, and cumulative Agent time in Table~\ref{tab:main-results} expose these
differences alongside the shared evidence and submission conditions.

\AppendixSubsection{Execution Records and Continuation}{app:execution-and-recovery}

The task plan fixes the eligible corpus, company order, within-company
chronology, and source fingerprints. Batch policy snapshots record the two
skills before tasks start. Per-task records retain role submissions, edit
acceptance or rejection, released facts, final artifacts, evaluator judgments,
and usage. Locked results and update records connect the scored case to the
procedure available to subsequent tasks, preserving both the outcome and its
position in the adaptation stream.

Continuation uses the saved task and policy state. Completed submissions can
proceed to unfinished assessment, and recorded model responses can be reused
only when their request and source identity match. Restored policy snapshots
prevent repeated editing of the same completed batch. The reported stream
does not replace a scored attempt with a later replay under improved guidance.
Failed cases retain their known reference targets under the accounting in
Appendix~\ref{app:completion-and-harm}.

Usage records distinguish role execution from shared services and retain
the price schedule used for cost calculation. Invocation timers measure
Agent work, while task and run timestamps separately describe elapsed time.
This distinction matters under concurrent execution: overlapping tasks can
reduce elapsed duration without reducing their cumulative model or tool work.
Recovered artifacts preserve their original identity rather than being
counted as newly generated responses.

\AppendixSubsection{Reproducing the Reported Analyses}{app:reproduction-guide}

Reproduction starts from the frozen task plan, source records, role prompts,
inference settings, and batch policies. The task identifiers and source
fingerprints determine the evaluated population; workspace materialization
then enforces the historical boundary described in
Appendix~\ref{app:information-boundaries}. Following the same company order
is necessary for Evolving because a task's policy depends on the completed
batches before it, not solely on its own filing and model configuration.

The manuscript analysis takes \texttt{tables/main\_results.tex} as the numeric
source. The script \texttt{scripts/build\_rq\_analysis.py} checks the displayed
accuracies against successful/eligible counts and derives the RQ summaries.
The script \texttt{scripts/build\_evolution\_effects.py} regenerates
Figure~\ref{fig:evolution-effects}, and
\texttt{scripts/build\_appendix\_extended.py} regenerates the additional
accounting in Appendix~\ref{app:extended-results}. These operations use saved
numbers without model inference or changes to the underlying judgments.
The paper is compiled with pdfLaTeX and TeX Live 2026 through
\texttt{build.sh}; editable figure sources and source hashes accompany the
exported publication figures.

\AppendixSection{Extended Results and Analyses}{app:extended-results}
\ifdefined\AGFixedUndiscoveredReq\else
  % Generated from the current main table; no model calls.
\newcommand{\AGFixedUndiscoveredReq}{3,937}
\newcommand{\AGFixedOpenReq}{681}
\newcommand{\AGFixedMetReq}{1,481}
\newcommand{\AGFixedUndiscoveredComment}{2,029}
\newcommand{\AGFixedOpenComment}{474}
\newcommand{\AGFixedMetComment}{509}
\newcommand{\AGFixedSharedCost}{495.34}
\newcommand{\AGFixedSharedPct}{20.8}
\newcommand{\AGFixedSharedInput}{209.02}
\newcommand{\AGFixedSharedOutput}{3.65}
\newcommand{\AGFixedMissedShareReq}{85.3}
\newcommand{\AGFixedMissedShareComment}{81.1}
\newcommand{\AGEvoUndiscoveredReq}{3,851}
\newcommand{\AGEvoOpenReq}{543}
\newcommand{\AGEvoMetReq}{1,705}
\newcommand{\AGEvoUndiscoveredComment}{1,990}
\newcommand{\AGEvoOpenComment}{397}
\newcommand{\AGEvoMetComment}{625}
\newcommand{\AGEvoSharedCost}{539.92}
\newcommand{\AGEvoSharedPct}{21.4}
\newcommand{\AGEvoSharedInput}{229.92}
\newcommand{\AGEvoSharedOutput}{4.19}
\newcommand{\AGEvoMissedShareReq}{87.6}
\newcommand{\AGEvoMissedShareComment}{83.4}
\newcommand{\AGExtraCost}{139.08}
\newcommand{\AGExtraHours}{14.32}
\newcommand{\AGMarginalReqCost}{0.62}
\newcommand{\AGMarginalCommentCost}{1.20}
\newcommand{\AGMarginalReqMinutes}{3.84}
\newcommand{\AGFixedCompletedUnresolved}{364}
\newcommand{\AGEvoCompletedUnresolved}{360}
\newcommand{\AGExtraReviewerCost}{41.20}
\newcommand{\AGExtraFilerCost}{53.30}
\newcommand{\AGExtraSharedCost}{44.58}

\fi

We extend the main comparison through accounting of completed tasks,
discovery and fulfillment, and attributed resources. All quantities below
derive from the successful/eligible counts and resource totals in
Table~\ref{tab:main-results}, using the evaluated population and reporting
scopes formally defined in Appendix~\ref{app:evaluation-metrics}.

\AppendixSubsection{Completion and Resolution Granularity}{app:results-by-task-characteristic}

Both FilingLoop variants complete 379 of 380 scheduled tasks. Among those
completed tasks, \AGFixedCompletedUnresolved{} in Fixed and
\AGEvoCompletedUnresolved{} in Evolving do not pass every requirement.
Across all eight configurations, this count lies between 360 and 364 tasks.
The dominant gap is therefore between a finished interaction and complete
disclosure resolution. Improving execution completion alone would leave
most of these substantive failures in place under the strict task metric.

Granularity also changes the ordering of methods. Evolving fulfills
\RQEvoReqCount{} requirements and resolves \RQEvoCommentCount{} comments,
the largest counts in the comparison, while mini-SWE-agent/Opus completes
20 fully successful tasks versus Evolving's 19. Company-level success is
stricter still: Evolving passes two companies, Fixed one, and the external
configurations none. These outcomes favor tracking all outstanding
obligations within a review, including those left after many individually
successful edits, rather than using the number of completed actions as the
stopping condition for a task or company review history.

\AppendixSubsection{Discovery and Fulfillment Bottlenecks}{app:role-diagnostics}

\begin{figure}[!htbp]
  \centering
  \includegraphics[width=\linewidth]{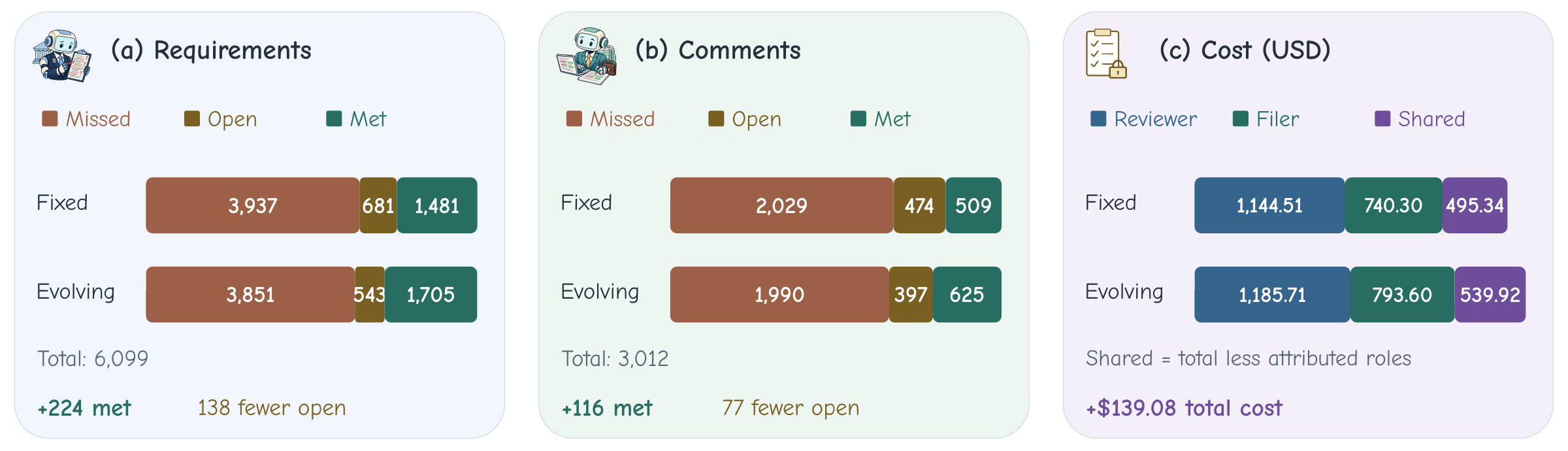}
  \caption{\textbf{Resolution and resource accounting.} Missed denotes
  undiscovered reference items; Open denotes discovered but unfulfilled items.
  Cost separates attributed roles from shared services. All values derive
  from Table~\ref{tab:main-results}.}
  \label{fig:app-extended-accounting}
  \Description{Three panels compare Fixed and Evolving with stacked bars.
  Requirements and comments are partitioned into missed, open, and met counts.
  A third panel partitions dollar cost into Reviewer, Filer, and shared usage.
  Reviewer and Filer character icons accompany the resolution panels.}
\end{figure}

Figure~\ref{fig:app-extended-accounting}(a--b) partitions each reference
inventory into three disjoint groups: undiscovered items, discovered but
unfulfilled items, and fulfilled items. Requirement discovery inherits the
parent-comment match, as in Appendix~\ref{app:accuracy-definitions}; a
comment enters the fulfilled group only when all its requirements pass.
For either counting level, the identity is
$N_{\mathrm{total}}=N_{\mathrm{missed}}+N_{\mathrm{open}}+N_{\mathrm{met}}$.
This exposes the work excluded from a conditional Filer accuracy without
changing any of the system denominators.

Fixed leaves \AGFixedUndiscoveredReq{} requirements undiscovered and
\AGFixedOpenReq{} discovered but unfulfilled. Evolving reduces these
counts to \AGEvoUndiscoveredReq{} and \AGEvoOpenReq{}, respectively.
At comment level, the corresponding groups change from
\AGFixedUndiscoveredComment{} missed and \AGFixedOpenComment{} open
comments to \AGEvoUndiscoveredComment{} and \AGEvoOpenComment{}.
Undiscovered items account for \AGEvoMissedShareReq{}\% of Evolving's
remaining unfulfilled requirements and \AGEvoMissedShareComment{}\% of
its unresolved reference comments. The residual error is thus concentrated
in concerns that never reach the revision workflow. Broader Reviewer search
must bring these omissions into the interaction before targeted revisions
and company-fact requests can address their disclosure requirements.

The discovered requirement count rises by 86 while the discovered but
unfulfilled count falls by 138, giving the aggregate gain of
$86+138=224$ fulfilled requirements. The comment-level accounting is
$39+77=116$ additional resolved comments. These are net count identities;
they do not identify which individual requirements changed state or assign
causal contributions to discovery and revision. They also explain why conditional Filer
accuracy should be read alongside Reviewer coverage: successful revision
of assigned concerns cannot resolve omissions that never enter that scope.

\AppendixSubsection{Outcome Gains and Incremental Expenditure}{app:evolution-analysis}

Evolving adds \$\AGExtraCost{} and \AGExtraHours{} recorded Agent hours
over Fixed across the complete stream. Dividing these increments by the
additional successful outcomes gives \$\AGMarginalReqCost{} per extra
fulfilled requirement and \$\AGMarginalCommentCost{} per extra fully
resolved comment. The time increment is \AGMarginalReqMinutes{} Agent
minutes per additional fulfilled requirement. These stream-level ratios
include the resources of role execution and shared services under the same
accounting used for the two system totals.

The improvement also increases output per unit of total expenditure.
Requirements fulfilled per \$100 rise from 62.2 to 67.7, and requirements
per recorded Agent hour rise from 3.25 to 3.63
(Table~\ref{tab:resource-efficiency}). Meanwhile, mean minutes per scheduled
task rise from \RQFixedMinutes{} to \RQEvoMinutes{}. The extra work can
therefore increase time per task while decreasing cost per successful
requirement. For disclosure agents, a useful optimization target combines
the cost of a task with the obligations actually resolved by its final
documents and responses.

\AppendixSubsection{Role and Shared-Service Resources}{app:resource-breakdown}

Table~\ref{tab:app-resource-composition} separates resources attributed to
Reviewer and Filer from the residual shared component. Shared cost is
\$\AGFixedSharedCost{} for Fixed and \$\AGEvoSharedCost{} for Evolving,
or \AGFixedSharedPct{}\% and \AGEvoSharedPct{}\% of their respective
system totals. It includes services that are not assigned to either role,
such as assessment, fact matching, and skill editing where applicable.
Figure~\ref{fig:app-extended-accounting}(c) visualizes this decomposition
without treating the entire shared component as the cost of evolution.

\begin{table}[!htbp]
  \captionsetup{skip=4pt}
  \caption{Attributed resources for Fixed and Evolving. Shared values subtract the two role rows from the system total.}
  \label{tab:app-resource-composition}
  \centering\normalsize
  \setlength{\tabcolsep}{6pt}\renewcommand{\arraystretch}{1.18}
  \begin{tabularx}{\linewidth}{Xrrrrrr}
    \toprule
    \rowcolor{MRPanel}
    & \multicolumn{3}{c}{\textbf{Fixed}} & \multicolumn{3}{c}{\textbf{Evolving}} \\
    \rowcolor{MRPanel}
    \textbf{Resource} & Reviewer & Filer & Shared & Reviewer & Filer & Shared \\
    \midrule
    Input tokens (M) & 3,041.21 & 2,848.36 & 209.02 & 3,126.36 & 3,076.23 & 229.92 \\
    Output tokens (M) & 15.47 & 16.47 & 3.65 & 16.17 & 17.54 & 4.19 \\
    Cost (USD) & 1,144.51 & 740.30 & 495.34 & 1,185.71 & 793.60 & 539.92 \\
    \bottomrule
  \end{tabularx}
\end{table}

The additional \$\AGExtraCost{} consists of
\$\AGExtraReviewerCost{} in Reviewer-attributed usage,
\$\AGExtraFilerCost{} in Filer-attributed usage, and
\$\AGExtraSharedCost{} in shared usage. The resource difference is
distributed across the workflow rather than confined to the editor call.
Systems that learn reusable review procedures must budget for applying
them and assessing their resulting artifacts as well as for producing the
procedure itself. Cumulative Agent hours remain a separate measure: they
sum recorded role execution and do not include independent assessment time.

\AppendixSection{Case Studies and Failure Analysis}{app:case-studies}

We examine three evidence-linked cases covering successful resolution,
partial fulfillment of a compound request, and procedural transfer to a
later task. Each example follows a selected reference concern through the
generated comment, relevant interaction, committed artifacts, and independent
requirement judgments. The figures condense the interaction wording while
retaining its numerical evidence and document targets. A requirement may
call for a filing amendment, an explanation to Staff, or a commitment in a
formal reply; the assessment checks the corresponding deliverable. These
concern-level outcomes remain distinct from the status of a broader agent
comment or the success of the complete review task, which may include
additional requests for company information and further amendments.

\AppendixSubsection{Evidence-Grounded Disclosure Resolution}{app:successful-resolution}

\begin{figure}[!htbp]
  \centering
  \includegraphics[width=\linewidth]{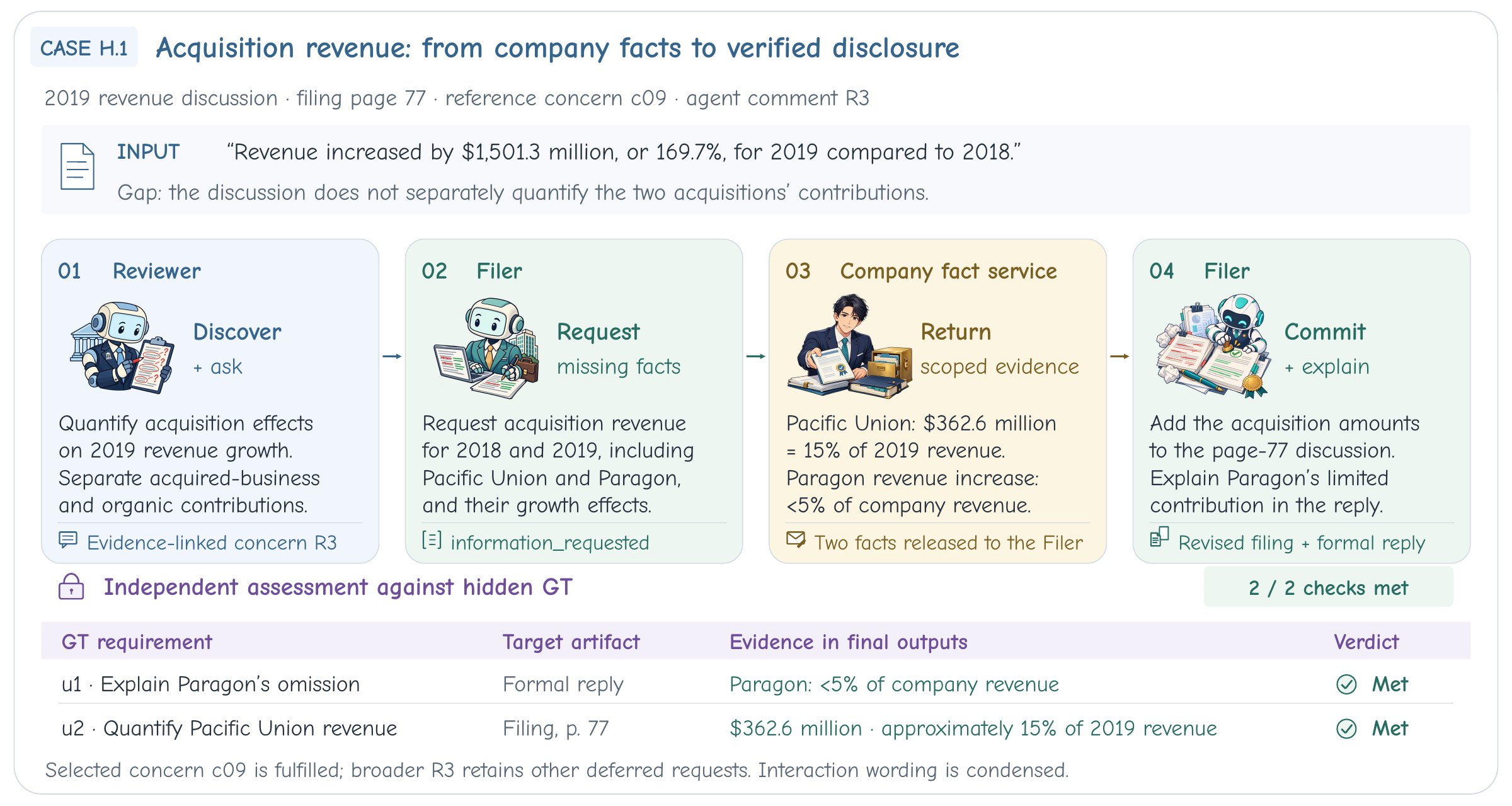}
  \caption{\textbf{Evidence-grounded resolution of an acquisition disclosure concern.}
  Scoped company facts enter the filing and formal reply, satisfying both
  selected requirements. Broader requests in agent comment R3 remain deferred.}
  \label{fig:app-case-resolution}
  \Description{Four illustrated stages show Reviewer identifying an acquisition
  revenue gap, Filer requesting facts, a company service returning Pacific Union
  and Paragon contributions, and Filer committing a revision and reply. The
  final assessment links each of two fulfilled requirements to its artifact.}
\end{figure}

The first case concerns acquisition contributions to 2019 revenue growth
(reference concern \texttt{r01:c09}). The reviewed draft reports a
\$1,501.3 million revenue increase but does not separately quantify the
contributions at issue. Reviewer comment R3 asks for acquisition effects
within a broader analysis of revenue and expense drivers. Filer requests
acquired-business revenue and growth contributions, and the company-fact
service releases two relevant facts: Pacific Union contributed
\$362.6 million, or 15\% of 2019 revenue, while the revenue increase
attributable to Paragon was less than 5\% of company revenue in 2019 or
any other period. Figure~\ref{fig:app-case-resolution} connects these
facts to the two required deliverables.

\noindent\textbf{Outcome and interpretation.}
Both selected requirements pass. The page-77 revenue discussion states
Pacific Union's dollar contribution and share of revenue, satisfying the
filing-amendment requirement. The formal reply supplies Paragon's relative
contribution, providing the basis for its earlier omission.
Pacific Union's full-year revenue contribution is not treated as its
incremental contribution to growth: the reply preserves that distinction
and identifies the broader decomposition still requiring company
information. The selected concern is therefore resolved even though R3
retains other deferred requests. This case shows why a useful disclosure
agent must route each supported fact to its required artifact and preserve
the scope of what that fact establishes. Acquiring the right number is
one step; incorporating it into the filing and the appropriate explanation
is what fulfills the disclosure obligation.

\AppendixSubsection{Compound Requirements and Failure Localization}{app:failure-taxonomy}

The second case concerns Evolus's amended purchase agreement
(reference concern \texttt{r03:c07}). A \$43.1 million estimated value of
revised obligations and a promissory note does not identify the underlying
contractual cash amounts. Reviewer R3 requests payment obligations and the
note principal. The released company facts include an approximately
\$9.2 million approval-triggered upfront payment, a \$20.0 million note,
its material terms, and reasons for withholding exact royalty rates.
The company states that precise rates are not material to investors and
that disclosure would reveal cost and margin benchmarks to negotiating
counterparties. Thus, both the numerical amendments and the supporting
confidentiality explanation have evidence available to Filer.

\begin{figure}[!htbp]
  \centering
  \includegraphics[width=\linewidth]{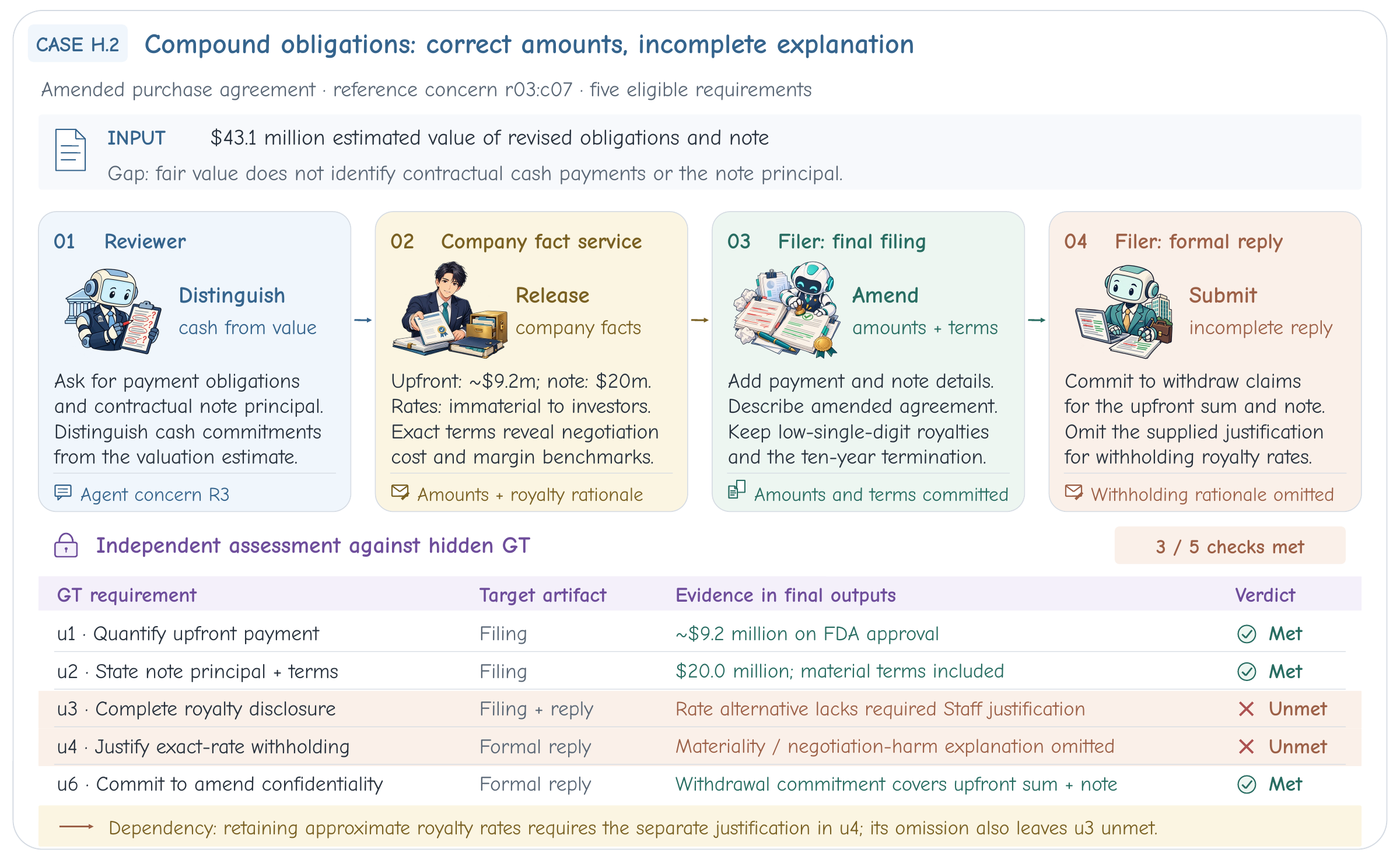}
  \caption{\textbf{Partial fulfillment of a compound request.}
  Payment amounts and note terms are corrected, but an available royalty
  withholding rationale is omitted from the reply, leaving two linked
  requirements unmet. Original eligible requirement identifiers are retained.}
  \label{fig:app-case-compound}
  \Description{Illustrated review, fact release, filing revision and reply
  stages show numerical amendments succeeding while the justification for
  withholding exact royalty rates is omitted. A five-row assessment marks
  three requirements met and two unmet, including their conditional dependency.}
\end{figure}

\noindent\textbf{Outcome and interpretation.}
Three of five eligible requirements pass. The filing corrects the amounts
and note terms; the reply commits to withdrawing their confidentiality
claims. However, retaining approximate royalty rates requires the separate
Staff explanation. Its omission leaves both that justification and the
selected disclosure alternative incomplete. Figure~\ref{fig:app-case-compound}
links the two failed checks to a shared failure: supplied evidence was not
incorporated into its required artifact. Agents should track conditional
obligations through the chosen response path, verifying that the supporting
explanation reaches its destination alongside the numerical amendments.

\AppendixSubsection{Procedural Transfer Across Cases}{app:skill-transfer-cases}

The third example follows an accepted Filer update into a later first-pass
task under the 16-task batch schedule. It comes from an earlier 64-task
run and supplies no additional observation to the main 380-task comparison.
In the first completed batch,
feedback identifies a filing whose market-maker status remains inconsistent
across the Summary, other filing text, and Exhibit~99.2
(reference concern \texttt{r01:c05}). The subsequent accepted guidance
instructs Filer to project its edits against the current draft, search the
entire projected filing for affected entities, numbers, statuses, and earlier
wording, and remove surviving contradictions. The skill is frozen for the
next batch; it transfers a verification procedure rather than the earlier
issuer's market-maker facts.

In the later case (reference concern \texttt{r01:c06}), a risk factor states
that the company irrevocably opted out of the JOBS Act extended accounting
transition period, while several other passages disclose an election to
use it. Reviewer R1 cites the conflicting locations. Filer replaces the
opt-out statement with the supported election, adds the resulting
financial-statement comparability risk, and makes conforming changes in
the accounting note. The committed risk-factor text satisfies both
selected requirements: it reconciles the election and explains that the
financial statements may not be comparable to those of issuers adopting
standards on public-company effective dates. Requests for additional
company confirmation and adoption details remain separately identified.

\begin{figure}[H]
  \centering
  \includegraphics[width=\linewidth]{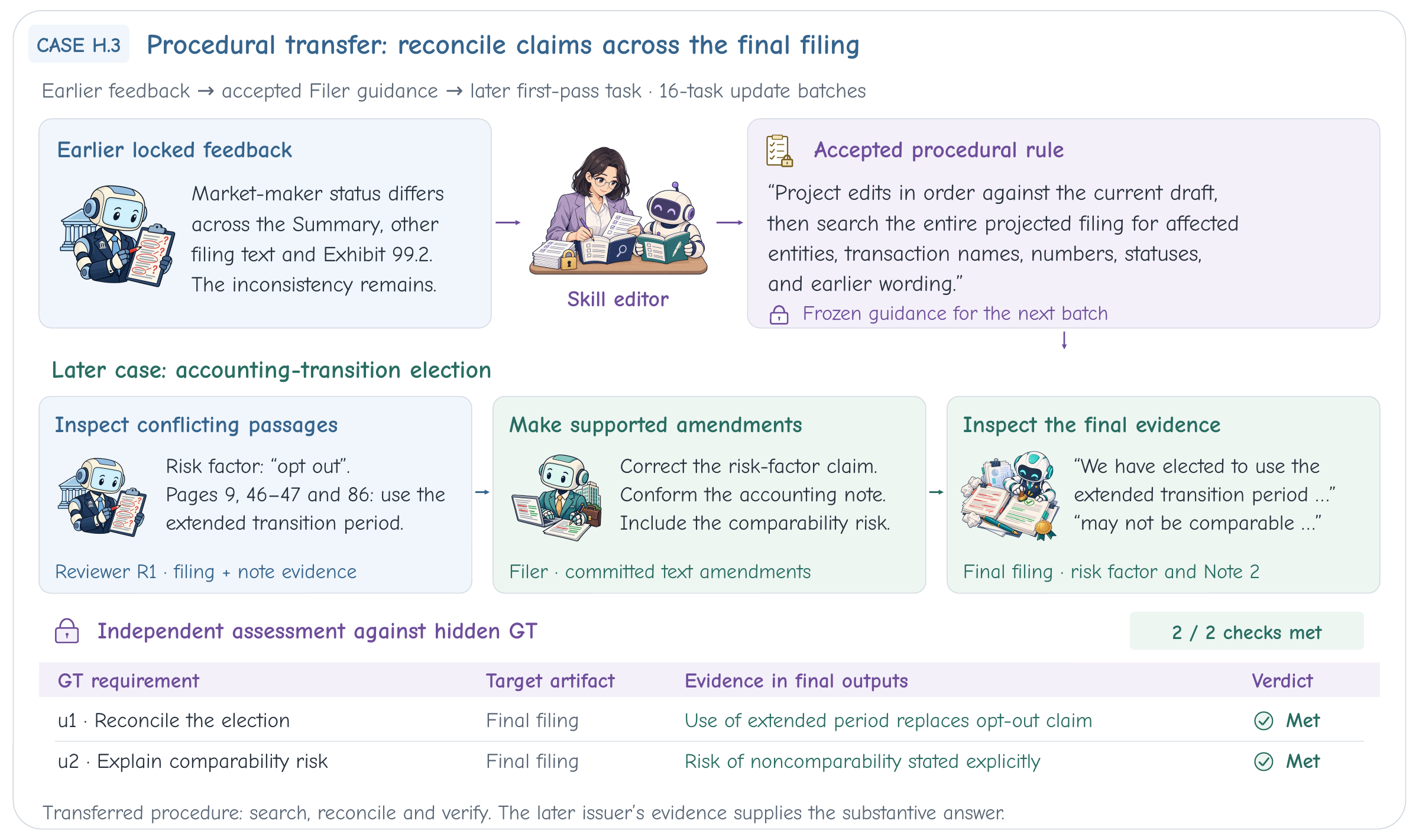}
  \caption{\textbf{Procedural transfer to a later task.}
  Contradiction feedback informs an accepted filing-wide verification rule.
  A later accounting-election case connects that procedure to committed
  revisions and two fulfilled requirements.}
  \label{fig:app-case-transfer}
  \Description{Earlier feedback about inconsistent market-maker statements
  leads through an illustrated editor to a procedural search rule. A later
  task reconciles conflicting accounting-transition statements in a risk
  factor and note, and the final assessment marks both selected checks met.}
\end{figure}

\noindent\textbf{Outcome and interpretation.}
Figure~\ref{fig:app-case-transfer} links the earlier inconsistency feedback,
the accepted rule, and the later document revisions. The subject changes
from market-maker arrangements to accounting elections, while the
transferable operation remains the same: locate competing statements,
reconcile the supported position across required locations, and inspect
the resulting text. Learning at this level supports a reusable action
sequence without turning a previous company's answer into evidence for
the current issuer. It also makes the learned procedure inspectable:
the rule specifies what to search, which state to check, and what counts
as completion in the final artifact.

% Keep this final illustration with H, ahead of the following appendix section.

\AppendixSection{Limitations and Responsible Data Use}{app:limitations}

FilingLoop evaluates evidence-backed disclosure resolution in reconstructed
review interactions. Its conclusions concern the supplied historical records,
task eligibility rules, agents, and online protocol. The following boundaries
guide interpretation of the results and reuse of the data and framework.

\AppendixSubsection{Historical Reconstruction and Coverage}{app:benchmark-limitations}

The corpus is concentrated on IPO registration review, with a deliberately
small annual-report extension from one existing issuer. It spans domestic and
foreign companies but does not represent every filing type, jurisdiction,
industry, or stage of regulatory review. The two annual-report tasks extend
the task format without establishing broad annual-report generalization.
Results should therefore be interpreted in relation to the corpus composition
in Appendix~\ref{app:data-construction}.

Historical correspondence records the concerns staff raised and the responses
that became available. It does not enumerate every defensible concern or
every acceptable resolution. FilingLoop consequently assesses additional
generated concerns on their evidence and substance rather than treating an
absent historical match as an error. Eligibility also selects requirements
whose requested actions and necessary facts have attributable support. This
makes fulfillment evaluable while limiting conclusions about cases requiring
unrecorded information, negotiations, or other unavailable company evidence.

Temporal reconstruction controls what the agent can inspect, but cannot
establish what a language model encountered during pretraining. Public filing
and correspondence text may have appeared in training data. Repeated rounds
and requirements from the same company are also related observations, so the
number of requirements should not be read as the number of independent
companies or independent regulatory decisions represented by the benchmark.

\AppendixSubsection{Evaluator Validity and Online Adaptation}{app:evaluation-limitations}

Model-assisted annotation and evaluation can misinterpret an accounting
question, overlook relevant context, or judge an alternative response
inconsistently. Independent annotation checks, source-linked requirements,
structured judgment schemas, and quotation validation improve traceability,
but mechanical validation does not establish semantic correctness. The
evaluator underwent manual review, and all configurations share its grading
model, prompts, and artifact checks. Human--grader agreement and cross-grader
sensitivity are not quantified. Results are specific to the evaluated task
sample and order in this single seeded stream. The
acting Reviewer and the independent evaluator have different responsibilities;
the former's closure decision cannot replace the latter's requirement check.

The bounded fact service supplies historical evidence associated with matched
concerns. It does not reproduce unrestricted access to company personnel or
guarantee answers to arbitrarily phrased questions. Its response reflects
recorded evidence coverage, and a failed match is distinct from an issuer's
inability to answer. The task therefore measures fact acquisition within this
defined interface rather than every aspect of a live counsel--company exchange.

Evolution uses feedback from earlier tasks in the evaluation stream, including
earlier rounds of issuers that can reappear later. The resulting comparison
measures online procedural adaptation under the specified order. It does not
establish company-disjoint transfer or robustness to every alternative stream
order. Textual filters on learned skills reject known identifiers and direct
reference copying, but cannot guarantee that all paraphrased case information
has been excluded. Fixed prompt precedence and case-specific evidence checks
remain necessary when applying learned procedures to subsequent disclosures.

Both roles receive procedural updates within the same interaction protocol.
The treatment in the paired comparison is therefore the joint evolving
policy, while role accuracies describe discovery and conditional resolution
within that treatment. Their denominators and information dependencies should
be preserved when interpreting contributions at each stage. Framework
comparisons likewise account for native tool loops and measured expenditure
alongside backbone identity and the shared artifact contract.

\AppendixSubsection{Source Attribution and Data Handling}{app:data-use}

The source records are public SEC filings and correspondence, and derived
tasks retain company, accession, date, and source-span provenance. Role-private
company answers are drawn from the same historical collection. Their visibility
is restricted to model the Filer's access to supporting evidence during an
interaction, without giving the Reviewer the hidden fact ledger. The term
private therefore describes a role boundary in the experiment and does not
imply that confidential issuer records or direct communications with company
personnel were obtained for evaluation. Released answers retain their
connection to the public source evidence used to construct each task.

Reuse should preserve document attribution and the distinction between
original statements, derived annotations, and model-generated revisions.
Names and other personal information can occur in public filings, so
distributing task examples requires attention to the specific fields and
source passages included. Credentials, account information, and unrelated
workspace files are excluded from task materials. Generated figures use
fictional characters and identify software roles rather than depicting SEC
officials or company representatives involved in the source reviews.

\AppendixSubsection{Intended Use and Human Oversight}{app:deployment-boundaries}

The framework supports research on document-grounded review, revision, and
procedural learning. Generated comments and amendments are experimental
artifacts; they are not SEC communications, executed agreements, obtained
consents, or submitted filings. A successful edit transaction proves that a
specified document operation was committed, not that the issuer performed
an external action described in the text.

Practical use requires qualified review of the source evidence, requested
changes, accounting interpretation, and final deliverables. Unsupported
company assertions or inconsistent revisions can remain consequential even
when aggregate accuracy improves. The separate harm audits and strict
comment, task, and company metrics expose these obligations rather than
allowing a high requirement count to substitute for complete verification
of the documents and commitments on which a real review decision depends.

\end{document}